\documentclass[10pt,twocolumn,letterpaper]{article}

\usepackage[pagenumbers]{cvpr}      
\definecolor{cvprblue}{rgb}{0.21,0.49,0.74}
\usepackage[pagebackref,breaklinks,colorlinks,allcolors=cvprblue]{hyperref}

\def\confName{CVPR}
\def\confYear{2027}

\usepackage{graphicx}
\usepackage{amsmath,amssymb}
\usepackage{booktabs}
\usepackage{microtype}
\usepackage{xcolor}

\usepackage{natbib}
\usepackage{placeins}
\usepackage{float}
\usepackage{fvextra}

\DefineVerbatimEnvironment{PaperPrompt}{Verbatim}{
    fontsize=\footnotesize,
    breaklines=true,
    breakanywhere=false
}

\input{glyphtounicode}
\title{Semantic Apertures: Recognizing Structure Under Competing Visual Content}

\author{
Kianoosh Vadaei$^{1}$\thanks{Equal contribution.},
Melika Shirian$^{1}$\footnotemark[1],
Mohammad Sabokrou$^{2}$\\
$^{1}$University of Isfahan, Isfahan, Iran\\
$^{2}$Okinawa Institute of Science and Technology (OIST), Okinawa, Japan\\
{\tt\small melika.shi@mehr.ui.ac.ir, k.vadaei@mehr.ui.ac.ir}\\
{\tt\small mohammad.sabokrou@oist.jp}
}

\hypersetup{
    pdftitle={
        Semantic Apertures: Recognizing Structure Under Competing Visual Content
    },
    pdfauthor={
        Melika Shirian, Kianoosh Vadaei, Mohammad Sabokrou
    },
    pdfsubject={
        Computer Vision, Shape Recognition, Visual Robustness,
        and Visual Representation Analysis
    },
    pdfkeywords={
        Semantic Apertures,
        Computer Vision,
        Shape Recognition,
        Shape Bias,
        Texture Bias,
        Visual Robustness,
        Object Recognition,
        Visual Perception,
        Shape-Appearance Conflict,
        Vision-Language Models,
        Visual Representation Learning,
        Representation Decodability,
        Linear Probing,
        Human-Machine Comparison,
        Benchmarking
    },
    pdfcreator={LaTeX},
    pdflang={en-US}
}

\begin{document}
\maketitle

\begin{abstract}
An object's outline can provide cues to its identity even
when it reveals unrelated visual content, as in a shape
wiped into a fogged mirror. We introduce
\emph{Semantic Apertures}\footnote{The dataset is publicly
available at \url{https://huggingface.co/datasets/smntccccc/semantic_apertures}.},
a benchmark of 14,400 composites built from 600 photographs
across 100 categories, in which a target-derived aperture exposes independently
assigned scene content through heavy fog, allowing us
to examine structural recognition under altered visibility
and visual appearance. Across
eight frozen checkpoints, top-1 accuracy falls from
12.44--65.06\% on isolated structure to 0.17--23.08\% on
final composites. A matched construction ladder shows that
the placement stage largely preserves recognition, while
accuracy collapses at the heavy-fog stage. In a separate
ten-choice human study on 500 selected final composites,
image-macro accuracy reaches 69.83\%. Paired
interventions show that correct geometry still increases
target evidence, even as changes to the exposed (carrier)
content alter predictions. Shape-trained linear probes
transfer poorly to final composites (1.33--12.21\%),
whereas separately fitted Final-trained probes recover
partial target-category information (up to 36.35\%)
within the tested rendering distribution. On the 498
images scored by both endpoints, using the same ten
options, \texttt{gpt-6-luna} and \texttt{gemini-3.8-flash}
reach 51.6\% and 79.3\% Final-target accuracy. GPT falls
short of the human rate and Gemini exceeds it, while both
lose recognition under composition. Together, these findings distinguish
cross-view readout portability from within-view category
decodability without establishing that the recovered
information is specifically geometric.
\end{abstract}


\begin{figure}[!t]
    \centering
    \includegraphics[width=\columnwidth]{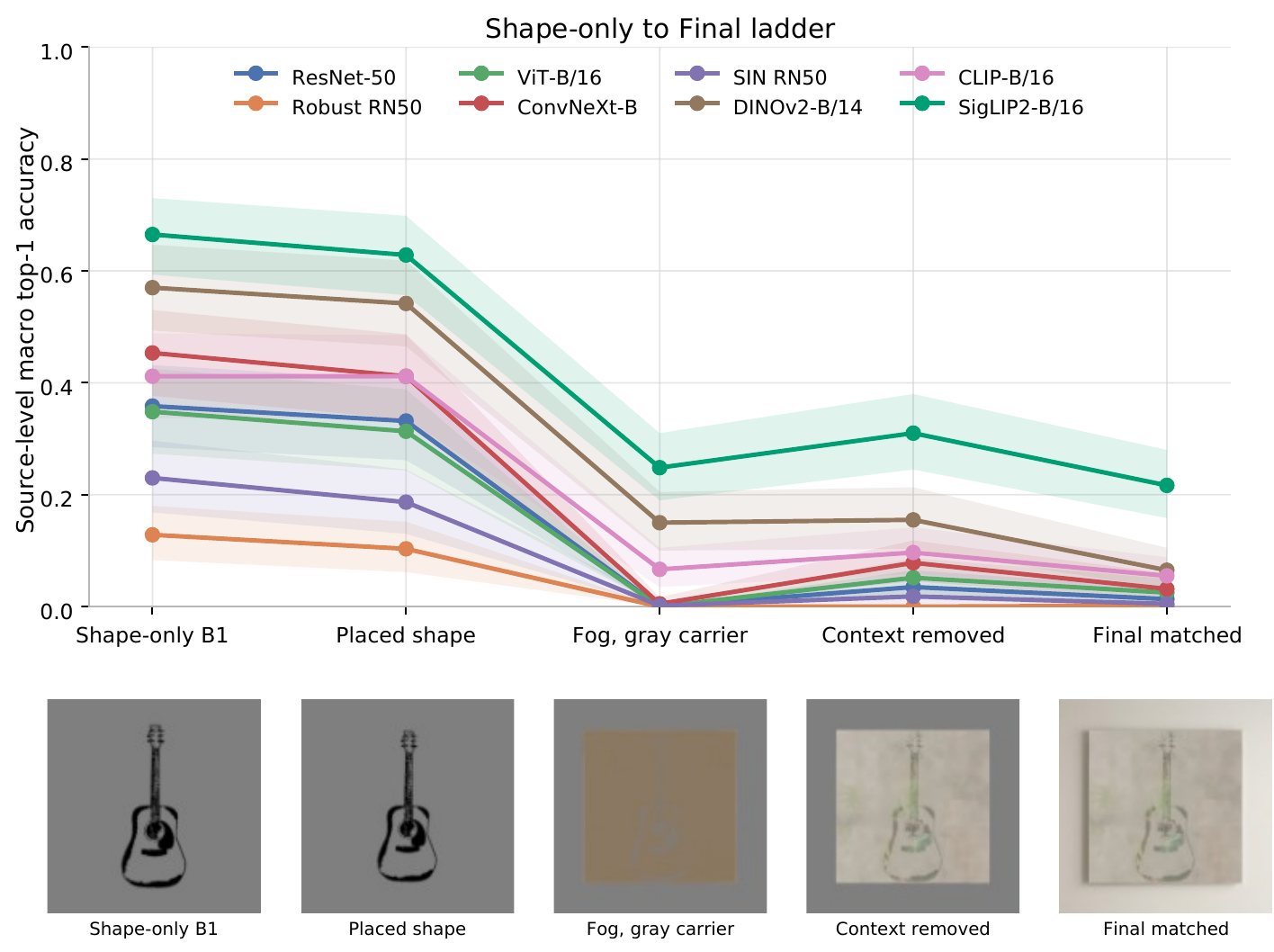}
    \caption{
    \textbf{Matched Shape-only to Final ladder.}
    Native ImageNet-1000 top-1 for eight frozen checkpoints
    on the same 600 B1 sources (95\% hierarchical bootstrap CIs).
    The lower row shows the matched stages for one acoustic-guitar
    source. The beige color at the gray-carrier step comes from fog.
    Final matched uses one render per source.
    }
    \label{fig:recognition_ladder}
\end{figure}

\begin{figure*}[t]
    \centering
    \includegraphics[
        width=0.84\textwidth,
    ]{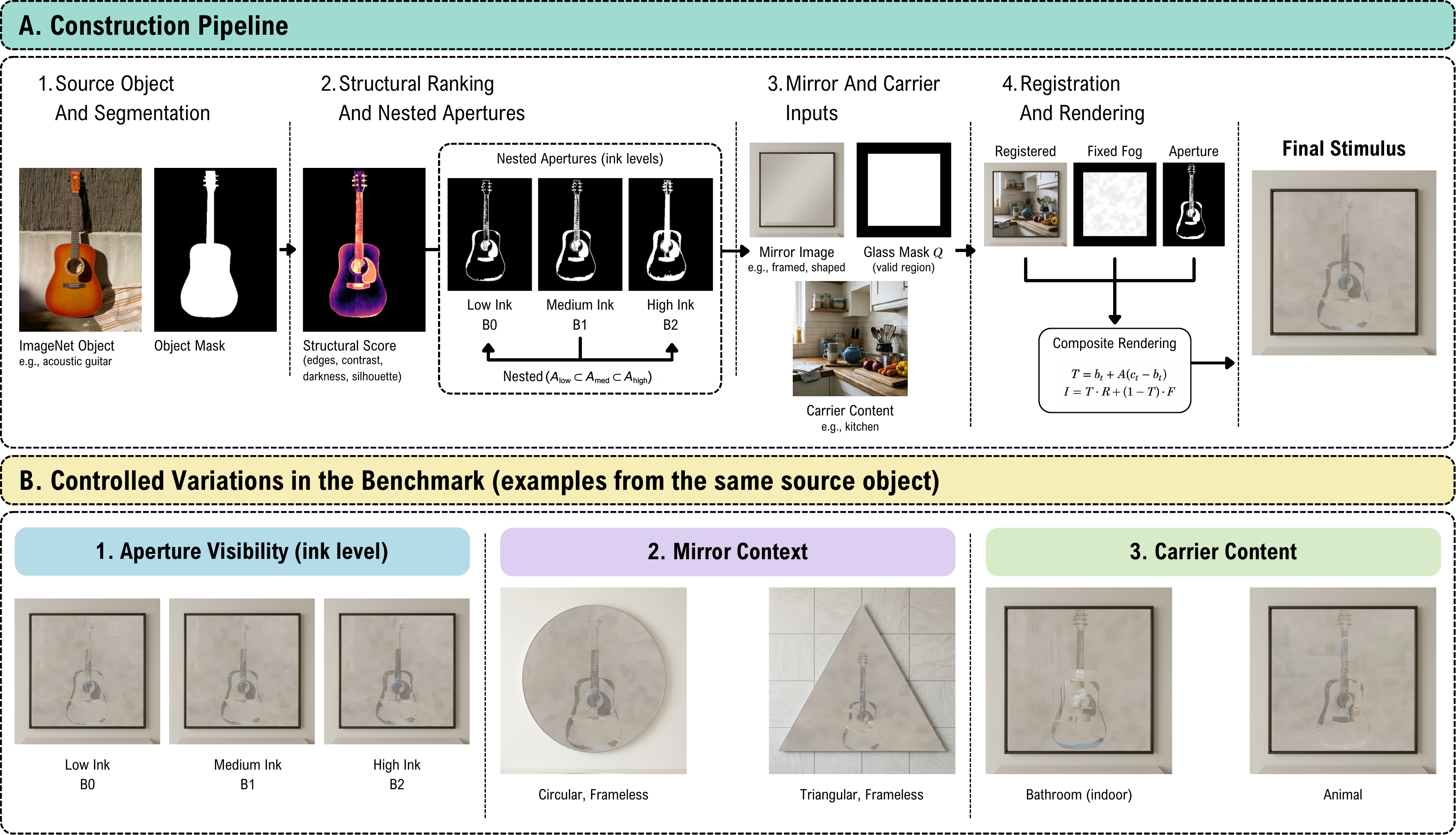}
    \caption{
    \textbf{Semantic Apertures construction and variations.}
    A source photograph and its object mask define three nested structural
    apertures. The selected aperture is placed within a mirror's glass
    region and reveals separately assigned carrier content through fog.
    The lower examples illustrate variations in aperture support,
    mirror/context asset, and carrier appearance.
    The recognition target is the source object's category.
    }
    \label{fig:semantic_apertures_overview}
\end{figure*}

\section{Introduction}
\label{sec:introduction}

Visual recognition relies on appearance and spatial
organization~\citep{geirhos2019texture,lonnqvist2025contour}.
Consider an object-shaped opening in fogged glass:
its geometry suggests one object, while its colors
and textures belong to another scene. How reliably
do recognition models use spatial structure when
geometry and appearance conflict?

Prior work documents texture bias and the effects
of stylized training~\citep{geirhos2019texture}.
\emph{Hidden in Plain Sight} introduced IllusionBench,
revealing VLM difficulties with shapes formed by
scene-element arrangements
~\citep{hemmat2024illusionbench}.
Yet recognizing isolated structure, responding to
its geometry within a composition, and transferring
a readout across views are distinct capabilities.
A failed prediction cannot distinguish absent structural
information from information unused by the native readout.

We introduce \emph{Semantic Apertures}, a synthetic,
mirror-inspired benchmark in which target-derived
apertures reveal independently assigned carrier scenes
(Figure~\ref{fig:semantic_apertures_overview}).
It comprises 14,400 composites from 600 photographs
across 100 categories, with three nested aperture
levels, two mirror assignments, and four carriers
per source under fixed heavy fog. Balanced assignments
support paired interventions, while diagnostic views
trace recognition from structure to composition.
Unlike background replacement, the target's
structure determines where independently assigned
scene content becomes visible, allowing paired
interventions to examine sensitivity to aperture
geometry and transmitted appearance.

Across eight frozen checkpoints, native ImageNet-1000
top-1 falls from 12.44--65.06\% on Shape-only to
0.17--23.08\% on Final. The matched ladder shows
that recognition largely survives placement but
declines under heavy fog on a gray carrier, before
photographic carrier content appears
(Figure~\ref{fig:recognition_ladder}).
Because aperture contrast and appearance also change,
the decline cannot be attributed solely to fog or
content conflict. Paired interventions show that
correct geometry increases target evidence, while
carrier appearance affects predictions across models.

Source-disjoint probes distinguish Shape-trained
readout transfer from within-Final category decoding.
Unchanged readouts transfer poorly, whereas Final-trained
probes recover partial category information. The
36.35\% DINOv2 result uses a render-level fit with
the Shape-selected penalty; a separate equal-source-weight
head evaluates aperture controls. Correct geometry
outperforms scrambled and absent apertures, while
replacements alter category predictions. Donor pools
include training sources, and the small asset-disjoint
diagnostic does not establish broad generalization.
These controls indicate category sensitivity without
isolating exclusively geometric information.

In a ten-choice human study on 500 selected Final
images, image-macro accuracy reaches 69.83\%.
The images and options were evaluated with
\texttt{gpt-6-luna} and \texttt{gemini-3.8-flash}.
On 498 common images, Final-target accuracy is
51.6\% and 79.3\%, respectively, versus 69.9\%
for human responses, showing no uniform
human--machine gap. Both lose target and carrier
recognition under composition, with
Shape-to-Final retention of 59.4\% and 84.3\%
(Section~\ref{sec:vlm_recognition}).
Across seven encoders, ten-way accuracy on 500
Final images is 3.8--30.2\%, below human and
VLM results. Prompts, response counts, and
aggregation still differ.

Our contributions are:
\begin{enumerate}
    \item \textbf{A benchmark for structure--content composition.}
    We introduce nested target-derived apertures,
    balanced mirror/carrier assignments, reproducible
    rendering, and diagnostic recognition views
    (Section~\ref{sec:semantic_apertures};
    Appendix~\ref{app:benchmark_construction}).

    \item \textbf{Paired evidence of geometry and carrier dependence.}
    Controlled interventions quantify sensitivity
    to spatial arrangement and rendered carrier
    appearance, alongside changes in target evidence
    (Section~\ref{sec:geometry_or_content};
    Appendix~\ref{app:geometry_carrier_results}).

    \item \textbf{Cross-view readout diagnostics and complementary recognition.}
    Source-disjoint probes distinguish Shape-only
    decoding, unchanged readout transfer, and
    within-Final category decoding. Aperture controls,
    human validation, matched VLM queries, and ten-way
    encoder evaluations provide complementary evidence
    under composition
    (Sections~\ref{sec:semantic_apertures},
    \ref{sec:structural_stability}, and
    \ref{sec:vlm_recognition};
    Appendices~\ref{app:probe_details},
    \ref{app:human_validation}, and
    \ref{app:extended_vlm}).
\end{enumerate}

\section{Related Work}
\label{sec:related_work}

\paragraph{Shape and appearance cues.}
\citet{baker2018global} show that CNNs may rely on local
contour features despite disruptions to global shape.
\citet{geirhos2019texture} demonstrate texture preference
in ImageNet-trained CNNs and show that stylized training
can increase shape bias. \citet{gavrikov2025seeing} further
show that language prompts influence VLM cue preferences.
We study a complementary setting in which object-derived
spatial structure reveals independently assigned scene
appearance, rather than texture belonging to the target
object itself.

\paragraph{Hidden-shape recognition.}
IllusionBench studies the recognition of abstract shapes
formed by arrangements of scene elements alongside scene
recognition \citep{hemmat2024illusionbench}.
Semantic Apertures instead combines target-derived
apertures with independently assigned carrier content
and matched construction views. Its construction ladder
locates a substantial recognition decline at the
heavy-fog/contrast stage, while paired geometry and
carrier interventions quantify sensitivity to spatial
arrangement and rendered appearance. Source-disjoint
probes additionally distinguish cross-view readout
portability from within-view category decodability.
These controlled diagnostics constitute our methodological
distinction, rather than a claim of performance superiority
over IllusionBench.

\paragraph{Spatial integration and configuration.}
\citet{lonnqvist2025contour} investigate contour
fragmentation and directional integration, while
\citet{doshi2025anagrams} examine spatial configuration
through patch-preserving object anagrams.
Our aperture interventions compare correct arrangements
with scrambled and replacement geometry under matched
mirror--carrier assignments. Although scrambles preserve
aperture area, other geometric properties are only
approximately matched
(Appendix~\ref{app:geometry_carrier_results}).

\paragraph{Background and contextual dependence.}
\citet{xiao2021background} show that classifiers exploit
background information, including misleading backgrounds.
\citet{rosenfeld2018elephant} show that object
transplantation can affect detections elsewhere in a scene.
Rather than recombining an intact foreground object with
a background, we vary the appearance transmitted through
a target-derived aperture and compare correct, scrambled,
wrong-class, and absent geometry.
Carrier swaps also modify the carrier-dependent fog plate;
thus, these interventions establish sensitivity to
rendered appearance, not an isolated semantic-content
effect.

\paragraph{Robustness benchmarks.}
ImageNet-C \citep{hendrycks2019corruptions} evaluates common
corruptions, ImageNet-R \citep{hendrycks2021manyfaces}
examines diverse visual renditions, and ImageNet-Sketch
\citep{wang2019sketch} tests sketch-like inputs.
These benchmarks characterize recognition under changes
in image appearance. Semantic Apertures instead combines
target-derived spatial structure with independently assigned
carrier appearance, using matched construction stages
and paired interventions to examine recognition decline
and decision sensitivity. This provides a complementary
diagnostic setting rather than a direct benchmark comparison.

\paragraph{Decodability and readout use.}
Linear probes measure accessible class information
\citep{alain2016probes}, while \citet{fu2025overlook}
show that VLM responses may underuse information
accessible through encoder readouts.
We separately evaluate Shape-only category decoding,
transfer of an unchanged Shape-trained readout to Final
images, and category decoding by independently fitted
Final-trained readouts on source-disjoint splits.
Comparison with native predictions restricted to the same
100 categories distinguishes cross-view readout
portability from within-view category decodability,
without assuming geometry-specific information recovery.

\section{Semantic Apertures}
\label{sec:semantic_apertures}

Semantic Apertures evaluates recognition when object-derived
spatial structure reveals independently assigned appearance.
The aperture determines where the carrier is visible; the
carrier determines what appears there. Target and carrier
categories may occasionally agree. Figure~\ref{fig:semantic_apertures_overview}
illustrates the construction.

\begin{table*}[t]
    \centering
    \caption{
    \textbf{Recognition across the Semantic Apertures construction.}
    Native ImageNet-1000 top-1 and top-5 target accuracy (\%),
    aggregated over 600 sources and 100 equally weighted categories.
    Shape-only averages three apertures per source; Final averages
    24 compositions per source.
    Bold indicates column maxima (including ties), not statistical
    significance. Confidence intervals are reported in
    Appendix~\ref{app:recognition_results}.
    }
    \label{tab:recognition_transition}

    \renewcommand{\arraystretch}{1.20}
    \begin{tabular*}{0.84\textwidth}{
        @{\extracolsep{\fill}}
        l
        *{8}{r}
        @{}
    }
        \toprule
        &
        \multicolumn{2}{c}{\textbf{Original}} &
        \multicolumn{2}{c}{\textbf{Object Gray}} &
        \multicolumn{2}{c}{\textbf{Shape-only}} &
        \multicolumn{2}{c}{\textbf{Final}} \\
        \cmidrule(lr){2-3}
        \cmidrule(lr){4-5}
        \cmidrule(lr){6-7}
        \cmidrule(lr){8-9}

        \textbf{Model}
        & Top-1 & Top-5
        & Top-1 & Top-5
        & Top-1 & Top-5
        & Top-1 & Top-5 \\
        \midrule

        ResNet-50
        & 91.0 & 98.8
        & 90.3 & 97.7
        & 32.2 & 49.1
        & 1.6 & 4.7 \\

        Robust RN50
        & 81.3 & 93.5
        & 72.3 & 89.5
        & 12.4 & 25.6
        & 0.2 & 0.5 \\

        ViT-B/16
        & 89.7 & 98.3
        & 88.3 & 96.2
        & 31.8 & 51.3
        & 2.0 & 5.3 \\

        ConvNeXt-B
        & \textbf{93.2} & 99.2
        & \textbf{91.3} & 97.8
        & 41.2 & 63.1
        & 2.2 & 7.7 \\

        SIN RN50
        & 76.0 & 90.2
        & 67.0 & 85.0
        & 21.3 & 40.3
        & 0.7 & 1.9 \\

        DINOv2-B/14
        & 92.8 & \textbf{99.7}
        & \textbf{91.3} & \textbf{98.7}
        & 55.4 & 76.3
        & 6.1 & 12.5 \\

        CLIP-B/16
        & 80.7 & 93.7
        & 76.5 & 92.7
        & 40.8 & 63.1
        & 5.9 & 15.0 \\

        SigLIP2-B/16
        & 91.0 & 99.2
        & 90.0 & 97.7
        & \textbf{65.1} & \textbf{82.2}
        & \textbf{23.1} & \textbf{39.6} \\

        \bottomrule
    \end{tabular*}
\end{table*}

\subsection{Source Selection and Structural Apertures}
\label{sec:aperture_construction}

We select 600 ImageNet validation
photographs~\citep{deng2009imagenet,russakovsky2015imagenet} with ImageNet-S
masks~\citep{gao2022luss}: six sources per category across
100 categories covering animals, vehicles, devices,
household objects, and food. Selection prioritizes segmentation geometry rather than
model confidence (Appendix~\ref{app:source_selection_details}).

For source $i$ with object support $M_i$, we construct
three nested binary apertures:
\begin{equation}
    A_i^{B0}
    \subseteq
    A_i^{B1}
    \subseteq
    A_i^{B2}
    \subseteq
    M_i.
    \label{eq:nested_ink}
\end{equation}
Nesting holds in source coordinates before placement.
$B0$ combines the silhouette rim, dark structural regions,
and internal edges using lightness, local contrast,
illumination-adjusted detail, and Sobel features.
$B1$ relaxes darkness and edge thresholds while retaining
$B0$; $B2$ can add further pixels.
Increasing aperture level changes both structural
support and transmitted carrier area, rather than isolating
contour detail. Extraction parameters appear in
Appendix~\ref{app:benchmark_construction}.

\subsection{Structure--Content Composition}
\label{sec:aperture_composition}

The mirror bank contains 45 assets crossing five shapes,
three surrounding types, and three frame styles, each with
a glass-region mask. The carrier bank contains 30 images
across ten categories. The fixed asset banks support
balanced, reproducible comparisons across geometry and
carrier assignments. The renderer registers carrier
appearance within the glass, producing synthetic
mirror-inspired compositions without modeling
three-dimensional reflection optics.

Placement is computed separately per level, so source-coordinate
nesting does not ensure identical rendered scales.

For softened, placed aperture $\bar{A}(u)\in[0,1]$,
carrier transmission is
\begin{equation}
    T(u)
    =
    t_{\mathrm{bg}}
    +
    \bar{A}(u)
    \left(
        t_{\mathrm{clear}}-t_{\mathrm{bg}}
    \right),
    \label{eq:transmission}
\end{equation}
where $t_{\mathrm{bg}}=0.006$ and
$t_{\mathrm{clear}}=0.52$ under fixed heavy fog.
The aperture therefore transmits more carrier appearance
than surrounding fogged glass.

Rendering blends carrier appearance with a fog plate in
linear color space, applies glass and edge treatments,
and restores pixels outside the glass mask before JPEG
encoding. Only one heavy-fog, no-droplet configuration
is evaluated.

The fog plate itself retains carrier-dependent appearance;
carrier swaps consequently affect visible content throughout
the glass, not solely within the aperture. Thus, paired
carrier interventions measure rendered-appearance
sensitivity rather than isolated aperture-content effects
(Appendix~\ref{app:benchmark_construction}).

\subsection{Balanced Assembly and Recognition Views}
\label{sec:benchmark_assembly}

Each of the 600 independent sources receives two differently
shaped mirrors, four carriers from distinct categories, and
three aperture levels, yielding 24 Final renders per source
and 14,400 in total. Global assignments balance mirror shape,
surroundings, frame style, and carrier-category usage
(Appendix~\ref{app:assembly_qc_details}).

Four recognition views trace construction:
\emph{Original} (source photograph),
\emph{Object Gray} (photographic object on gray),
\emph{Shape-only} (black aperture on gray), and
\emph{Final} (carrier-filled mirror composite).
They contain 600, 600, 1,800, and 14,400 images,
respectively.

Shape-only-to-Final jointly changes placement, scale,
contrast, fog, context, and transmitted appearance.
A matched ladder separates these factors on one B1
assignment per source, using the mirror, carrier, and
placement of the existing context-removed reference.
Each step contains 600 images (Figure~\ref{fig:recognition_ladder}).
Final matched uses one paired render per source; the Final view
above averages 24 compositions. Appendix~\ref{app:recognition_ladder}
gives the five matched conditions and complete results.

\subsection{Human Recognizability Validation}
\label{sec:human_recognition}

To assess human recognizability, we evaluated a selected
subset of 500 Final images (five per category) through
ten-choice survey forms. Across 51 complete response
records and 2,550 judgments, image-macro target accuracy
reached \textbf{69.83\%} (95\% CI: 65.41--74.10\%;
pooled accuracy: 69.57\%). Of 500 images, 188 received
only correct judgments, while 31 received none.
Distinct participant identities could not be verified
across forms. Participants provided informed consent;
no formal institutional ethics review was conducted.
Selection details, uncertainty analysis, and limitations
appear in Appendix~\ref{app:human_validation}.

\section{Experimental Setup}
\label{sec:experimental_setup}

We evaluate recognition across construction views,
paired geometry and carrier interventions, and
structural readout transfer. Eight frozen checkpoints
provide native predictions; seven also provide features
for interventions and probes.

A separate paired VLM study uses the human study's
500 Final images and ten-choice target options.
For each image, five queries assess target recognition
(Original, Placed Shape-only, Final) and carrier
recognition (clean carrier, Final). The endpoints are
\texttt{gpt-6-luna} and \texttt{gemini-3.8-flash},
both unpinned provider aliases, with 2{,}500 scheduled
queries each. The primary cohort contains 498 images
with scored responses to all five queries from both
endpoints. Invalid outputs count as incorrect; technical
failures remain missing. Accuracy and paired changes
are category-macro, while conditional retention is
image-pooled (Section~\ref{sec:vlm_recognition};
Appendix~\ref{app:extended_vlm}).

\subsection{Models and Native Readouts}
\label{sec:model_suite}

We evaluate supervised ResNet-50~\citep{he2016resnet},
ViT-B/16~\citep{dosovitskiy2021vit}, and
ConvNeXt-B~\citep{liu2022convnext};
adversarially trained ResNet-50 (Robust RN50,
$\ell_\infty=4/255$)~\citep{salman2020transfer};
Stylized-ImageNet ResNet-50~\citep{geirhos2019texture};
self-supervised DINOv2-B/14~\citep{oquab2023dinov2};
CLIP-B/16~\citep{radford2021clip}; and
SigLIP2-B/16~\citep{tschannen2025siglip2}.
All encoders remain frozen with native preprocessing.
Classification checkpoints retain pretrained heads;
DINOv2 uses its supplied ImageNet classifier, while
CLIP and SigLIP2 use ensembled text prompts with
normalized image--text similarity and sigmoid scores,
respectively. SigLIP2 contributes native results for
the four recognition views, matched ladder, and
locked-fog comparison. Probes use pre-classifier
features or normalized CLIP embeddings
(Appendix~\ref{app:evaluation_details}).

We report native ImageNet-1000 top-1/top-5 accuracy.
The matched ladder uses the same frozen readouts
on one B1 image per source at each step
(Section~\ref{sec:benchmark_assembly}).
\emph{Native Benchmark-100} restricts unchanged
scores to the 100 benchmark categories, matching
the probe label space without refitting weights.

\subsection{Carrier, Geometry, and Context Interventions}
\label{sec:intervention_setup}

Carrier interventions vary four carriers at fixed
source, mirror, and aperture level, yielding 3,600
blocks per checkpoint. We measure prediction
switching and original-target score variation.

Geometry interventions compare one deterministic
B1 mirror--carrier reference per source against
no aperture, three scrambles (tile, row-band, and
column-band), and wrong-class geometry. Scrambles
preserve area exactly but only approximately match
other geometric properties
(Appendix~\ref{app:geometry_carrier_results}).
Context removal replaces the surround with gray;
mirror-region cropping also changes effective scale,
so it cannot isolate localization alone.
All geometry and context contrasts are source-paired
against the same existing Final reference.
The locked-fog comparison uses the same 600 B1
sources (Section~\ref{sec:recognizing_structure}).

\subsection{Source-Disjoint Structural Readouts}
\label{sec:probe_setup}

We fit ridge classifiers~\citep{hoerl1970ridge}
on standardized frozen features across six
source-disjoint rotations, each with 400/100/100
training/validation/test sources. All views of a
source remain in the same fold. Shape-trained
probes select $C\in\{0.01,0.1,1,10\}$ by Shape-only
validation accuracy, breaking ties toward smaller
$C$. Training-fitted standardization and weights
transfer unchanged to held-out Shape-only and Final
images. Categories and rendering asset banks remain
shared across folds (Appendix~\ref{app:probe_details}).

The 36.35\% DINOv2 Final result uses a render-level
ridge fit with the Shape-selected $C$, refitting
standardization and weights on 9,600 Final renders
from 400 training sources. For aperture tracking
(Section~\ref{sec:structural_stability}), we instead
fit a source-weighted multinomial logistic regression
head on Final features, selecting $C$ by source-macro
Final validation accuracy. Its scaler and weights
are frozen before evaluating aperture controls and
the exploratory 32-source asset-disjoint test
(Appendix~\ref{app:aperture_tracking_details}).
These heads use different fitting and regularization
procedures. Chance accuracy over 100 classes is $1/100$.

\section{Recognizing Structure Under Competing Content}
\label{sec:recognizing_structure}

We evaluate native ImageNet-1000 recognition across
Original, Object Gray, Shape-only, and Final views
(Table~\ref{tab:recognition_transition}).
Full intervals and Benchmark-100 results appear in
Appendix~\ref{app:recognition_results}.

\paragraph{Recognition declines with structural isolation.}
Original top-1 accuracy is 76.0--93.2\%, remaining
67.0--91.3\% on Object Gray before falling to
12.4--65.1\% on Shape-only. SigLIP2 performs best
(65.1\%), while robust RN50 reaches 12.4\%,
compared with 32.2\% for standard RN50 and
21.3\% for SIN RN50. Neither variant outperforms
standard RN50 on Shape-only; this descriptive
comparison does not isolate training effects.

\paragraph{Composition sharply reduces recognition.}
Final top-1 falls to 0.17--23.08\%; DINOv2 drops
from 55.4\% to 6.1\%, and CLIP from 40.8\%
to 5.9\%. In the matched B1 ladder, placement largely
preserves recognition, while heavy fog on gray lowers
top-1 to 0.0--24.8\% before photographic carrier
content appears (Figure~\ref{fig:recognition_ladder};
Appendix~\ref{app:recognition_ladder}).
DINOv2 falls from 54.2\% after placement to 15.0\%
with fog on gray and 6.5\% on Final matched.
Because fog, contrast, and appearance change together,
their individual effects remain unresolved.
Section~\ref{sec:geometry_or_content} examines
geometry and carrier sensitivity.

\begin{table*}[t]
    \centering
    \small
    \caption{
    \textbf{Incongruent carrier content behind one heavy-fog plate.}
    Native ImageNet-1000 top-1 (\%).
    Each source uses one B1 aperture, one incongruent carrier,
    one fog plate, and a gray surround.
    The phase scramble matches the photograph's mean and
    standard deviation inside the aperture. The constant fill
    uses that mean.
    Paired differences are percentage points, class-macro
    over 600 sources, with 95\% hierarchical intervals.
    Negative differences favor the photograph.
    Bold indicates the highest accuracy in each content column.
    }
    \label{tab:fog_content}
    \setlength{\tabcolsep}{4pt}
    \begin{tabular*}{0.92\textwidth}{@{\extracolsep{\fill}}lccccc@{}}
        \toprule
        Model
        & Photograph
        & Phase scramble
        & Constant
        & \shortstack{Scramble$-$photo\\(pp)}
        & \shortstack{Constant$-$photo\\(pp)} \\
        \midrule
        ResNet-50
        & 3.83 & 1.00 & 0.33
        & $-2.83\;[-5.50,-0.67]$
        & $-3.50\;[-6.50,-1.17]$ \\
        Robust RN50
        & 0.17 & 0.00 & 0.00
        & $-0.17\;[-1.00,0.00]$
        & $-0.17\;[-1.00,0.00]$ \\
        ViT-B/16
        & 6.17 & 2.50 & 4.17
        & $-3.67\;[-7.00,-0.83]$
        & $-2.00\;[-6.00,1.67]$ \\
        ConvNeXt-B
        & 5.67 & 2.83 & 2.33
        & $-2.83\;[-6.17,0.00]$
        & $-3.33\;[-6.50,-0.50]$ \\
        SIN RN50
        & 1.50 & 0.33 & 0.17
        & $-1.17\;[-3.17,0.17]$
        & $-1.33\;[-3.17,-0.17]$ \\
        DINOv2-B/14
        & 15.33 & 14.00 & 25.17
        & $-1.33\;[-5.83,3.17]$
        & $9.83\;[3.67,16.17]$ \\
        CLIP-B/16
        & 5.83 & 5.33 & 16.17
        & $-0.50\;[-4.17,3.17]$
        & $10.33\;[5.83,15.17]$ \\
        SigLIP2-B/16
        & \textbf{31.50} & \textbf{31.50} & \textbf{41.50}
        & $0.00\;[-6.00,5.83]$
        & $10.00\;[4.00,16.17]$ \\
        \bottomrule
    \end{tabular*}
\end{table*}

\paragraph{Carrier content behind a locked fog plate.}
On 600 B1 sources, we compare photographic, phase-scrambled,
and constant carrier content under a shared heavy-fog plate
(Table~\ref{tab:fog_content}). None restores placed-shape
accuracy. Phase scrambling reduces top-1 for ResNet-50
($-2.83$ pp) and ViT-B/16 ($-3.67$ pp), with 95\% CIs
excluding zero; the other six intervals include zero.
Constant fill improves DINOv2, CLIP, and SigLIP2 by
9.83--10.33 pp but reduces ResNet-50 and ConvNeXt-B
accuracy. Carrier effects are therefore model-dependent,
not a uniform content-conflict penalty, and cannot
explain the earlier gray-carrier decline.
Further contrasts appear in
Appendix~\ref{app:recognition_results}.

\paragraph{Candidate restriction yields partial recovery.}
Restricting native scores to the 100 benchmark categories
raises Final top-1 to 1.44--17.81\%, including CLIP
(5.9\% to 17.8\%) and DINOv2 (6.1\% to 15.0\%).
Out-of-benchmark competition contributes to failure
but does not fully explain it. This provides the
matched label-space baseline for
Section~\ref{sec:structural_stability}.

\section{What Controls the Decision: Geometry or Content?}
\label{sec:geometry_or_content}

\begin{figure}[t]
    \centering
    \includegraphics[width=\columnwidth]
    {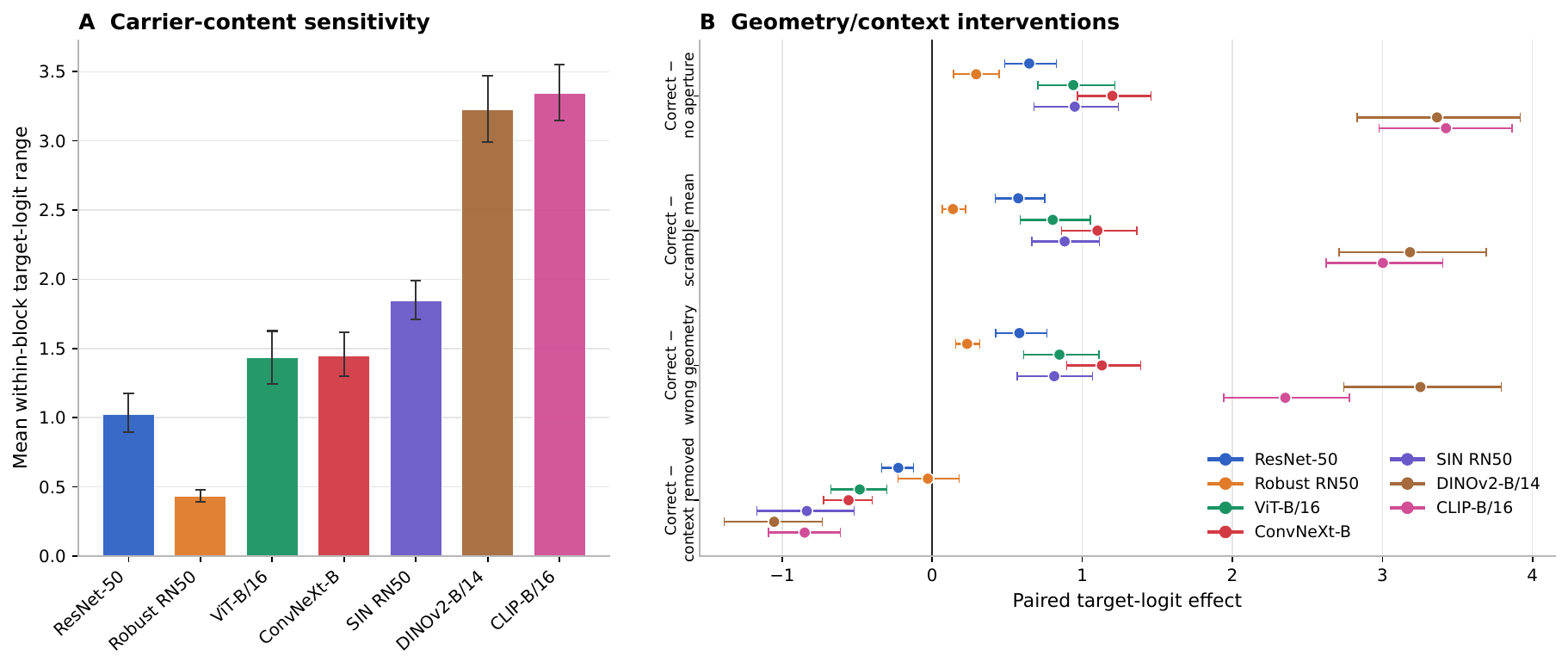}
    \caption{
    \textbf{Carrier dependence and geometry sensitivity.}
    Left: target-score ranges across four carriers in
    3,600 fixed-geometry blocks. Right: paired
    correct-minus-control effects at $B1$ (95\% CIs).
    Context effects use correct-minus-context-removed;
    negative values favor context removal
    (Section~\ref{sec:ablations_robustness}).
    Score scales are checkpoint-specific.
    }
    \label{fig:carrier_geometry}
\end{figure}

Low Final accuracy does not reveal which cues drive
predictions. We vary carriers at fixed geometry and
aperture geometry within fixed mirror--carrier pairs
(Figure~\ref{fig:carrier_geometry};
Table~\ref{tab:geometry_and_carrier}).

\paragraph{Carrier appearance influences predictions.}
Across 3,600 four-carrier blocks per checkpoint,
fixed-geometry top-1 switching ranges from 43.42\%
to 80.50\% (70.25\% for ResNet-50).
Switching measures prediction disagreement, not
successful recognition: robust RN50 switches least
despite 0.17\% Final top-1 accuracy.
Because carrier swaps also alter fog throughout the
glass, we repeat them with a fixed fog plate on 600
B1 sources. Across eight checkpoints, switching
remains 38.5--73.0\%, exceeding fog-seed
(8.3--42.7\%) and JPEG (2.5--38.2\%) switching
for every checkpoint. Image differences concentrate
inside the aperture, although nuisance sensitivity
remains substantial. These results indicate greater
sensitivity to carrier appearance than to the tested
nuisances, not semantic recognition of carrier content
(Appendix~\ref{app:carrier_nuisance_controls}).

\paragraph{Correct geometry increases target evidence.}
We compare correct $B1$ apertures with absent,
scrambled, and wrong-class geometry under fixed
mirror--carrier assignments. For native target score
$s_y(x)$, the paired effect of control $g$ is
\begin{equation}
    \Delta s_y(g)
    =
    s_y(x_{\mathrm{correct}})
    -
    s_y(x_g).
    \label{eq:geometry_effect}
\end{equation}
All three primary contrasts have positive 95\% CIs
and Holm-adjusted $p<0.01$ across seven checkpoints;
all 21 individual scramble-family intervals also
favor correct geometry
(Appendix~\ref{app:geometry_carrier_results}).
For DINOv2, correct geometry improves target rank
by 223.1 positions relative to the scramble mean
(95\% CI: $[188.9,258.7]$).
Native top-1 also improves to 0.17--6.67\% from
0\% for the mean scramble, with positive unadjusted
paired intervals for five of seven checkpoints.
These gains remain small relative to the overall
recognition decline. Same-category replacements
yield intervals covering zero for target logits,
log-probability, and rank across all checkpoints
(Appendix~\ref{app:same_category_geometry}).

\paragraph{Control matching and category shifts.}
Scrambles preserve area exactly but only approximately
match other geometry; tile, row-band, and column-band
QC pass rates are 38.17\%, 67.33\%, and 71.67\%.
All 21 correct-minus-scramble target-logit and rank
intervals remain positive on QC-passing subsets,
with the same direction on failing subsets
(Appendix~\ref{app:individual_scramble_results}).
Post-render audits find small mean-luminance residuals
but remaining contrast and gradient differences;
transmitted aperture area and boundary matching
remain unverified
(Appendix~\ref{app:post_render_scramble_audit}).
Wrong-class replacements reduce original-class
evidence and increase donor-class evidence across
all seven checkpoints, although the robust RN50
category margin includes zero. Donor assignments
are unbalanced (1--13 uses per class), and these
relative score shifts do not establish top-1
donor recognition (Table~\ref{tab:donor_shift}).

\paragraph{Prediction alignment with visible objects.}
Context agreement exceeds carrier agreement for five
of seven checkpoints, but 71.7--99.2\% of predictions
fall into Other. These labels measure semantic
agreement, not the source of model decisions
(Figure~\ref{fig:prediction_alignment};
Appendix~\ref{app:prediction_alignment}).

\section{Cross-View Readout Transfer and Final-View Decodability}
\label{sec:structural_stability}

We compare unchanged Shape-trained probes on held-out
Final images with native Benchmark-100 recognition and
separately fitted Final-trained readouts
(Table~\ref{tab:structural_transfer}).
A distinct equal-source-weight DINOv2 head evaluates
aperture controls (Table~\ref{tab:final_render_probe}).

\begin{table*}[t]
    \centering
    \small
    \caption{
    \textbf{Cross-view readout transfer and within-Final
    category decoding.}
    Source-disjoint 100-class accuracy (\%).
    Arrows indicate training-to-test views.
    Native Benchmark-100 is evaluated on Final;
    paired gaps compare Shape$\to$Final against native
    recognition (pp; 95\% CIs from unrounded estimates).
    Final$\to$Final refits weights and standardization
    on individual Final renders and reuses the
    Shape-selected penalty.
    Bold indicates column maxima.
    }
    \label{tab:structural_transfer}

    \setlength{\tabcolsep}{3pt}
    \renewcommand{\arraystretch}{1.10}

    \begin{tabular*}{0.84\textwidth}{
        @{\extracolsep{\fill}}
        lccccc
        @{}
    }
        \toprule
        \textbf{Model}
        & \textbf{Shape$\to$Shape}
        & \textbf{Shape$\to$Final}
        & \textbf{Native 100}
        & \textbf{Final$\to$Final}
        & \textbf{Paired gap (pp)} \\
        \midrule

        ResNet-50
        & 65.2 & 4.1 & 7.8 & 11.9
        & $-3.65\;[-5.72,-1.78]$ \\

        Robust RN50
        & 57.2 & 1.3 & 1.4 & 1.8
        & $-0.10\;[-1.65,+1.40]$ \\

        ViT-B/16
        & 65.7 & 7.1 & 7.6 & 16.4
        & $-0.47\;[-2.57,+1.48]$ \\

        ConvNeXt-B
        & 78.2 & 11.3 & 12.0 & 25.0
        & $-0.73\;[-2.61,+1.12]$ \\

        SIN RN50
        & 55.7 & 3.6 & 3.6 & 5.9
        & $-0.05\;[-1.66,+1.43]$ \\

        DINOv2-B/14
        & \textbf{83.2} & \textbf{12.2}
        & 15.0 & \textbf{36.4}
        & $-2.84\;[-6.17,+0.49]$ \\

        CLIP-B/16
        & 66.8 & 11.2 & \textbf{17.8} & 25.0
        & $-6.64\;[-9.87,-3.35]$ \\

        \bottomrule
    \end{tabular*}
\end{table*}

\paragraph{Cross-view transfer and native recognition.}
Shape-trained probes achieve 55.67--83.17\% on
held-out Shape-only images but only 1.33--12.21\%
on Final with unchanged weights and standardization.
DINOv2 falls from 83.17\% to 12.21\%; conditional
retention of correct Shape-only predictions is
1.41--14.54\% across checkpoints
(Appendix~\ref{app:probe_details}).
Source-disjoint folds prevent training/test photograph
overlap. On identical Final images and 100 labels,
all transferred-probe-minus-native point estimates
are negative, with paired intervals excluding zero
for ResNet-50 and CLIP
(Table~\ref{tab:structural_transfer}).

\paragraph{Sensitivity of the ridge readout.}
Feature-distribution shift contributes to poor
transfer: recentering DINOv2 Final features raises
the unchanged Shape-trained probe from 12.21\%
to 24.62\%, and scale calibration reaches 25.49\%,
still below its 83.17\% Shape-only accuracy.
The remaining gap reflects limited readout
portability, not necessarily lost geometric information.
A three-view Final fit reduces Final-to-Final
accuracy from 36.35\% to 24.25\%
(Appendix~\ref{app:readout_sensitivity}).

\paragraph{Aperture-category tracking.}
The 36.35\% DINOv2 result uses a render-level ridge
fit with the Shape-selected penalty. A separate
equal-source-weight logistic head achieves 23.64\%
on held-out Final images (95\% CI: $[18.93,28.58]$)
and 25.00\% on correct geometry
($[18.83,31.33]$). Scrambles, no aperture, and the
mirror--carrier identifier score 0--1.33\%.
Correct geometry exceeds the scramble mean by
24.11 pp ($[18.17,30.45]$) and no aperture
by 24.33 pp ($[18.17,31.17]$).
Same-class replacements reach 42.67\%, while
wrong-class replacements yield 0\% target and
53.17\% donor accuracy. These comparisons are
descriptive because replacement donors may originate
from training sources.

In a limited source-and-asset-disjoint evaluation
of 32 sources from 32 classes, Final accuracy is
15.76\% and correct-geometry accuracy is 12.50\%
(95\% CI: $[3.13,25.00]$), exceeding the scramble
mean by 10.42 pp ($[1.04,21.88]$).
The small sample and wide intervals preclude strong
claims about unseen-asset generalization.
Table~\ref{tab:final_render_probe} gives condition-level
results; Appendix~\ref{app:aperture_tracking_details}
details paired contrasts, QC, and donor provenance.

\begin{table}[t]
    \centering
    \small
    \caption{
    \textbf{Render-level DINOv2 readout by aperture condition.}
    Class-macro top-1 (\%). Source-disjoint uses 600 sources from
    100 classes; source-and-asset-disjoint uses 32 from 32 classes.
    Scrambles are tile, row-band, and column-band. Donor is the
    wrong aperture's category.
    This equal-source-weight head is separate from the 36.35\%
    Final$\to$Final fit in Table~\ref{tab:structural_transfer}.
    Asset-disjoint results are exploratory due to the limited
    number of test sources.
    }
    \label{tab:final_render_probe}
    \setlength{\tabcolsep}{3pt}
    \renewcommand{\arraystretch}{1.08}
    \begin{tabular*}{\columnwidth}{@{\extracolsep{\fill}}lcc@{}}
        \toprule
        \textbf{Condition}
        & \textbf{Source-disjoint}
        & \textbf{Asset-disjoint} \\
        \midrule
        Final & 23.64 & 15.76 \\
        Correct geometry & 25.00 & 12.50 \\
        Same-class aperture & 42.67 & 37.50 \\
        Wrong-class target & 0.00 & 0.00 \\
        Donor class & 53.17 & 43.75 \\
        Scrambles & 0.50 / 0.83 / 1.33 & 0.00 / 3.13 / 3.13 \\
        No aperture & 0.67 & 0.00 \\
        Mirror--carrier identifier & 0.00 & 0.00 \\
        \bottomrule
    \end{tabular*}
\end{table}

\section{How Do VLMs Respond to Semantic Apertures?}
\label{sec:vlm_recognition}

We evaluate zero-shot recognition on the 500 Final
images from the human study, five per category.
Five queries per image assess target recognition
(Original, Placed Shape-only, Final) and carrier
recognition (clean carrier, Final). Placed Shape-only
preserves mirror placement, unlike the centered
encoder diagnostic. Target queries use the human
study's ten options; carrier queries use ten categories
(Appendix~\ref{app:extended_vlm}).

\paragraph{Paired evaluation protocol.}
We query \texttt{gpt-6-luna} and
\texttt{gemini-3.8-flash}, both unpinned provider
aliases, 2{,}500 times each. The paired analysis
uses 498 common images spanning all 100 categories,
with scored responses to all five queries from
both endpoints. Invalid outputs count as incorrect;
technical failures remain missing. Accuracy and
paired differences are category-macro, while
conditional retention is image-pooled
(Appendix~\ref{app:extended_vlm}).

\begin{figure}[t]
    \centering
    \includegraphics[width=\columnwidth]
    {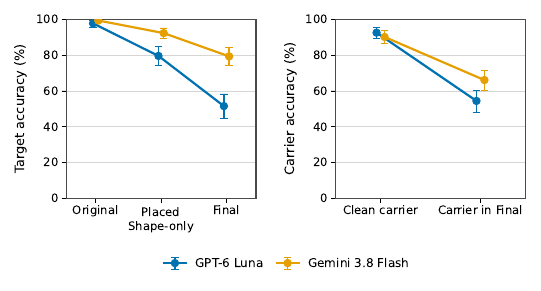}
    \caption{
    \textbf{Target and carrier recognition under composition.}
    Target accuracy across Original, Placed Shape-only,
    and Final (left); carrier accuracy on clean and
    Final views (right). Category-macro estimates
    use 498 common images (95\% hierarchical CIs).
    }
    \label{fig:vlm_paired_recognition}
\end{figure}

\paragraph{Recognition under composition.}
Original target accuracy is 97.8\% for GPT and
99.4\% for Gemini. From Placed Shape-only to Final,
GPT falls from 79.6\% to 51.6\% ($-28.0$ pp;
95\% CI: $[-34.8,-21.5]$), and Gemini from
92.4\% to 79.3\% ($-13.0$ pp; $[-17.8,-8.4]$).
Conditional retention is 59.4\% ($[52.3,66.5]$)
for GPT and 84.3\% ($[79.5,88.7]$) for Gemini.
Carrier accuracy also declines, from 92.6\% to
54.4\% for GPT and 90.2\% to 66.1\% for Gemini
(Figure~\ref{fig:vlm_paired_recognition}).
These transitions jointly change appearance,
fog, and context, rather than isolating semantic
interference (Appendix~\ref{app:extended_vlm}).

\section{Ablations and Robustness Checks}
\label{sec:ablations_robustness}

Final top-1 does not increase monotonically with aperture
support: five of seven checkpoints score higher at B1 than B2.
Only DINOv2's paired top-1 interval excludes zero
($+2.21$ pp, 95\% CI $[1.19,3.42]$; Holm-adjusted
$p=0.00035$). Aperture levels jointly change structural
support, transmitted appearance, and potentially scale.

Removing the surround increases target scores for six
checkpoints (Holm-adjusted $p<0.01$). DINOv2 gains 1.05
score units (95\% CI $[0.73,1.39]$), while context-removed
accuracy remains far below placed shape
(Figure~\ref{fig:recognition_ladder}).
Target-score profiles also vary across mirror shape, frame,
surround and carrier category. Appendices~\ref{app:critical_ablations}
and~\ref{app:factor_heterogeneity} give the full results.

\section{Discussion, Limitations, and Conclusion}
\label{sec:discussion_limitations}

\noindent\textbf{Discussion.}
The matched ladder locates the largest recognition
decline at the heavy-fog gray-carrier stage, before
photographic carrier content appears. Because fog
transmission, aperture contrast, and appearance change
together, their individual effects remain unresolved.
Nevertheless, correct geometry increases target
evidence, carrier appearance affects predictions, and
Final-trained probes recover partial category information.
These findings distinguish recognition failure from
cue sensitivity and category decodability, without
establishing an exclusively geometric mechanism.

\noindent\textbf{Limitations.}
Only one heavy-fog setting is tested, and carrier swaps
also alter fog across the glass. Geometry controls
match area but not all low-level properties; cropping
changes scale, and post-render aperture area and
boundary matching remain unverified.
The 36.35\% DINOv2 Final head was not evaluated on
geometry controls. The separate aperture-tracking head
has a 32-source asset-disjoint test with wide intervals,
and some replacement apertures originate from training
sources. Intervals crossing zero do not establish
equivalence. Finite assets, no physical validation,
and one robust checkpoint limit generalization.
Human judgments are sparse (3--11 per image), respondent
uniqueness is unverified, and no human Shape-only baseline
was collected. Unpinned VLM aliases, missing responses,
and differences in prompts and aggregation limit
human--model comparisons. Separate target and carrier
queries do not establish semantic interference
(Appendix~\ref{app:extended_vlm}).

\noindent\textbf{Conclusion.}
Semantic Apertures reveals substantial recognition
loss under heavy fog alongside measurable geometry
and carrier sensitivity. Shape-trained readouts
transfer poorly, while Final-trained probes recover
partial category evidence. On matched ten-choice
Final images, GPT and Gemini reach 51.6\% and 79.3\%
target accuracy versus 69.9\% for human responses,
without a uniform human--machine gap.


\section*{LLM Usage Statement}
LLMs and a generative image model assisted with the
manuscript, code, and stimulus assets.
Two further models were used only as the
vision-language systems under evaluation.
No model is an author.
The authors reviewed the assisted text, code, and
generated assets and are responsible for this submission.

\paragraph{Manuscript and research.}
ChatGPT, GPT-6, and Claude assisted with editing the
manuscript and with research discussion, including
wording, organization, and clarification of technical
points. They also assisted with providing reviewer-style feedback on the full manuscript or substantial sections of it, including comments on wording, organization, clarity, and technical presentation.
These tools did not produce the experimental
measurements or the reported numerical results.

\paragraph{Code.}
Grok assisted with drafting and revising analysis
and experiment code.
The authors reviewed and executed that code before
relying on it.

\paragraph{Stimulus assets.}
Gemini 3.1 Flash Image
(\texttt{gemini-3.1-flash-image}) generated the fixed
banks of 45 mirror/context images and 30 carrier
images from text-only prompts, with no input image
and no conversation history.
The requested identifier is a provider-hosted alias
rather than a pinned checkpoint.
The saved banks are reused for compositing and
recognition and are not regenerated at evaluation
time.
Prompt text, factor design, and generation records
are reported in
Appendix~\ref{app:asset_generation_prompts}.

\paragraph{Evaluation endpoints.}
GPT-6 Luna (\texttt{gpt-6-luna}) and Gemini 3.8 Flash
(\texttt{gemini-3.8-flash}) were queried as the
vision-language models under evaluation.
Each call was an independent zero-shot request
containing one image and a task-specific prompt.
GPT requests used low reasoning effort; Gemini
requests used low thinking and temperature zero.
These endpoints were not used to draft the paper or
to select reported outcomes.
Both identifiers are provider-hosted aliases rather
than pinned checkpoints.
Coverage, prompts, and scoring are reported in
Appendix~\ref{app:extended_vlm}.



{
\small
\bibliographystyle{ieeenat_fullname}
\bibliography{main}
}

\clearpage

\appendix

\section*{Appendix Part I: Supporting Methods and Results}

\section{Benchmark Construction and Provenance}
\label{app:benchmark_construction}

This appendix details the benchmark construction
described in Section~\ref{sec:semantic_apertures},
including source selection, structural aperture extraction,
asset generation, image compositing, assignment balance,
and reproducibility. The source photograph is the
independent sampling unit; its derived apertures and
composite renders constitute repeated observations.

\subsection{Source Categories, Masks, and Selection}
\label{app:source_selection_details}

The benchmark includes 100 ImageNet categories,
comprising animals (45), vehicles (15), devices (20),
household objects (15), and food (5). Six distinct
ImageNet validation photographs are selected per
category, yielding 600 sources. Category identities
are represented by ImageNet synsets, with associated
labels and segmentation identifiers retained in the
source manifest.

\paragraph{Segmentation processing.}
Object masks are obtained from ImageNet-S annotations~\citep{gao2022luss}.
For RGB-encoded annotations, the segmentation identifier
is decoded as the red-channel value plus 256 times
the green-channel value. Integer-valued annotations
are used directly. The binary mask retains pixels
associated with the source category's segmentation
identifier. Images with empty masks, invalid identifiers,
or inconsistent image and mask dimensions are excluded.

\paragraph{Geometry-based source selection.}
Source selection prioritizes segmentation geometry
rather than recognition confidence.

Let $f_{\mathrm{mask}}$ denote the foreground fraction,
$f_{\mathrm{box}}$ the fraction occupied by its tight
bounding box, $n_{\mathrm{comp}}$ the number of
eight-connected foreground components, and $f_{\max}$
the fraction of foreground pixels in the largest
component. Let $n_{\mathrm{border}}$ count the image
borders contacted by the foreground. Candidate
quality is defined as

\begin{equation}
\begin{aligned}
q ={}& f_{\max}
      -0.06\max(0,n_{\mathrm{comp}}-1)\\
    &-0.04n_{\mathrm{border}}\\
    &+0.12\mathbf{1}
      [0.08\leq f_{\mathrm{mask}}\leq0.55]\\
    &+0.35\mathbf{1}[n_{\mathrm{comp}}=1].
\end{aligned}
\label{eq:source_selection_score}
\end{equation}

Preferred candidates satisfy
$f_{\mathrm{mask}}\in[0.02,0.85]$ and
$f_{\mathrm{box}}\in[0.035,0.85]$.
These candidates are organized into three
priority tiers: single-component masks contacting
at most two image borders; masks containing at
most two components with $f_{\max}\geq0.85$;
and all remaining candidates satisfying the
size constraints.

Within each tier, candidates are ranked by
descending quality score, rounded to five
decimal places, with filenames used to resolve ties.
When fewer than six preferred sources are available,
nonempty masks outside the preferred size ranges
are considered in descending priority. These
selections are identified as relaxed-filter cases.

Each category must contain six distinct usable
photographs. Source identifiers and within-category
image hashes are checked for duplication.
The selection procedure does not use predictions,
confidence scores, or features from the evaluated
recognition models.

After selection, image and mask pairs whose
longer dimension exceeds 1,200 pixels are resized
while preserving aspect ratio. Images use Lanczos
interpolation, while segmentation masks use
nearest-neighbor interpolation. Candidate ranking
is performed before resizing.

\subsection{Nested Aperture Extraction}
\label{app:aperture_details}

Structural apertures are extracted within the
source-object support $M$.
For a field $z$, masked percentile normalization
is defined as

\begin{equation}
\mathcal{N}_{a,b}(z)
=
\operatorname{clip}
\left(
\frac{z-P_a(z\mid M)}
     {\max(P_b(z\mid M)-P_a(z\mid M),10^{-6})},
0,1
\right),
\label{eq:masked_normalization}
\end{equation}

where $P_a(z\mid M)$ denotes percentile $a$
over foreground pixels.

\paragraph{B0: baseline source structure.}
We extract CIELAB lightness and apply bilateral
filtering with diameter 9, color parameter 0.10,
and spatial parameter 9, followed by CLAHE
with clip limit 1.8 and an $8\times8$ tile grid.

Let $L$ denote the resulting lightness field.
A Gaussian illumination estimate $I$ is computed
using standard deviation
$\max(8,\min(H,W)/25)$.
The illumination-adjusted detail field is

\begin{equation}
R_{\mathrm{detail}}
=
\mathcal{N}_{2,98}
\left(
\log(L+0.035)-\log(I+0.035)
\right).
\end{equation}

Darkness combines normalized lightness and
illumination-adjusted detail:

\begin{equation}
D
=
0.56\left(1-\mathcal{N}_{2,98}(L)\right)
+
0.44\left(1-R_{\mathrm{detail}}\right).
\label{eq:aperture_darkness}
\end{equation}

Edge strength is computed from the magnitude
of $3\times3$ Sobel derivatives of $L$ and
normalized using foreground percentiles
4 and 99.

The initial B0 aperture combines pixels above
the foreground 0.76 quantile of darkness,
pixels at or above the 93rd percentile of edge
strength, and an interior silhouette rim.

The rim is obtained by subtracting an eroded
mask from $M$, using radius

\[
r=\max\left(
1,\operatorname{round}(0.004\min(H,W))
\right)
\]

and a square erosion kernel of size
$(2r+1)\times(2r+1)$.

Morphological cleanup removes connected components
smaller than

\[
\max\left(
4,
\left\lfloor
\frac{\max(6,\lfloor0.000055HW\rfloor)}{2}
\right\rfloor
\right)
\]

pixels. Components of the complement within
$M$ containing at most
$\max(1,\lfloor0.00008|M|\rfloor)$ pixels
are filled.

\paragraph{B1: expanded source structure.}
B1 applies the same extraction procedure
with relaxed darkness and edge thresholds:
darkness quantile 0.68 and edge percentile 88.

The resulting support is unioned with B0,
preserving all pixels retained at the
baseline level.

\paragraph{B2: additional support.}
B2 preserves B1 and introduces additional
structural support when foreground coverage
is below 0.55.

The expansion priority combines normalized
darkness, edge magnitude, and local contrast
with weights 0.50, 0.30, and 0.20.

For this calculation, lightness is processed
using CLAHE with clip limit 2.0 and an
$8\times8$ tile grid. Local contrast is
the absolute difference between lightness
and its Gaussian-smoothed version with
standard deviation 2.0.

Each component is normalized using
$\mathcal{N}_{2,98}$, and the combined
priority field is smoothed with standard
deviation 1.15.

Pixels outside $M$ and those already retained
by B1 are excluded from expansion. The number
of additional pixels is

\begin{equation}
n_{\mathrm{add}}
=
\max\left(
0,
\operatorname{round}
\left(0.55|M|-|A^{B1}|\right)
\right),
\end{equation}

subject to the available candidate count.

If B1 already reaches the coverage target,
B2 remains identical to B1. The value 0.55
therefore represents an expansion target,
not an upper bound on aperture coverage.

The three binary supports satisfy the nesting
relation in Equation~\ref{eq:nested_ink}
in source coordinates. Nesting is verified
before independent aperture placement.

\paragraph{Interpretation of aperture levels.}
Figure~\ref{fig:semantic_apertures_overview}
illustrates the extraction process using an
acoustic-guitar source and its segmentation mask.
The three aperture levels represent progressively
expanded source-coordinate support rather than
fixed percentages of object-mask coverage.

The final B0, B1, and B2 apertures are defined
by the extraction procedures above, not by
the fixed coverage fractions selected during
historical calibration
(Appendix~\ref{app:pilot_calibration}).

Although nesting holds before composition,
aperture placement and resizing are performed
independently for each level. Consequently,
the rendered apertures may differ in scale
and need not preserve pixelwise nesting
after registration.

\subsection{Asset Generation and Prompt Provenance}
\label{app:asset_generation_prompts}

The benchmark uses two fixed collections of
synthetic visual assets: a bank of 45
mirror/context images and a bank of 30
carrier images.

These assets provide controlled variation in
mirror geometry, surrounding context, frame
appearance, and independently assigned visual
content. They are generated separately from
the source photographs and structural apertures.

\paragraph{Generation procedure.}
The asset-generation procedure specifies the
provider-hosted \texttt{gemini-3.1-flash-image}
alias through AvalAI.

Each request consists of a single user message
containing a text prompt, without an input
image or conversation history.
Both generation interfaces request square
images using \texttt{aspectRatio=1:1}.

The image-size configuration supports
\texttt{1K}, \texttt{2K}, and \texttt{4K},
with \texttt{1K} as the default.
Neither generation procedure explicitly
sets a random seed or sampling temperature.

The requested model identifier is a
provider-hosted alias rather than a
fixed checkpoint revision.

The first returned image is converted to RGB
and saved in PNG format after verification
of its square dimensions.

Generation records include the intended asset
filename, prompt text and hash, requested
model configuration, generation timestamp,
returned image dimensions, and image hashes.

The final benchmark uses the fixed generated
assets rather than regenerating images
during recognition evaluation.

\paragraph{Mirror asset bank.}
The 45 mirror/context images cross five
mirror geometries, three surrounding conditions,
and three frame styles.

The geometries are circle, square, upright
pentagon, upright triangle, and an irregular
organic shape identified as \texttt{deformed}.
The surrounding conditions are
\texttt{none}, \texttt{texture}, and
\texttt{object}; the frame styles are
\texttt{frameless}, \texttt{regular},
and \texttt{ornate}.

Mirror-generation instructions emphasize
a dominant, approximately frontal mirror
with its complete boundary visible and
sufficient unobstructed glass area for
subsequent compositing.

Surrounding elements are constrained to
preserve visibility of the mirror opening
and frame. The requested reflected
appearance is neutral and visually
unobtrusive rather than a recognizable
reflected scene.

Fog, condensation, droplets, and aperture
marks are excluded from the base mirror
descriptions because these effects are
introduced by the renderer.

Each mirror-generation request combines
an asset-specific description with shared
generation instructions.

\begin{figure*}[p]
    \centering
    \includegraphics[
        width=\textwidth,
        height=0.80\textheight,
        keepaspectratio
    ]{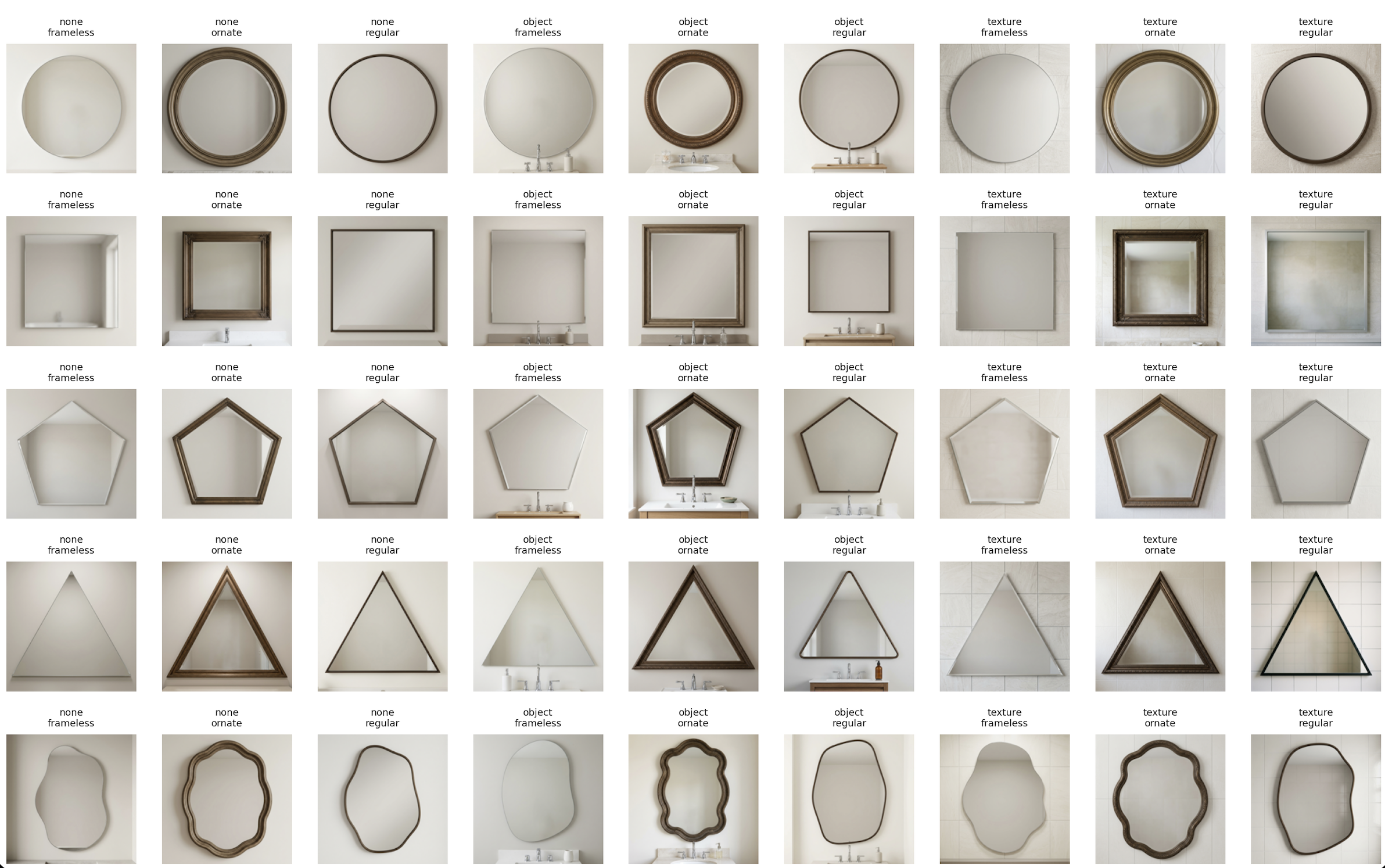}
    \caption{
    \textbf{Complete mirror/context asset bank.}
    The 45 assets cover five mirror shapes,
    three surrounding conditions, and three
    frame styles.
    Rows show circular, square, pentagonal,
    triangular, and irregular organic mirrors,
    respectively.
    Column groups show the surrounding
    conditions \texttt{none}, \texttt{object},
    and \texttt{texture}; within each group,
    frame styles appear in the order
    \texttt{frameless}, \texttt{ornate},
    and \texttt{regular}.
    These base assets provide the mirror opening
    and surrounding context for subsequent
    aperture placement and compositing.
    }
    \label{fig:mirror_asset_bank}
\end{figure*}

Figure~\ref{fig:mirror_asset_bank}
displays the complete 45-image mirror/context
bank and its factor organization.

\paragraph{Mirror-generation prompt example.}
The following is the original asset-specific
opening paragraph for
\texttt{mirror\_circle\_none\_frameless.png}.
It is an excerpt of the generation request,
rather than its complete prompt.

\begin{PaperPrompt}
Create a square, highly photorealistic close frontal photograph of a **large perfectly circular frameless wall mirror**. Make the complete circle as large as possible while keeping its boundary visible, occupying roughly 75-84% of the image width and height. Mount it on a smooth warm off-white or light beige bathroom wall. Keep the surrounding area exceptionally minimal, with only subtle wall material and a very narrow hint of vanity surface at the bottom if needed for realism. No decorative objects around the mirror. The polished mirror edge should be thin and clean, with no visible frame. Follow all shared benchmark requirements above.
\end{PaperPrompt}

\paragraph{Mirror prompt composition.}
The mirror-generation requests include
asset-specific descriptions together with
shared benchmark instructions.
The request structure also includes a
catalogue of mirror specifications.

The example above reproduces only the
asset-specific opening paragraph.
In particular, its reference to shared
benchmark requirements refers to additional
instructions within the generation request.

Consequently, this excerpt should not be
interpreted as the complete serialized
prompt or as sufficient by itself to
recreate the original generation conditions.

\paragraph{Carrier asset bank.}
The carrier bank contains 30 generated images,
organized into ten categories with three
exemplars each:

Bathroom, Kitchen, Bicycle/garage,
Animals/recognizable objects,
Chairs/furniture, Textiles/materials,
Organized repetition, Dense mixed clutter,
Vegetation, and Architectural geometry.

Carrier-generation prompts request
photorealistic scene images containing
recognizable objects, surfaces, and spatial
relationships, rather than images already
depicting mirror reflections.

The instructions emphasize recognizable
visual structure across the central region,
plausible perspective and depth, natural
lighting, and sufficient detail to remain
recognizable after resizing.

The mirror composition, aperture, and fog
are introduced subsequently by the renderer.

The nine Bathroom, Kitchen, and
Bicycle/garage prompts use one shared
instruction template, while the remaining
21 prompts use a second template.
Each request includes an asset-specific
scene-description paragraph.

\begin{figure*}[p]
    \centering
    \includegraphics[
        width=\textwidth,
        height=0.80\textheight,
        keepaspectratio
    ]{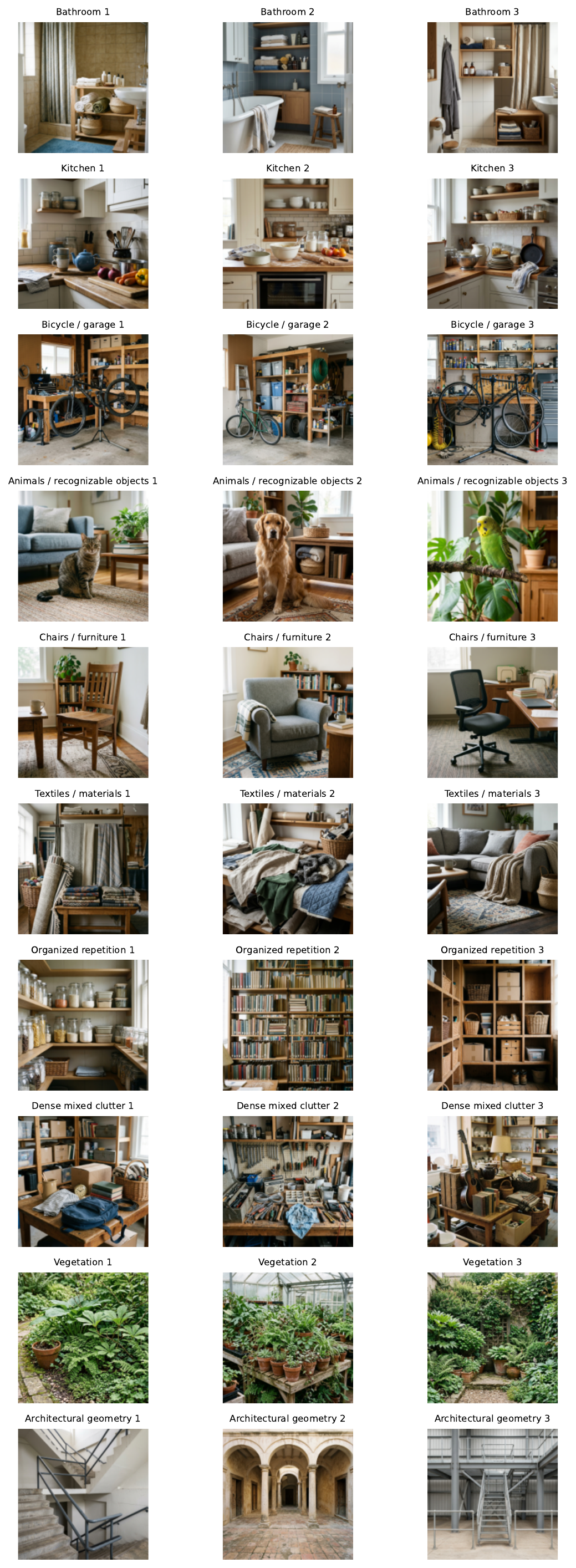}
    \caption{
    \textbf{Complete carrier asset bank.}
    Each row contains three exemplars from
    one carrier category.
    From top to bottom, the categories are:
    Bathroom,
    Kitchen,
    Bicycle/garage,
    Animals/recognizable objects,
    Chairs/furniture,
    Textiles/materials,
    Organized repetition,
    Dense mixed clutter,
    Vegetation,
    and Architectural geometry.
    The images are shown before glass-related
    preprocessing, registration, aperture
    transmission, and fog compositing.
    Carrier appearance is assigned separately
    from the source-object category.
    }
    \label{fig:carrier_asset_bank}
\end{figure*}

Figure~\ref{fig:carrier_asset_bank}
displays the complete carrier bank before
glass-related processing.

\paragraph{Carrier-generation prompt: first template.}
The following opening and closing paragraphs
are used for the nine Bathroom, Kitchen, and
Bicycle/garage assets.

The bracketed scene-description marker
indicates the position of the corresponding
asset-specific paragraph; it is not literal
text included in the generation request.

\begin{PaperPrompt}
Create one square, highly photorealistic color photograph of the detailed scene described below. Show the scene directly, as a normal camera photograph. It is a coherent real-world environment containing several recognizable objects, not an isolated object on a blank backdrop.

[ASSET-SPECIFIC SCENE PARAGRAPH]

Use a level camera at a natural viewing position, a moderate field of view, straight architectural lines, and believable perspective without fisheye distortion. Distribute several medium-size recognizable objects and textured surfaces across the central two-thirds of the image, with foreground, middle-distance, and background structure. Avoid both large empty central areas and tiny distant clutter that turns into noise. Allow natural partial overlaps while preserving enough visible structure to identify several objects. Keep key scene-defining objects inside the frame; peripheral furniture may continue naturally beyond its edges. Use soft diffuse daylight, neutral white balance, moderate contrast, realistic material detail, and enough depth of field for objects at several depths to be recognizable. The room should feel naturally used and visually rich, without surreal piles or exaggerated disorder. Include no people, faces, human silhouettes, mirrors, reflected duplicates, fog, steam, condensation, water droplets on the camera, drawings on glass, readable text, logos, watermarks, borders, collages, dramatic cinematic lighting, extreme bokeh, or illustration. No unrelated objects fused together or physically impossible structures.
\end{PaperPrompt}

\paragraph{Asset-specific carrier description.}
For example, the following is the original
scene-description paragraph for
\texttt{scene\_bathroom\_01.png}:

\begin{PaperPrompt}
A believable used but well-maintained bathroom with warm beige tiles. In the central portion, show an open wooden shelving unit with folded and loosely rolled towels, several unbranded shampoo and lotion containers of different ordinary shapes, a ceramic soap dish, and a woven storage basket. A partial white basin with a metal faucet appears toward one side, and a fabric shower curtain with subtle folds occupies part of the rear wall. A small floor mat adds another surface. The towels, containers, basket, and tiled boundaries form distinct overlapping groups rather than a symmetrical display. Keep the room dry and naturally detailed. There is no mirror anywhere in the scene.
\end{PaperPrompt}

\paragraph{Carrier-generation prompt: second template.}
The remaining 21 carrier assets use the
following opening and closing paragraphs,
with an asset-specific scene description
inserted between them.

\begin{PaperPrompt}
Create one square, highly photorealistic color photograph of the scene described below. Show a believable environment directly, as a normal camera photograph. The image is intended to contain meaningful visual structure across its central region, not an isolated product on an empty backdrop.

[ASSET-SPECIFIC SCENE PARAGRAPH]

Use a level camera, natural perspective, and a moderate field of view without fisheye distortion. Put the important objects and visual structures across the central two-thirds of the frame, rather than only at the edges. Include enough mid-size detail to remain recognizable when the image is reduced, with plausible depth, natural contact shadows, and clear physical relationships. Keep central subjects complete wherever their identity depends on their outline; peripheral scenery may continue beyond the frame. Use soft diffuse daylight, neutral white balance, moderate contrast, realistic materials, and sufficient depth of field for the central content to remain clear. Do not add people, faces, human silhouettes, mirrors, distinct reflected duplicates, fog, steam, condensation, camera droplets, drawings on glass, readable text, logos, watermarks, borders, collages, glowing screens, dramatic cinematic lighting, extreme bokeh, or illustration. Avoid malformed or fused objects, impossible supports, and repeated artificial copies.
\end{PaperPrompt}

\paragraph{Prompt and asset reproducibility.}
The prompt excerpts and templates above
document the generation instructions for
the illustrated assets and their respective
request families.

Complete request reconstruction additionally
requires the asset-specific descriptions,
shared instructions, and any additional
content present in the original serialized
generation requests.

The fixed mirror and carrier images, together
with their associated generation records,
constitute the rendering inputs used by
the benchmark.

Prompt text describes intended visual
properties but does not guarantee that
every generated image satisfies each
instruction. Furthermore, reissuing an
identical prompt through a provider-hosted
model alias does not guarantee pixel-identical
output.

Exact reproduction of the benchmark
composites therefore depends on the
frozen asset images and corresponding
rendering configurations, rather than
regeneration from the example prompts alone.

\paragraph{Construction illustration.}
Figure~\ref{fig:semantic_apertures_overview}
summarizes the construction process,
from a source photograph and segmentation
mask to structural aperture extraction,
mirror placement, carrier registration,
and Final-image compositing.

The illustrated variations show changes
in aperture support, mirror/context
appearance, and carrier content.
They illustrate the composition procedure
rather than measured recognition outcomes.

The transmission diagram summarizes the
spatially varying visibility of carrier
appearance. The implemented renderer
additionally includes fog-field generation,
carrier-dependent scattering, aperture-edge
treatment, glass-mask blending, and image
encoding, as detailed in
Appendix~\ref{app:rendering_details}.

\subsection{Fog and Image Compositing}
\label{app:rendering_details}

The frozen benchmark uses a single
heavy-fog configuration without droplets.

Fog fields are generated deterministically
for each source--mirror pair, using a seed
derived from the global seed, source
identifier, and mirror filename.
The resulting fields are reused across
assigned carriers and aperture levels.

\paragraph{Fog generation.}
The no-droplet opacity field is

\begin{equation}
\label{eq:fog_opacity}
\begin{aligned}
\alpha = Q\,\operatorname{clip}\bigl(
    &0.97+0.15(D_f-0.5)\\
    &+0.105(C_{\mathrm{large}}-0.5)\\
    &+0.055(C_{\mathrm{mid}}-0.5)\\
    &+0.018(G_f-0.5)+0.022V_f,\\
    &0.93,\,0.995\bigr).
\end{aligned}
\end{equation}

Here, $Q$ is the glass-region mask;
$D_f$, $C_{\mathrm{large}}$,
$C_{\mathrm{mid}}$, and $G_f$
denote density, coarse-cloud, medium-cloud,
and grain fields, respectively.
The vertical field $V_f$ ranges from
$-0.5$ to $0.5$.

The renderer supports reference-derived
fog maps when a reference asset is provided
and deterministic procedural fog generation
otherwise. The corresponding field-generation
inputs are part of the rendering configuration.

\paragraph{Fog appearance.}
Let $R_{\mathrm{reg}}$ denote the registered
carrier image.

The scattering color combines 95\% fog color
with 5\% blurred carrier appearance.
The carrier blur uses Gaussian standard
deviation

\[
\sigma_{\mathrm{carrier}}
=
\max(10,\max(H,W)/26).
\]

The fog plate combines scattering color
and registered carrier appearance according
to the opacity field, followed by bloom
and reference-shadow/highlight adjustments.

Because the fog plate partly depends on
the carrier image, changing the carrier
can alter appearance throughout the glass
region, including regions outside the
aperture. Consequently, carrier interventions
do not isolate semantic content visible
exclusively through the aperture.

\paragraph{Aperture placement.}
Each aperture is cropped to its support and centered within
a safe glass region. A 14-step scale search limits clipping
to 0.5\%; masks undergo nearest-neighbor resizing and
mild softening. Placement is computed separately per level,
so source-coordinate nesting does not ensure identical
rendered scales.

\paragraph{Aperture transmission and blending.}
Final compositing is performed in linear RGB.

Registered carrier appearance is mixed
with 14\% of its Gaussian-blurred version,
using standard deviation 1.05.

The placed aperture is softened with
Gaussian standard deviation 0.48.
The transmission field follows
Equation~\ref{eq:transmission}, with

\[
t_{\mathrm{bg}}=0.006,
\qquad
t_{\mathrm{clear}}=0.52.
\]

The renderer blends transmitted carrier
appearance with the fog plate according
to the spatially varying transmission field.

A mild aperture-edge treatment applies
inner-boundary darkening with coefficient
0.026 and outer-boundary highlighting
with coefficient 0.018.

The glass boundary is feathered using
Gaussian standard deviation 2.5,
with blending restricted to the
glass-region mask $Q$.

After conversion to sRGB, pixels outside
$Q$ are restored from the mirror/context
image. Final composites are encoded as
JPEG images at quality 95.

Restoration is performed before JPEG
encoding; compression can introduce
small differences in decoded pixels
outside the glass mask.

\subsection{Global Assignment and Balance}
\label{app:assembly_qc_details}

Asset assignments are generated globally
using seed 42 before partitioning sources
into computational shards.

Each source receives two mirrors with
different shapes and four carriers from
distinct carrier categories.

The same four carriers are reused across
both mirrors and all three aperture levels,
producing

\[
2\times4\times3=24
\]

Final composites per source.

Across 600 sources, this gives 14,400
Final composites.

\paragraph{Balanced assignments.}
Mirror assignment balances shape pairs
and individual asset usage.
Surrounding conditions and frame styles
are subsequently balanced across the
assigned mirrors.

Carrier categories are assigned in groups
of four distinct categories per source,
with global adjustments providing equal
category usage. Individual carrier
exemplars are distributed across their
category assignments.

Table~\ref{tab:benchmark_balance}
summarizes the resulting global marginals
at the assignment and rendering levels.

These balances apply globally; they do
not imply equal representation of every
factor combination within each target
category.

\begin{table*}[t]
    \centering
    \small
    \caption{
    \textbf{Global construction balance.}
    Assignment counts refer to distinct
    source--mirror or source--carrier pairings
    before crossing the remaining factors.
    Final-render counts include repeated
    observations of the same source photographs.
    }
    \label{tab:benchmark_balance}

    \begin{tabular}{@{}lrrr@{}}
        \toprule
        Factor
        & Levels
        & Assignments per level
        & Final renders per level \\
        \midrule
        Mirror shape
            & 5  & 240       & 2,880 \\
        Surrounding type
            & 3  & 400       & 4,800 \\
        Frame style
            & 3  & 400       & 4,800 \\
        Individual mirror asset
            & 45 & 26 or 27  & 312 or 324 \\
        Carrier category
            & 10 & 240       & 1,440 \\
        Individual carrier asset
            & 30 & 80        & 480 \\
        \bottomrule
    \end{tabular}
\end{table*}

Sharding partitions sources after assignment,
preserving their globally determined mirror
and carrier configurations.
Different shards do not introduce distinct
experimental conditions or independently
sampled asset banks.

\subsection{Quality Control and Reproducibility}
\label{app:construction_provenance}

\paragraph{Construction validation.}
The construction pipeline validates source
counts and uniqueness, segmentation-mask
integrity, source-coordinate aperture nesting,
support confinement, placement clipping,
and completeness of the expected renders.

Additional checks address global assignment
balance, unique sample identifiers, image
availability, and consistency between
render metadata and the corresponding
source and asset assignments.

The renderer also verifies restoration
outside the glass region before lossy
image encoding.

\paragraph{Data and rendering provenance.}
The source manifest associates each
photograph with its segmentation mask,
selection properties, and image identifiers.

Assignment metadata identifies the
selected mirror and carrier assets,
while render metadata associates each
composite with its source photograph,
aperture level, rendering condition,
placement measurements, and output identity.

The rendering configuration retains the
global seed, asset assignments,
fog-generation inputs, output dimensions,
and encoding parameters.

Together, these records support source-level
grouping, consistency checks, and reproduction
using the frozen inputs.

\paragraph{Dataset accounting.}
The four primary recognition views contain
600 Original photographs,
600 Object Gray images,
1,800 Shape-only images, and
14,400 Final composites.

The geometry and context intervention set
contains 4,800 condition records,
including 600 correct-reference images
already present among the Final composites.

These derived images and intervention records
must not be treated as independent source
photographs. Recognition estimates therefore
aggregate repeated observations at the
source level before category-level analysis.

\paragraph{Scope of qualitative evidence.}
The construction examples illustrate how
target-derived spatial structure is extracted
and combined with separately assigned
visual content.

They do not constitute a quantitative
evaluation of recognition success or failure.
The recognition, geometry-intervention,
carrier-sensitivity, and structural-transfer
findings are established by the corresponding
quantitative evaluations in
Sections~\ref{sec:recognizing_structure}--%
\ref{sec:structural_stability}.

Similarly, the variation shown across
the 45 mirror assets and 30 carrier assets
characterizes the fixed rendering banks.
It does not establish generalization to
unseen asset collections or photographs
of physical fogged mirrors.

\paragraph{Reproduction scope.}
Reproducing the exact benchmark requires
the selected source photographs and masks,
fixed mirror and carrier assets, assignment
metadata, fog-generation inputs, and
rendering implementation.

The benchmark uses controlled synthetic
compositing rather than a physically
calibrated simulation of reflection
optics or condensation.

Its construction does not model full
three-dimensional reflection geometry,
physical scene depth, or measured
condensation dynamics.

\subsection{Relationship to Prior Work}
\label{app:related_work_comparison}
Table~\ref{tab:related_work_comparison} summarizes the diagnostics
compared in Section~\ref{sec:related_work}.

\begin{table}[t]
\centering
\caption{Relationship to prior work. Semantic Apertures
provides complementary diagnostic evaluations, not
head-to-head evidence of superiority.}
\label{tab:related_work_comparison}
\footnotesize
\setlength{\tabcolsep}{3pt}
\renewcommand{\arraystretch}{1.12}
\begin{tabular}{@{}p{0.32\columnwidth}p{0.61\columnwidth}@{}}
\hline
\textbf{Prior work} & \textbf{Complementary evaluation in our work} \\
\hline
Baker; Geirhos
& Target-derived apertures exposing unrelated scene
appearance rather than target-object texture. \\
Xiao; Rosenfeld
& Paired geometry interventions and carrier-assignment
sensitivity within a shared rendering pipeline. \\
ImageNet-C / R / Sketch
& Recognition across matched construction stages,
including the heavy-fog/contrast transition. \\
IllusionBench
& Explicit aperture interventions, matched construction
ladder, and cross-view readout diagnostics. \\
Representation probing
& Unchanged Shape$\rightarrow$Final transfer versus
separately fitted Final readouts and native recognition. \\
\hline
\end{tabular}
\end{table}


\section{Models, Readouts, and Statistical Details}
\label{app:evaluation_details}

This appendix expands Section~\ref{sec:experimental_setup}.
All eight visual encoders remain frozen. Native predictions
are computed for every checkpoint. Intervention effects and
fitted-readout evaluations use the seven checkpoints with
saved features; SigLIP2-B/16 contributes prediction rows only.

\subsection{Checkpoint Identities}
\label{app:checkpoint_details}

Table~\ref{tab:checkpoint_identifiers} lists the checkpoint
identifiers specified by the evaluation runner. DINOv2 uses its
supplied ImageNet linear classifier for native evaluation.
CLIP and SigLIP2 use an ImageNet text-template ensemble.
None of these readouts is fitted on Semantic Apertures.

\begin{table*}[t]
    \centering
    \small
    \caption{
    Checkpoint identifiers specified by the evaluation code.
    Available checkpoint hashes and preprocessing metadata are
    stored with the corresponding evaluation outputs.
    }
    \label{tab:checkpoint_identifiers}
    \begin{tabular}{@{}l p{0.70\textwidth}@{}}
        \toprule
        Model & Checkpoint specification \\
        \midrule
        ResNet-50 &
        \texttt{ResNet50\_Weights.IMAGENET1K\_V2} \\

        Robust RN50 &
        \texttt{madrylab/resnet50\_linf\_eps4.0};
        adversarial training with $\ell_\infty$ radius $4/255$~\citep{salman2020transfer} \\

        ViT-B/16 &
        TorchVision
        \texttt{ViT\_B\_16\_Weights.IMAGENET1K\_V1} \\

        ConvNeXt-B &
        TorchVision
        \texttt{ConvNeXt\_Base\_Weights.IMAGENET1K\_V1} \\

        SIN RN50 &
        Geirhos Stylized-ImageNet checkpoint,
        \nolinkurl{resnet50_train_60_epochs-c8e5653e.pth.tar} \\

        DINOv2-B/14 &
        Torch Hub \texttt{facebookresearch/dinov2:main},
        entry point \texttt{dinov2\_vitb14\_lc} \\

        CLIP-B/16 &
        OpenCLIP~\citep{schuhmann2022laionb, radford2021clip, cherti2023reproducible} architecture \texttt{ViT-B-16},
        pretrained weights \texttt{openai} \\

        SigLIP2-B/16 &
        OpenCLIP \texttt{hf-hub:timm/ViT-B-16-SigLIP2}~\citep{tschannen2025siglip2};
        80 OpenAI ImageNet templates;
        $\operatorname{sigmoid}(\mathrm{scale}\cdot\mathrm{similarity}+\mathrm{bias})$ \\
        \bottomrule
    \end{tabular}
\end{table*}

The checkpoint collection supports comparisons among the evaluated
systems. Architecture, pretraining data, training objective, and
native readout differ across systems, so these comparisons do not
isolate the causal effect of a training regime.

\subsection{Image Preprocessing and Text Readouts}
\label{app:preprocessing_details}

ResNet-50, ViT-B/16, and ConvNeXt-B use the preprocessing transforms
attached to their specified TorchVision weights.
Robust RN50, SIN RN50, and DINOv2-B/14 use bicubic resizing of the
shorter dimension to 256 pixels, with antialiasing, followed by a
$224\times224$ center crop. RGB channels are normalized with means
$(0.485,0.456,0.406)$ and standard deviations
$(0.229,0.224,0.225)$.
CLIP and SigLIP2 use the preprocessing returned by OpenCLIP for
their checkpoints.

For CLIP, each class name is inserted into the active template
ensemble. Individual text embeddings are normalized, averaged
within class, and normalized again. Image embeddings are normalized,
and class scores are their similarities to the resulting text
embeddings, multiplied by the checkpoint's learned logit scale.

SigLIP2 inserts each class name into the 80-template OpenAI
ensemble recorded in its metadata. Its class score is the sigmoid
of the scaled similarity plus the checkpoint's learned logit bias.

For CLIP, the runner attempts to load OpenCLIP's ImageNet class names and
OpenAI ImageNet templates. Its fallback uses class names from the
ImageNet index and four templates:
\emph{a photo of a \{class\}},
\emph{a blurry photo of a \{class\}},
\emph{a close-up photo of a \{class\}}, and
\emph{a photo of the \{class\}}.
The active ensemble size is recorded in model metadata.
Reproduction should use that recorded configuration rather than
assume a template count from the checkpoint name.

\subsection{Native Scores and Answer Spaces}
\label{app:native_readout_details}

Let $z_j(x)$ be the saved native score for ImageNet category $j$.
The native ImageNet-1000 prediction is
\begin{equation}
\widehat{y}_{1000}(x)
=
\arg\max_{j\in\{1,\ldots,1000\}}z_j(x).
\end{equation}
Top-1 and top-5 accuracy measure whether the source target appears
in the corresponding predicted set.

Let $\mathcal{Y}_{100}$ contain the ImageNet indices associated
with the 100 benchmark synsets. Native Benchmark-100 restricts
the unchanged scores to these indices:
\begin{equation}
\widehat{y}_{100}(x)
=
\arg\max_{j\in\mathcal{Y}_{100}}z_j(x).
\label{eq:native_benchmark_100}
\end{equation}
This restriction introduces no fitted parameters. It supplies the
same answer space as the 100-label probes while retaining the
checkpoint's native scoring rule.

Secondary recognition metrics include the original-target score,
its native probability, and reciprocal rank.
For every checkpoint except SigLIP2, that probability is the
1,000-way softmax. SigLIP2 stores the sigmoid score instead,
and the two are not compared.
The saved target rank is
\begin{equation}
r_y(x)
=
1+\sum_{j=1}^{1000}
\mathbf{1}[z_j(x)>z_y(x)].
\end{equation}
These metrics describe different aspects of recognition.
In particular, a target-score increase need not change the
predicted category. Raw score magnitudes are interpreted within
checkpoint because their scales differ across models.

\subsection{Frozen Features and Numerical Precision}
\label{app:feature_extraction_details}

For models with a supplied 1,000-class linear classifier,
the runner captures its input through a forward pre-hook.
These pre-classifier features are used for the fitted-readout
experiments. CLIP instead supplies normalized image embeddings.

The runner enables CUDA automatic mixed precision with
\texttt{float16} autocasting. Inference scores and features are
converted to \texttt{float32} before export preparation.
Saved logits use \texttt{float32}, while saved feature arrays
use \texttt{float16}. The analysis loads features into
\texttt{float32} arrays for fitting. Thus, the probe evaluations
use the stored feature representation and its associated
numerical precision.
SigLIP2 exports prediction rows only. It has no saved feature
array and is excluded from the fitted readouts.

\subsection{Source and Category Aggregation}
\label{app:aggregation_details}

Renders are averaged within source before aggregation across
categories. For metric $m$, condition $v$, category $c$, and
source $i$, define
\begin{equation}
\overline{m}_{c,i,v}
=
\frac{1}{|\mathcal{R}_{c,i,v}|}
\sum_{r\in\mathcal{R}_{c,i,v}}m(r),
\end{equation}
where $\mathcal{R}_{c,i,v}$ contains that source's relevant
renders. The reported category-macro estimate is
\begin{equation}
\widehat{\mu}_v
=
\frac{1}{|\mathcal{C}|}
\sum_{c\in\mathcal{C}}
\frac{1}{n_c}
\sum_{i=1}^{n_c}\overline{m}_{c,i,v}.
\label{eq:category_source_aggregation}
\end{equation}

For the complete encoder benchmark, $|\mathcal{C}|=100$ and
$n_c=6$. This aggregation gives each category equal weight
and prevents repeated renders from being treated as independent
photographs. The same principle applies to carrier statistics,
paired interventions, and probe correctness.

\subsection{Bootstrap Confidence Intervals}
\label{app:bootstrap_details}

Primary encoder estimates use 10,000 bootstrap replicates~\citep{efron1979bootstrap}.
The implementation first constructs a source-resampled mean
for every category in each replicate, drawing $n_c$ source
observations with replacement. It independently draws category
multiplicities
\begin{equation}
(w_1^{*},\ldots,w_C^{*})
\sim
\operatorname{Multinomial}
\left(C;\frac{1}{C},\ldots,\frac{1}{C}\right),
\end{equation}
and combines them as
\begin{equation}
\widehat{\mu}^{*}
=
\frac{1}{C}
\sum_{c=1}^{C}w_c^{*}\overline{m}_c^{*}.
\label{eq:bootstrap_estimate}
\end{equation}
When a category receives multiplicity greater than one, its
source-resampled mean is reused with that multiplicity.
This specifies the implementation used for the stored intervals.

The 2.5th and 97.5th percentiles form the reported 95\%
confidence interval. Exploratory rendering-factor strata use
2,000 replicates under the local analysis configuration.
The base analysis and probe seed is 20260925, with deterministic
metric-specific offsets.

For paired effects, source-level differences are formed before
resampling, preserving condition pairing. Probe-versus-native
accuracy gaps are likewise computed from paired correctness
on the same Final renders before source aggregation.
The bootstrap does not refit probe heads or resample asset banks;
its intervals are conditional on the evaluated checkpoints,
rendering assets, and fitted heads.

Human and VLM uncertainty use their separately specified
procedures in Appendices~\ref{app:human_validation}
and~\ref{app:extended_vlm}, respectively.

\subsection{Paired Tests and Multiple Comparisons}
\label{app:paired_testing_details}

For conditions $a$ and $b$, the paired source difference is
\begin{equation}
d_{c,i}
=
\overline{m}_{c,i,a}
-
\overline{m}_{c,i,b}.
\end{equation}
The geometry/context tests use original-target scores and
600 paired sources. Because every category contributes six
sources, the mean source difference equals the category-macro
difference.

Two-sided Monte Carlo sign-flip tests use 20,000 draws.
Each draw independently multiplies source differences by
$-1$ or $+1$. The $p$-value is the proportion of randomized
absolute means at least as large as the observed absolute mean,
with one added to both the exceedance count and denominator~\citep{phipson2010permutation}.
Interpretation relies on sign symmetry of paired differences
under the null.

The primary correction family contains 28 tests:
seven checkpoints crossed with correct-minus-no-aperture,
correct-minus-mean-scramble, correct-minus-wrong-geometry,
and correct-minus-context-removed contrasts.
Holm adjustment~\citep{holm1979simple} is applied across this complete family.
The three individual scramble contrasts across seven checkpoints
form a separate 21-test family.

Presenting context removal with the opposite sign reverses
the effect and interval endpoints while retaining the original
two-sided $p$-value and correction family.
Intervals for individual aperture levels do not constitute
paired tests between those levels.

\subsection{Probe Procedure and Analysis Integrity}
\label{app:analysis_integrity}

The source-disjoint probe procedure uses six rotations, each
with four training sources, one validation source, and one
test source per category. All views derived from a source
inherit its fold. Ridge readouts use training-fold feature
standardization and select
$C\in\{0.01,0.1,1,10\}$ on validation accuracy, with ties
favoring smaller $C$. Final-trained diagnostics fit fresh
standardization and weights while reusing the corresponding
Shape-selected $C$. Full fitting and retention definitions
appear in Appendix~\ref{app:probe_details}.

Before analysis, the code checks schema compatibility,
checkpoint/preprocessing consistency, external-shard coverage,
manifest row coverage, duplicate identifiers, and alignment
between prediction records and saved arrays. Missing outputs
are not counted as incorrect classifications. The complete-data
configuration rejects incomplete encoder runs.
These checks support traceability between reported metrics
and the evaluated image records.

\section{Complete Recognition and Readout Results}
\label{app:recognition_results}

This appendix provides the complete accuracy estimates supporting
Section~\ref{sec:recognizing_structure}. The four reported views
are Original, Object Gray, Shape-only, and Final.
Section~\ref{app:recognition_ladder} reports the matched B1 ladder.
All estimates aggregate 600 source photographs across 100
categories using the procedure in
Appendix~\ref{app:evaluation_details}.

\subsection{Recognition Across Construction Views}
\label{app:recognition_transition_details}

Table~\ref{tab:recognition_intervals} reports native ImageNet-1000
top-1 and top-5 accuracy with their stored 95\% confidence
intervals. Original and Object Gray each contribute one image
per source, Shape-only contributes three aperture levels,
and Final contributes 24 renders. Repeated views are averaged
within source before category aggregation.

\begin{figure}[H]
    \centering
    \includegraphics[width=\columnwidth]
    {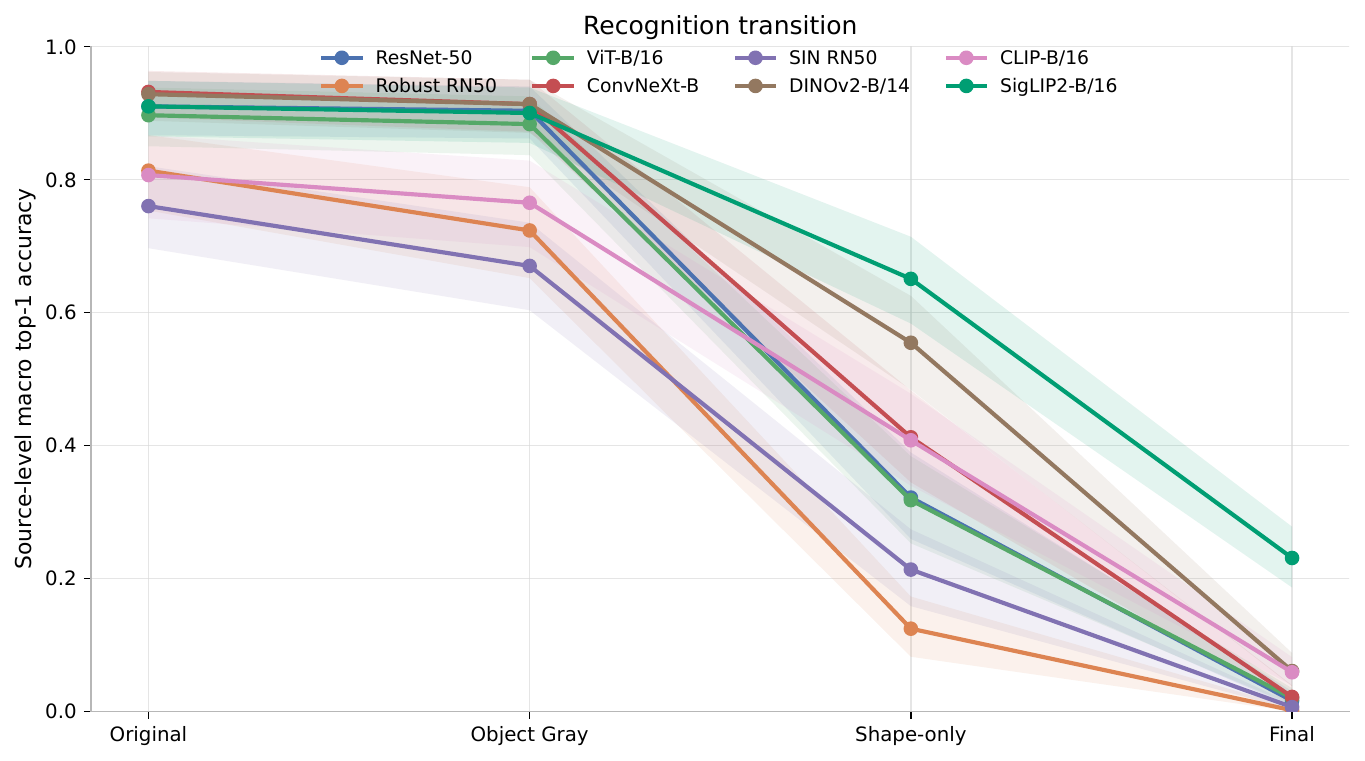}
    \caption{
    Native ImageNet-1000 top-1 recognition across the four
    reported views. Points show category-macro estimates after
    source aggregation; shaded bands show stored 95\% confidence
    intervals at each view. Lines connect construction stages
    for readability. Shape-only averages three apertures and
    Final averages 24 renders per source.
    Table~\ref{tab:ladder_intervals} reports the matched
    decomposition.
    }
    \label{fig:recognition_transition_supp}
\end{figure}

Figure~\ref{fig:recognition_transition_supp} shows the corresponding
top-1 trajectories. Its connecting lines organize the four
construction views; they do not represent a continuous experimental
variable. Shape-only averages three centered apertures and Final
averages 24 composites per source.
Table~\ref{tab:ladder_intervals} separates placement, heavy fog,
surrounding context, and the paired carrier on one B1 image
per source.

\begin{table*}[t]
    \centering
    \small
    \caption{
    Complete native ImageNet-1000 recognition estimates.
    Accuracies and interval endpoints are percentages.
    Each row aggregates 600 sources and 100 categories.
    Intervals describe individual estimates, rather than
    paired differences between views.
    }
    \label{tab:recognition_intervals}
    \begin{tabular}{@{}llrcrc@{}}
        \toprule
        Model & View
        & Top-1 & 95\% CI
        & Top-5 & 95\% CI \\
        \midrule
        ResNet-50 & Original
            & 91.00 & [86.67, 94.83]
            & 98.83 & [97.00, 99.83] \\
        & Object Gray
            & 90.33 & [86.17, 94.00]
            & 97.67 & [95.33, 99.33] \\
        & Shape-only
            & 32.17 & [25.94, 38.94]
            & 49.11 & [41.67, 56.33] \\
        & Final
            & 1.61 & [0.65, 3.03]
            & 4.72 & [2.85, 7.00] \\
        \midrule
        Robust RN50 & Original
            & 81.33 & [75.33, 86.67]
            & 93.50 & [90.00, 96.50] \\
        & Object Gray
            & 72.33 & [65.17, 78.83]
            & 89.50 & [84.67, 93.67] \\
        & Shape-only
            & 12.44 & [8.22, 17.28]
            & 25.61 & [19.61, 32.17] \\
        & Final
            & 0.17 & [0.00, 0.64]
            & 0.45 & [0.03, 1.23] \\
        \midrule
        ViT-B/16 & Original
            & 89.67 & [85.00, 93.83]
            & 98.33 & [96.33, 99.67] \\
        & Object Gray
            & 88.33 & [83.67, 92.50]
            & 96.17 & [93.17, 98.50] \\
        & Shape-only
            & 31.78 & [25.28, 38.39]
            & 51.28 & [43.94, 58.83] \\
        & Final
            & 2.01 & [0.94, 3.49]
            & 5.34 & [3.15, 7.93] \\
        \midrule
        ConvNeXt-B & Original
            & 93.17 & [89.33, 96.33]
            & 99.17 & [97.67, 100.00] \\
        & Object Gray
            & 91.33 & [87.17, 95.00]
            & 97.83 & [95.50, 99.50] \\
        & Shape-only
            & 41.22 & [34.44, 48.28]
            & 63.11 & [55.94, 70.06] \\
        & Final
            & 2.18 & [0.99, 3.93]
            & 7.72 & [5.14, 10.71] \\
        \midrule
        SIN RN50 & Original
            & 76.00 & [69.67, 82.00]
            & 90.17 & [85.83, 93.83] \\
        & Object Gray
            & 67.00 & [60.33, 73.50]
            & 85.00 & [79.83, 89.67] \\
        & Shape-only
            & 21.33 & [15.83, 27.39]
            & 40.28 & [33.33, 47.39] \\
        & Final
            & 0.65 & [0.25, 1.23]
            & 1.93 & [0.94, 3.33] \\
        \midrule
        DINOv2-B/14 & Original
            & 92.83 & [88.83, 96.17]
            & 99.67 & [98.67, 100.00] \\
        & Object Gray
            & 91.33 & [87.00, 95.00]
            & 98.67 & [96.83, 99.83] \\
        & Shape-only
            & 55.44 & [48.39, 62.50]
            & 76.33 & [69.78, 82.61] \\
        & Final
            & 6.07 & [3.79, 8.79]
            & 12.53 & [8.94, 16.59] \\
        \midrule
        CLIP-B/16 & Original
            & 80.67 & [74.17, 86.67]
            & 93.67 & [89.33, 97.33] \\
        & Object Gray
            & 76.50 & [69.83, 82.83]
            & 92.67 & [88.50, 96.17] \\
        & Shape-only
            & 40.78 & [33.94, 47.78]
            & 63.11 & [56.06, 69.72] \\
        & Final
            & 5.89 & [3.93, 8.16]
            & 15.05 & [11.40, 18.97] \\
        \midrule
        SigLIP2-B/16 & Original
            & 91.00 & [86.50, 94.83]
            & 99.17 & [97.67, 100.00] \\
        & Object Gray
            & 90.00 & [85.50, 93.83]
            & 97.67 & [95.33, 99.33] \\
        & Shape-only
            & 65.06 & [58.33, 71.39]
            & 82.22 & [77.05, 86.94] \\
        & Final
            & 23.08 & [18.64, 27.78]
            & 39.62 & [34.16, 45.33] \\
        \bottomrule
    \end{tabular}
\end{table*}

The measurements characterize the selected source population.
Original accuracy should therefore be interpreted on these
geometrically selected photographs, rather than as an estimate
for the full ImageNet validation distribution.

\subsection{Matched Shape-only to Final Ladder}
\label{app:recognition_ladder}

Table~\ref{tab:ladder_intervals} reports the five matched views
defined in Section~\ref{sec:benchmark_assembly}. Each view uses
one B1 image per source. Shape-only B1 is the centered stencil.
Placed shape uses the composite's placement on gray, without fog.
Fog, gray carrier adds the heavy-fog plate of the paired Final
render, with a gray carrier and gray pixels outside the glass.
Context removed and Final matched are the paired existing renders.
Aggregation and hierarchical 95\% intervals follow
Table~\ref{tab:recognition_intervals}.
Figure~\ref{fig:recognition_ladder} plots the top-1 trajectories.

Placement leaves top-1 close to the centered stencil. Heavy fog
on a gray carrier accounts for the collapse. Context removal
leaves point estimates at or above the fog step. For DINOv2,
fog lowers top-1 from 54.17\% to 15.00\%, and the paired Final
render is 6.50\% [3.17, 10.50].

\begin{table*}[t]
    \centering
    \small
    \caption{
    Matched B1 ladder. Native ImageNet-1000 top-1 and top-5
    accuracy. Accuracies and interval endpoints are percentages.
    Each row aggregates the same 600 sources and 100 categories.
    Intervals describe individual estimates, rather than
    paired differences between views. Final matched is the
    single paired render, while Final in
    Table~\ref{tab:recognition_intervals} averages 24 renders.
    }
    \label{tab:ladder_intervals}
    \begin{tabular}{@{}llrcrc@{}}
        \toprule
        Model & View
        & Top-1 & 95\% CI
        & Top-5 & 95\% CI \\
        \midrule
        ResNet-50 & Shape-only B1
            & 35.83 & [28.50, 43.17]
            & 52.50 & [44.67, 60.33] \\
        & Placed shape
            & 33.17 & [26.17, 40.50]
            & 50.17 & [42.50, 58.00] \\
        & Fog, gray carrier
            & 0.00 & [0.00, 0.00]
            & 0.50 & [0.00, 1.67] \\
        & Context removed
            & 3.50 & [1.33, 6.50]
            & 8.17 & [4.67, 12.33] \\
        & Final matched
            & 1.33 & [0.00, 3.33]
            & 5.33 & [2.50, 8.83] \\
        \midrule
        Robust RN50 & Shape-only B1
            & 12.83 & [8.33, 18.00]
            & 26.17 & [19.67, 33.17] \\
        & Placed shape
            & 10.33 & [6.17, 15.17]
            & 21.83 & [15.83, 28.50] \\
        & Fog, gray carrier
            & 0.00 & [0.00, 0.00]
            & 0.00 & [0.00, 0.00] \\
        & Context removed
            & 0.00 & [0.00, 0.00]
            & 0.17 & [0.00, 1.00] \\
        & Final matched
            & 0.17 & [0.00, 1.00]
            & 0.50 & [0.00, 1.83] \\
        \midrule
        ViT-B/16 & Shape-only B1
            & 34.83 & [27.33, 42.50]
            & 55.17 & [46.83, 63.17] \\
        & Placed shape
            & 31.33 & [24.33, 38.83]
            & 50.33 & [42.33, 58.17] \\
        & Fog, gray carrier
            & 0.00 & [0.00, 0.00]
            & 0.50 & [0.00, 1.83] \\
        & Context removed
            & 5.17 & [2.33, 8.67]
            & 12.67 & [8.00, 17.83] \\
        & Final matched
            & 2.50 & [0.67, 5.00]
            & 6.50 & [3.33, 10.33] \\
        \midrule
        ConvNeXt-B & Shape-only B1
            & 45.33 & [37.67, 53.00]
            & 65.50 & [58.00, 72.67] \\
        & Placed shape
            & 41.17 & [33.50, 48.67]
            & 61.50 & [54.00, 68.83] \\
        & Fog, gray carrier
            & 0.50 & [0.00, 1.67]
            & 0.83 & [0.00, 2.33] \\
        & Context removed
            & 7.83 & [4.50, 11.83]
            & 15.83 & [10.83, 21.50] \\
        & Final matched
            & 3.17 & [1.17, 6.00]
            & 9.17 & [5.33, 13.50] \\
        \midrule
        SIN RN50 & Shape-only B1
            & 23.00 & [16.83, 29.67]
            & 42.50 & [34.83, 50.17] \\
        & Placed shape
            & 18.67 & [13.00, 24.50]
            & 34.17 & [27.17, 41.33] \\
        & Fog, gray carrier
            & 0.17 & [0.00, 1.00]
            & 2.17 & [0.50, 4.50] \\
        & Context removed
            & 1.83 & [0.33, 3.83]
            & 3.50 & [1.17, 6.50] \\
        & Final matched
            & 0.50 & [0.00, 1.67]
            & 1.17 & [0.00, 3.00] \\
        \midrule
        DINOv2-B/14 & Shape-only B1
            & 57.00 & [49.33, 64.67]
            & 79.00 & [72.33, 85.17] \\
        & Placed shape
            & 54.17 & [46.50, 61.83]
            & 75.33 & [68.50, 82.00] \\
        & Fog, gray carrier
            & 15.00 & [10.00, 20.50]
            & 30.83 & [24.17, 38.00] \\
        & Context removed
            & 15.50 & [10.33, 21.33]
            & 26.67 & [20.17, 33.50] \\
        & Final matched
            & 6.50 & [3.17, 10.50]
            & 13.83 & [9.00, 19.33] \\
        \midrule
        CLIP-B/16 & Shape-only B1
            & 41.17 & [33.67, 48.83]
            & 64.50 & [57.17, 71.50] \\
        & Placed shape
            & 41.17 & [33.67, 48.50]
            & 63.00 & [55.67, 70.33] \\
        & Fog, gray carrier
            & 6.67 & [3.50, 10.50]
            & 15.00 & [10.17, 20.33] \\
        & Context removed
            & 9.67 & [5.83, 14.17]
            & 21.67 & [15.83, 28.00] \\
        & Final matched
            & 5.50 & [2.67, 8.83]
            & 13.50 & [9.00, 18.50] \\
        \midrule
        SigLIP2-B/16 & Shape-only B1
            & 66.50 & [59.33, 73.00]
            & 83.00 & [77.67, 88.00] \\
        & Placed shape
            & 62.83 & [55.67, 69.83]
            & 81.33 & [75.67, 86.50] \\
        & Fog, gray carrier
            & 24.83 & [19.00, 31.00]
            & 41.33 & [34.50, 48.50] \\
        & Context removed
            & 31.00 & [24.50, 38.00]
            & 47.83 & [40.67, 55.17] \\
        & Final matched
            & 21.67 & [15.83, 28.00]
            & 41.83 & [34.67, 49.17] \\
        \bottomrule
    \end{tabular}
\end{table*}

Table~\ref{tab:fog_content_intervals} gives the intervals for
the separate comparison in Section~\ref{sec:recognizing_structure}.
One incongruent carrier is rendered as the photograph, a phase
scramble matched to its mean and standard deviation, and a
constant fill at that mean. All three images share one heavy-fog
plate and a gray surround.

\paragraph{Photometric quality control.}
Of 3,000 post-composite photometric comparisons, five
(0.17\%) exceeded the predefined tolerances. Checks covered
aperture mean luminance and center-crop mean luminance for
both modified contents, and aperture luminance standard
deviation for the phase scramble only. All geometry checks
passed. The individual flagged records were not retained,
so their source-level effects could not be assessed.
All 600 paired sources were retained.

The phase-scramble minus photograph target-logit interval
excludes zero for ResNet-50, ViT-B/16, ConvNeXt-B, SIN RN50,
CLIP, and SigLIP2, all in favor of the photograph
(Table~\ref{tab:fog_content_logit}).
Robust RN50 and DINOv2 include zero.
The constant-fill minus photograph interval excludes zero for
six checkpoints. DINOv2 favors the constant fill, at $2.674$
$[2.186,3.156]$. CLIP and SigLIP2 include zero.

\begin{table*}[t]
    \centering
    \small
    \caption{
    Incongruent carrier content behind one heavy-fog plate.
    Native ImageNet-1000 top-1 accuracy (\%).
    Each cell aggregates 600 sources and 100 categories.
    These images are separate from the gray-carrier row in
    Table~\ref{tab:ladder_intervals}.
    }
    \label{tab:fog_content_intervals}
    \begin{tabular}{@{}lcccccc@{}}
        \toprule
        Model
        & Photograph & 95\% CI
        & Phase scramble & 95\% CI
        & Constant & 95\% CI \\
        \midrule
        ResNet-50
            & 3.83 & [1.50, 6.83]
            & 1.00 & [0.00, 2.67]
            & 0.33 & [0.00, 1.33] \\
        Robust RN50
            & 0.17 & [0.00, 1.00]
            & 0.00 & [0.00, 0.00]
            & 0.00 & [0.00, 0.00] \\
        ViT-B/16
            & 6.17 & [3.00, 10.17]
            & 2.50 & [0.67, 5.17]
            & 4.17 & [1.83, 7.17] \\
        ConvNeXt-B
            & 5.67 & [2.67, 9.33]
            & 2.83 & [0.67, 5.67]
            & 2.33 & [0.67, 4.67] \\
        SIN RN50
            & 1.50 & [0.17, 3.50]
            & 0.33 & [0.00, 1.33]
            & 0.17 & [0.00, 1.00] \\
        DINOv2-B/14
            & 15.33 & [10.17, 21.17]
            & 14.00 & [9.17, 19.33]
            & 25.17 & [18.83, 32.00] \\
        CLIP-B/16
            & 5.83 & [3.00, 9.50]
            & 5.33 & [2.50, 8.83]
            & 16.17 & [11.00, 21.83] \\
        SigLIP2-B/16
            & 31.50 & [25.00, 38.17]
            & 31.50 & [24.67, 38.67]
            & 41.50 & [34.50, 48.83] \\
        \bottomrule
    \end{tabular}
\end{table*}

\begin{table*}[t]
    \centering
    \small
    \caption{
    Paired target-logit differences behind the locked fog plate.
    Scramble minus photograph, and constant fill minus photograph.
    Negative values favor the photograph.
    Each interval is a 95\% hierarchical interval over 600 sources.
    }
    \label{tab:fog_content_logit}
    \begin{tabular}{@{}lcc@{}}
        \toprule
        Model
        & Scramble$-$photo
        & Constant$-$photo \\
        \midrule
        ResNet-50
            & $-0.341\;[-0.525,-0.165]$
            & $-0.509\;[-0.733,-0.304]$ \\
        Robust RN50
            & $0.066\;[-0.135,0.268]$
            & $-0.551\;[-0.734,-0.370]$ \\
        ViT-B/16
            & $-0.569\;[-0.858,-0.299]$
            & $-0.906\;[-1.232,-0.595]$ \\
        ConvNeXt-B
            & $-0.296\;[-0.544,-0.061]$
            & $-0.661\;[-0.900,-0.426]$ \\
        SIN RN50
            & $-1.015\;[-1.386,-0.648]$
            & $-0.617\;[-0.939,-0.307]$ \\
        DINOv2-B/14
            & $0.039\;[-0.437,0.510]$
            & $2.674\;[2.186,3.156]$ \\
        CLIP-B/16
            & $-0.542\;[-0.875,-0.207]$
            & $0.241\;[-0.092,0.586]$ \\
        SigLIP2-B/16
            & $-0.467\;[-0.838,-0.090]$
            & $0.408\;[-0.022,0.843]$ \\
        \bottomrule
    \end{tabular}
\end{table*}

\subsection{Native Recognition Under Two Answer Spaces}
\label{app:native_answer_space_results}

Table~\ref{tab:native_readout_space} compares Final top-1
recognition using the full ImageNet vocabulary and the
100 benchmark categories. Both decisions use the same saved
scores and images. Restriction removes competing labels
outside the benchmark without training a new head.

Every checkpoint has a higher point estimate under the
restricted answer space. Nevertheless, Benchmark-100 Final
accuracy ranges from 1.44\% to 17.81\%, leaving substantial
recognition failure even after this restriction.
These results also provide the native baseline for the
100-label probe comparisons in
Section~\ref{sec:structural_stability}.

\begin{table*}[t]
    \centering
    \small
    \caption{
    Native Final top-1 accuracy under two answer spaces.
    Values and interval endpoints are percentages.
    Scores are restricted without fitting.
    Intervals are taken from the answer-space analysis export;
    separately seeded bootstrap calls can produce slightly
    different endpoints from Table~\ref{tab:recognition_intervals}
    for the same ImageNet-1000 point estimates.
    }
    \label{tab:native_readout_space}
    \begin{tabular}{@{}lrcrc@{}}
        \toprule
        &
        \multicolumn{2}{c}{ImageNet-1000} &
        \multicolumn{2}{c}{Benchmark-100} \\
        \cmidrule(lr){2-3}
        \cmidrule(lr){4-5}
        Model & Top-1 & 95\% CI & Top-1 & 95\% CI \\
        \midrule
        ResNet-50
            & 1.61 & [0.66, 3.06]
            & 7.77 & [5.31, 10.69] \\
        Robust RN50
            & 0.17 & [0.00, 0.64]
            & 1.44 & [0.45, 2.88] \\
        ViT-B/16
            & 2.01 & [0.97, 3.47]
            & 7.60 & [4.99, 10.63] \\
        ConvNeXt-B
            & 2.18 & [0.99, 3.85]
            & 12.03 & [8.73, 15.76] \\
        SIN RN50
            & 0.65 & [0.25, 1.24]
            & 3.63 & [2.03, 5.56] \\
        DINOv2-B/14
            & 6.07 & [3.74, 8.73]
            & 15.05 & [11.03, 19.43] \\
        CLIP-B/16
            & 5.89 & [3.98, 8.17]
            & 17.81 & [14.02, 21.80] \\
        \bottomrule
    \end{tabular}
\end{table*}

\subsection{Secondary Metrics and Numerical Supplement}
\label{app:secondary_recognition_metrics}

The machine-readable supplement
\texttt{recognition\_full\_selected.csv} contains 140 estimates:
seven checkpoints, four views, and five metrics.
SigLIP2-B/16 top-1 and top-5 on all four views are reported
in the tables above; its five secondary metrics are not in
this file.
Each entry includes the point estimate, confidence interval,
source count, and category count.
The matched ladder is stored separately in
\texttt{T2\_ladder\_transition.csv}: eight checkpoints,
five views, and the same five metrics.

The five metrics are top-1 accuracy, top-5 accuracy,
original-target score, original-target probability under the
checkpoint's native link, and reciprocal target rank.
Their definitions appear in
Appendix~\ref{app:native_readout_details}.
The CSV stores accuracies and probabilities as fractions;
the displayed accuracy tables multiply these values by 100.

These secondary metrics distinguish changes in target evidence
from changes in the winning category. They do not provide
a common score scale across checkpoints or, by themselves,
identify the visual mechanism responsible for an error.
The matched ladder in Section~\ref{app:recognition_ladder}
identifies where top-1 accuracy falls.
Geometry interventions, carrier swaps, and fitted-readout
experiments address those complementary questions in
Appendices~\ref{app:geometry_carrier_results}
and~\ref{app:probe_details}.

\section{Geometry and Carrier Interventions}
\label{app:geometry_carrier_results}

This appendix provides the intervention definitions and supporting
results for Section~\ref{sec:geometry_or_content}. The original carrier
analyses use the complete Final benchmark, while fixed-plate and
nuisance controls use 600 matched B1 sources. Geometry interventions
use one deterministically selected B1 reference per source, giving
600 paired reference conditions across 100 categories.

\subsection{Carrier Blocks and Sensitivity Measures}
\label{app:carrier_sensitivity_details}

A carrier block holds the source, mirror, and aperture level fixed
and contains the four assigned carriers. Each source contributes
six blocks: two mirrors crossed with three aperture levels.
There are therefore 3,600 blocks and 14,400 Final images per
checkpoint.

For block $b$, let $x_{b,r}$ denote its image under carrier $r$.
The original-target score range is
\begin{equation}
R_b
=
\max_{r\in\{1,\ldots,4\}}z_y(x_{b,r})
-
\min_{r\in\{1,\ldots,4\}}z_y(x_{b,r}).
\label{eq:carrier_score_range}
\end{equation}
Prediction switching is defined as
\begin{equation}
S_b
=
\mathbf{1}
\left[
\left|
\left\{
\widehat{y}_{1000}(x_{b,r})
\right\}_{r=1}^{4}
\right|>1
\right].
\label{eq:carrier_block_switch}
\end{equation}
Thus, switching records whether the four predictions are not
all identical. It is not the fraction of pairwise carrier swaps
that change a prediction, and it does not measure correct
identification of carrier content.

Table~\ref{tab:carrier_sensitivity_full} reports score ranges
and switching. Mean ranges and switching rates average the six
blocks within each source before category aggregation.
Median ranges are descriptive medians over all 3,600 blocks.

\begin{table*}[t]
    \centering
    \small
    \caption{
    Complete carrier-sensitivity summaries.
    Score ranges use checkpoint-specific native score units;
    switching values and their intervals are percentages.
    Confidence intervals use source/category resampling.
    Median ranges summarize the block distribution without
    an associated interval.
    }
    \label{tab:carrier_sensitivity_full}
    \begin{tabular}{@{}lrcrrc@{}}
        \toprule
        Model
        & Mean range & 95\% CI
        & Median range
        & Switching & 95\% CI \\
        \midrule
        ResNet-50
            & 1.026 & [0.896, 1.174]
            & 0.578 & 70.25 & [66.67, 73.75] \\
        Robust RN50
            & 0.431 & [0.391, 0.479]
            & 0.314 & 43.42 & [39.44, 47.53] \\
        ViT-B/16
            & 1.431 & [1.246, 1.626]
            & 0.837 & 68.83 & [64.78, 72.81] \\
        ConvNeXt-B
            & 1.449 & [1.298, 1.614]
            & 0.957 & 53.42 & [49.39, 57.47] \\
        SIN RN50
            & 1.843 & [1.708, 1.988]
            & 1.418 & 73.06 & [69.50, 76.47] \\
        DINOv2-B/14
            & 3.225 & [2.990, 3.470]
            & 2.758 & 64.25 & [60.47, 67.95] \\
        CLIP-B/16
            & 3.345 & [3.145, 3.551]
            & 2.938 & 80.50 & [77.19, 83.58] \\
        \bottomrule
    \end{tabular}
\end{table*}

Carrier swaps retain aperture placement and the underlying fog
fields. However, the fog plate depends partly on carrier appearance
(Appendix~\ref{app:rendering_details}), so their visual effects
extend beyond the clear aperture within the glass.
These measurements establish carrier-assignment sensitivity
without isolating a purely semantic content mechanism.

\begin{table}[t]
    \centering
    \footnotesize
    \caption{
    \textbf{Carrier dependence and geometry sensitivity.}
    Switching (\%) measures prediction variation across
    four carriers at fixed geometry. Other columns report
    paired correct-minus-control target-score effects
    ($\Delta s_y$) over 600 sources at $B1$.
    Second lines give 95\% CIs. All geometry contrasts
    have positive CIs and Holm-adjusted $p<0.01$.
    Scores are comparable only within checkpoints.
    }
    \label{tab:geometry_and_carrier}

    \setlength{\tabcolsep}{2.6pt}
    \renewcommand{\arraystretch}{1.10}

    \begin{tabular*}{\columnwidth}{
        @{\extracolsep{\fill}}
        lcccc
        @{}
    }
        \toprule
        \textbf{Model}
        & \shortstack{\textbf{Switch}\\(\%)}
        & \shortstack{\textbf{No}\\\textbf{aperture}}
        & \shortstack{\textbf{Mean}\\\textbf{scramble}}
        & \shortstack{\textbf{Wrong}\\\textbf{geometry}} \\
        \midrule

        ResNet-50
        & \shortstack{70.3\\{\scriptsize 66.7--73.8}}
        & \shortstack{0.646\\{\scriptsize 0.485--0.828}}
        & \shortstack{0.573\\{\scriptsize 0.420--0.749}}
        & \shortstack{0.581\\{\scriptsize 0.422--0.763}} \\
        \addlinespace[1pt]

        Robust RN50
        & \shortstack{43.4\\{\scriptsize 39.4--47.5}}
        & \shortstack{0.293\\{\scriptsize 0.142--0.446}}
        & \shortstack{0.138\\{\scriptsize 0.068--0.221}}
        & \shortstack{0.233\\{\scriptsize 0.158--0.317}} \\
        \addlinespace[1pt]

        ViT-B/16
        & \shortstack{68.8\\{\scriptsize 64.8--72.8}}
        & \shortstack{0.940\\{\scriptsize 0.704--1.218}}
        & \shortstack{0.803\\{\scriptsize 0.587--1.054}}
        & \shortstack{0.848\\{\scriptsize 0.608--1.111}} \\
        \addlinespace[1pt]

        ConvNeXt-B
        & \shortstack{53.4\\{\scriptsize 49.4--57.5}}
        & \shortstack{1.200\\{\scriptsize 0.969--1.458}}
        & \shortstack{1.101\\{\scriptsize 0.863--1.363}}
        & \shortstack{1.131\\{\scriptsize 0.896--1.388}} \\
        \addlinespace[1pt]

        SIN RN50
        & \shortstack{73.1\\{\scriptsize 69.5--76.5}}
        & \shortstack{0.950\\{\scriptsize 0.679--1.241}}
        & \shortstack{0.883\\{\scriptsize 0.664--1.115}}
        & \shortstack{0.813\\{\scriptsize 0.568--1.068}} \\
        \addlinespace[1pt]

        DINOv2-B/14
        & \shortstack{64.2\\{\scriptsize 60.5--67.9}}
        & \shortstack{3.364\\{\scriptsize 2.832--3.920}}
        & \shortstack{3.184\\{\scriptsize 2.711--3.692}}
        & \shortstack{3.253\\{\scriptsize 2.744--3.793}} \\
        \addlinespace[1pt]

        CLIP-B/16
        & \shortstack{80.5\\{\scriptsize 77.2--83.6}}
        & \shortstack{3.423\\{\scriptsize 2.978--3.864}}
        & \shortstack{3.003\\{\scriptsize 2.624--3.401}}
        & \shortstack{2.353\\{\scriptsize 1.942--2.780}} \\

        \bottomrule
    \end{tabular*}
\end{table}

\subsection{Fixed-Plate Carrier Switching and Nuisance Controls}
\label{app:carrier_nuisance_controls}

The full-benchmark carrier blocks change carrier assignment while
allowing the carrier-dependent fog appearance to change within the
glass. To separate this variation from carrier replacement, we use
one matched B1 reference for each of 600 sources (100 categories).
We compare four carriers with the reference fog plate held fixed.
Two nuisance controls instead hold carrier content fixed while
varying four fog seeds or JPEG quality (95, 85, 75, and 60).
Each condition thus has four images per source. Switching uses
Equation~\ref{eq:carrier_block_switch}, now with one four-image
block per source, and is not a pairwise flip rate. Estimates are
category-macro percentages with 95\% hierarchical class-then-source
bootstrap intervals.

\begin{table*}[t]
    \centering
    \footnotesize
    \setlength{\tabcolsep}{5pt}
    \renewcommand{\arraystretch}{1.12}
    \caption{
    \textbf{Carrier switching under a fixed fog plate and nuisance controls.}
    Each entry is the percentage of 600 B1 sources whose four native
    ImageNet-1000 top-1 predictions are not all identical, followed
    by a 95\% hierarchical interval. The carrier condition changes
    photographs behind one locked fog plate; the other conditions
    keep the carrier fixed and vary the fog seed or JPEG quality.
    These controls are not matched for perturbation magnitude.
    }
    \label{tab:carrier_nuisance_controls}
    \begin{tabular*}{0.89\textwidth}{@{\extracolsep{\fill}}lccc@{}}
        \toprule
        Model & Fixed-plate carrier & Fog seed & JPEG quality \\
        \midrule
        ResNet-50
            & \shortstack{67.2\%\\{\scriptsize [60.8, 73.2]}}
            & \shortstack{38.3\%\\{\scriptsize [32.2, 44.8]}}
            & \shortstack{32.0\%\\{\scriptsize [26.0, 38.2]}} \\
        Robust RN50
            & \shortstack{38.5\%\\{\scriptsize [32.0, 45.2]}}
            & \shortstack{8.3\%\\{\scriptsize [5.0, 12.2]}}
            & \shortstack{2.5\%\\{\scriptsize [0.8, 5.0]}} \\
        ViT-B/16
            & \shortstack{68.0\%\\{\scriptsize [61.8, 74.0]}}
            & \shortstack{33.8\%\\{\scriptsize [27.8, 40.0]}}
            & \shortstack{28.8\%\\{\scriptsize [22.7, 35.2]}} \\
        ConvNeXt-B
            & \shortstack{53.5\%\\{\scriptsize [47.3, 59.8]}}
            & \shortstack{26.8\%\\{\scriptsize [21.3, 32.7]}}
            & \shortstack{37.8\%\\{\scriptsize [31.8, 44.0]}} \\
        SIN RN50
            & \shortstack{72.0\%\\{\scriptsize [66.3, 77.7]}}
            & \shortstack{42.7\%\\{\scriptsize [36.3, 49.2]}}
            & \shortstack{37.2\%\\{\scriptsize [31.2, 43.5]}} \\
        DINOv2-B/14
            & \shortstack{54.7\%\\{\scriptsize [48.2, 61.2]}}
            & \shortstack{35.0\%\\{\scriptsize [28.8, 41.3]}}
            & \shortstack{33.5\%\\{\scriptsize [27.5, 39.7]}} \\
        CLIP-B/16
            & \shortstack{73.0\%\\{\scriptsize [67.0, 78.7]}}
            & \shortstack{41.0\%\\{\scriptsize [34.5, 47.5]}}
            & \shortstack{38.2\%\\{\scriptsize [31.8, 44.7]}} \\
        SigLIP2-B/16
            & \shortstack{67.0\%\\{\scriptsize [60.7, 73.0]}}
            & \shortstack{36.3\%\\{\scriptsize [30.2, 42.5]}}
            & \shortstack{26.3\%\\{\scriptsize [20.8, 32.3]}} \\
        \bottomrule
    \end{tabular*}
\end{table*}

With fog fixed, carrier switching remains 38.5--73.0\% across the
eight evaluated checkpoints, compared with 8.3--42.7\% for fog-seed
variation and 2.5--38.2\% across JPEG qualities. For every
checkpoint, the reported carrier interval lies above the separate
intervals for both nuisance controls; these are not confidence
intervals for paired differences. Matching the baseline render to
the stored Final JPEG gives mean RGB reconstruction error 0.93 on
the 0--255 scale. Under carrier replacement, mean RGB differences
are 22.45 within the aperture and 0.33 inside the glass but
outside the aperture. Changes are therefore concentrated in the
intended region, although the controls are not equal in visual
magnitude. Substantial fog- and JPEG-only switching also shows
that predictions can be unstable under nonsemantic changes.
The comparison supports greater sensitivity to the tested carrier
replacements than to these nuisance variations; it does not
establish semantic recognition of carrier content.

\subsection{Reference Selection and Control Rendering}
\label{app:geometry_control_construction}

For each source, the reference is the first B1 Final render under
deterministic ordering by mirror and carrier filenames.
The intervention manifest records its sample identifier,
mirror, carrier, and aperture path.

The renderer reconstructs this reference using the corresponding
assets and fog-generation inputs. Before creating controls, it
checks that the reconstruction's mean absolute RGB difference
from the decoded stored Final image is at most 2.0 on the
0--255 scale. The correct-reference condition uses that stored
image; modified apertures are rendered with the reconstructed
carrier and fog configuration.

The eight condition records per source are the correct reference,
no aperture, three scramble families, wrong-class geometry,
context removal, and mirror-region crop. The 4,800 records
include the 600 correct references derived from Final.

\paragraph{No aperture.}
The placed aperture is replaced by a zero mask. The registered
carrier, fog plate, mirror, and surrounding context are retained.

\paragraph{Scrambled geometry.}
The placed correct B1 mask is transformed using the procedures
below, then passed through the same compositing routine.

\paragraph{Wrong-class geometry.}
A B1 aperture from another target category is selected by
descriptor matching and independently placed in the reference
mirror.

\paragraph{Context removal and crop.}
Context removal replaces pixels outside the glass mask by RGB
value 127 in the decoded reference array. The mirror-region
condition crops that array to the glass-mask bounding box.
Both are subsequently JPEG-encoded. Their interpretation and
paired context results are discussed in
Appendix~\ref{app:critical_ablations}.

\subsection{Scramble Generation and Matching}
\label{app:scramble_matching_details}

Three transformations operate on the tight bounding box of the
placed reference mask:
\begin{enumerate}
    \item \textbf{Tile scramble:}
    downsample the mask to $8\times8$, permute its 64 values,
    and upsample to the original bounding-box dimensions.

    \item \textbf{Row-band scramble:}
    partition the bounding box into eight horizontal bands
    and permute their order.

    \item \textbf{Column-band scramble:}
    partition it into eight vertical bands and permute
    their order.
\end{enumerate}

Each candidate is Gaussian-smoothed with standard deviation
sampled uniformly from $[0.8,20.0]$. An exact-area operation
then selects the highest-valued pixels within the reference
bounding box until the candidate has the same foreground-pixel
count as the reference. Small random perturbations break ties.

For each family and source, 64 candidates are generated.
Let $K$ and $K'$ denote the reference and candidate masks.
Their descriptors are area $a$, centroid $\mathbf{c}$,
bounding-box width $w$ and height $h$, and perimeter $p$.
Perimeter is the summed length of external contours.
The matching residuals are
\begin{equation}
\begin{aligned}
e_a &=
\frac{|a(K')-a(K)|}{\max(a(K),1)},\\
e_c &=
\frac{\|\mathbf{c}(K')-\mathbf{c}(K)\|_2}
     {\max(\sqrt{w(K)^2+h(K)^2},1)},\\
e_b &=
\max\left(
\frac{|w(K')-w(K)|}{\max(w(K),1)},
\frac{|h(K')-h(K)|}{\max(h(K),1)}
\right),\\
e_p &=
\frac{|p(K')-p(K)|}{\max(p(K),1)}.
\end{aligned}
\label{eq:scramble_matching_residuals}
\end{equation}
The selected candidate minimizes
$e_a+2e_c+e_b+e_p$.
Its QC flag requires
$e_a\leq0.005$, $e_c\leq0.05$,
$e_b\leq0.10$, and $e_p\leq0.30$.
The selected candidate is retained regardless of its QC flag.

Table~\ref{tab:scramble_qc} distinguishes exact area counts from
approximate matching of the other descriptors. These residuals
describe the binary control masks before compositing. Glass-mask
restriction and aperture softening are subsequent operations.
A separate post-render pixel-level audit is reported in
Appendix~\ref{app:post_render_scramble_audit}; it does not
establish identical transmitted aperture area or boundary
properties after rendering.

\begin{table*}[t]
    \centering
    \small
    \caption{
    Scramble matching diagnostics.
    Residual columns report means using the normalization in
    Equation~\ref{eq:scramble_matching_residuals}.
    QC pass rates are percentages.
    Each family contains 600 source-specific controls;
    their repeated checkpoint records do not constitute
    additional independent masks.
    }
    \label{tab:scramble_qc}
    \begin{tabular}{@{}lrrrrr@{}}
        \toprule
        Family & QC pass
        & Area & Centroid & Bounding box & Perimeter \\
        \midrule
        Tile
            & 38.17 & 0.0000 & 0.0372 & 0.0064 & 0.3046 \\
        Row band
            & 67.33 & 0.0000 & 0.0280 & 0.0314 & 0.1318 \\
        Column band
            & 71.67 & 0.0000 & 0.0262 & 0.0254 & 0.1339 \\
        \bottomrule
    \end{tabular}
\end{table*}

\subsection{Effects for Individual Scramble Families}
\label{app:individual_scramble_results}

Table~\ref{tab:individual_scramble_effects} reports all 21
correct-minus-scrambled target-score effects.
Every stored confidence interval lies above zero.
The three families therefore support the same directional
finding as their combined contrast in
Table~\ref{tab:geometry_and_carrier}.

These results show sensitivity to the retained spatial
organization under the evaluated transformations.
Because matching is incomplete and local image statistics
change, they do not identify a geometry-exclusive mechanism.

\begin{table*}[t]
    \centering
    \small
    \renewcommand{\arraystretch}{1.15}
    \caption{
    Correct-minus-scrambled original-target scores.
    Each cell gives the mean effect and its 95\% confidence
    interval over 600 paired sources and 100 categories.
    All 21 Holm-adjusted $p$-values are approximately 0.00105
    in the individual-scramble testing family.
    Score units differ across checkpoints.
    }
    \label{tab:individual_scramble_effects}
    \begin{tabular}{@{}lccc@{}}
        \toprule
        Model & Tile & Row band & Column band \\
        \midrule
        ResNet-50
            & \shortstack{0.601\\{}[0.434, 0.785]}
            & \shortstack{0.572\\{}[0.414, 0.750]}
            & \shortstack{0.546\\{}[0.397, 0.723]} \\
        Robust RN50
            & \shortstack{0.173\\{}[0.073, 0.281]}
            & \shortstack{0.111\\{}[0.047, 0.187]}
            & \shortstack{0.131\\{}[0.064, 0.208]} \\
        ViT-B/16
            & \shortstack{0.880\\{}[0.642, 1.144]}
            & \shortstack{0.783\\{}[0.560, 1.036]}
            & \shortstack{0.746\\{}[0.537, 0.983]} \\
        ConvNeXt-B
            & \shortstack{1.173\\{}[0.919, 1.445]}
            & \shortstack{1.067\\{}[0.831, 1.323]}
            & \shortstack{1.063\\{}[0.823, 1.321]} \\
        SIN RN50
            & \shortstack{0.903\\{}[0.638, 1.179]}
            & \shortstack{0.893\\{}[0.670, 1.131]}
            & \shortstack{0.853\\{}[0.637, 1.080]} \\
        DINOv2-B/14
            & \shortstack{3.359\\{}[2.859, 3.892]}
            & \shortstack{3.109\\{}[2.621, 3.620]}
            & \shortstack{3.083\\{}[2.617, 3.592]} \\
        CLIP-B/16
            & \shortstack{3.631\\{}[3.207, 4.068]}
            & \shortstack{2.644\\{}[2.261, 3.044]}
            & \shortstack{2.734\\{}[2.360, 3.134]} \\
        \bottomrule
    \end{tabular}
\end{table*}

Table~\ref{tab:scramble_qc_pass} repeats the same contrast on
the masks that pass the published QC rule. Tile retains 229
sources from 92 classes, row band 404 sources from 98 classes,
and column band 430 sources from 98 classes. Every interval
lies above zero. The failing masks do as well. Halving and
doubling the four residual tolerances, a choice fixed before
the scores were examined, also leaves all 21 target-logit
intervals above zero. The strict tile subset contains 87
sources from 56 classes. This is a sensitivity analysis.
It is not a replacement for
Table~\ref{tab:individual_scramble_effects}, and passing QC
does not isolate global geometry from the remaining
low-level differences.

\begin{table*}[t]
    \centering
    \small
    \renewcommand{\arraystretch}{1.15}
    \caption{
    Correct-minus-scrambled target logit on QC-passing masks.
    Each cell is the class macro over the classes that retain
    a passing source, with its 95\% interval.
    Counts are 229, 404, and 430 sources.
    The analysis is exploratory and is separate from the
    Holm family of
    Table~\ref{tab:individual_scramble_effects}.
    }
    \label{tab:scramble_qc_pass}
    \begin{tabular}{@{}lccc@{}}
        \toprule
        Model & Tile & Row band & Column band \\
        \midrule
        ResNet-50
            & \shortstack{0.727\\{}[0.500, 0.997]}
            & \shortstack{0.475\\{}[0.320, 0.657]}
            & \shortstack{0.454\\{}[0.306, 0.635]} \\
        Robust RN50
            & \shortstack{0.145\\{}[0.014, 0.280]}
            & \shortstack{0.101\\{}[0.038, 0.167]}
            & \shortstack{0.119\\{}[0.056, 0.189]} \\
        ViT-B/16
            & \shortstack{1.057\\{}[0.699, 1.460]}
            & \shortstack{0.707\\{}[0.474, 0.975]}
            & \shortstack{0.629\\{}[0.414, 0.882]} \\
        ConvNeXt-B
            & \shortstack{1.265\\{}[0.935, 1.646]}
            & \shortstack{0.953\\{}[0.707, 1.226]}
            & \shortstack{0.886\\{}[0.652, 1.139]} \\
        SIN RN50
            & \shortstack{0.869\\{}[0.522, 1.232]}
            & \shortstack{0.813\\{}[0.573, 1.060]}
            & \shortstack{0.785\\{}[0.554, 1.033]} \\
        DINOv2-B/14
            & \shortstack{3.492\\{}[2.804, 4.224]}
            & \shortstack{3.036\\{}[2.492, 3.603]}
            & \shortstack{2.865\\{}[2.362, 3.398]} \\
        CLIP-B/16
            & \shortstack{3.564\\{}[2.998, 4.154]}
            & \shortstack{2.519\\{}[2.111, 2.948]}
            & \shortstack{2.577\\{}[2.200, 2.976]} \\
        \bottomrule
    \end{tabular}
\end{table*}

\subsection{Post-Render Scramble Audit}
\label{app:post_render_scramble_audit}

We compared the saved correct and scrambled B1 JPEGs
over all 600 paired sources. Table~\ref{tab:post_render_scramble}
summarizes pixel-level differences measured inside
the mirror glass. The correct and scrambled images
share the same source, mirror, and carrier assignment.

\begin{table}[t]
\centering
\footnotesize
\caption{Post-render correct-versus-scramble comparisons.
Values are medians over 600 source pairs.
MAE and luminance differences use 0--255 intensity units.
The final column reports RGB MAE more than 32 pixels
outside the glass.}
\label{tab:post_render_scramble}
\setlength{\tabcolsep}{3pt}
\begin{tabular}{@{}lcccc@{}}
\toprule
Scramble & Glass MAE & $|\Delta Y|$ & SSIM
& Outside MAE \\
\midrule
Tile   & 6.46 & 0.36 & 0.862 & 0.052 \\
Row    & 5.54 & 0.25 & 0.864 & 0.052 \\
Column & 5.52 & 0.26 & 0.866 & 0.052 \\
\bottomrule
\end{tabular}
\end{table}

All 21 correct-minus-scramble target-logit intervals
remain positive on the pre-render QC-passing subsets.
Target-rank intervals also favor the correct aperture
in all 21 comparisons. The same holds for target logits
in the lowest terciles of glass-region pixel, luminance,
contrast, and gradient differences.

The three scrambled renders are pixel-identical to
one another beyond 32 pixels outside the glass.
The small correct-versus-scramble residual there is
consistent with differences between the stored
reference JPEG and reconstructed controls.
Within the glass, local contrast and gradients also
change. Because post-render aperture masks were
unavailable, effective transmitted area and boundary
matching were not directly verified. These findings
support sensitivity to the tested spatial arrangements
but do not isolate geometry from all low-level cues.

\subsection{Wrong-Class Geometry and Category Competition}
\label{app:wrong_geometry_details}

Replacement apertures are selected from sources belonging to
a different target category. Matching uses B1 foreground
density over the stencil canvas and perimeter normalized
by the square root of foreground area.
The sum of absolute descriptor differences, each divided
by its standard deviation across sources, determines the
selected replacement. Standard deviations have a minimum
of $10^{-6}$, and source identifiers break ties.
The replacement aperture is then placed independently
in the reference mirror.

This matching does not enforce equal placed area, identical
scale, or balanced replacement-category frequencies.
All 100 classes appear as donors. A donor class is used
between 1 and 13 times, and the 600 sources form 569
original--donor pairs. The resulting control tests this
particular assignment. It is not relabeled as balanced.

Within a wrong-geometry image $x_i^{\mathrm{wrong}}$, define
the original-versus-replacement category margin as
\begin{equation}
\Gamma_i
=
z_{y_i}(x_i^{\mathrm{wrong}})
-
z_{y_i^{\mathrm{replacement}}}
(x_i^{\mathrm{wrong}}).
\label{eq:wrong_geometry_category_margin}
\end{equation}
Negative values indicate greater native score for the
replacement category than for the original target.
The paired shift compares that margin with the same two
scores on the correct image:
\begin{equation}
\begin{aligned}
\Delta_i
&=
\left[z_{z_i}(x_i^{\mathrm{wrong}})
-
z_{y_i}(x_i^{\mathrm{wrong}})\right]\\
&\quad-
\left[z_{z_i}(x_i^{\mathrm{correct}})
-
z_{y_i}(x_i^{\mathrm{correct}})\right],
\end{aligned}
\label{eq:donor_shift}
\end{equation}
where $z_i$ is the donor category. A positive value means
donor evidence rose relative to original-class evidence.
The two score changes are reported separately, because the
shift can come from a fall in the original class, a rise
in the donor class, or both.

\begin{figure}[H]
    \centering
    \includegraphics[width=0.85\columnwidth]
    {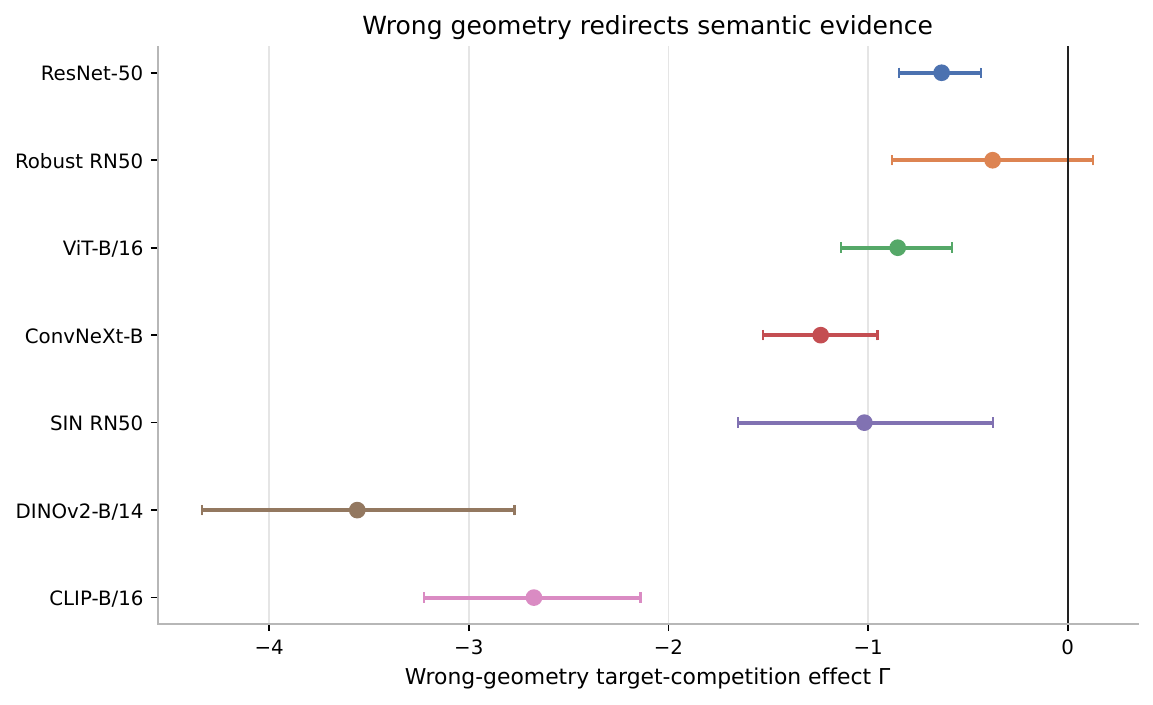}
    \caption{
    Original-target minus replacement-category score within
    wrong-geometry images.
    Points show category-macro margins after source aggregation;
    bars show stored 95\% confidence intervals over 600 sources.
    The vertical line marks equal scores for the two categories.
    Score scales are checkpoint-specific.
    }
    \label{fig:wrong_geometry_supp}
\end{figure}

Figure~\ref{fig:wrong_geometry_supp} and
Table~\ref{tab:wrong_geometry_margin} report $\Gamma_i$.
Six checkpoint intervals lie below zero; the robust RN50
interval includes zero. This within-image comparison does
not measure a paired increase in replacement-category score
relative to the correct reference, or establish that the
replacement category is the top-1 prediction.
Table~\ref{tab:donor_shift} reports $\Delta_i$. Every
shift interval lies above zero. Every original-class
change lies below zero, and every donor-class change lies
above zero. The shift is exploratory and is not added to
the Holm family of Table~\ref{tab:geometry_and_carrier}.

\begin{table}[t]
    \centering
    \footnotesize
    \caption{
    Within-image wrong-geometry category margins and
    95\% confidence intervals.
    }
    \label{tab:wrong_geometry_margin}
    \begin{tabular}{@{}lrc@{}}
        \toprule
        Model & Margin & 95\% CI \\
        \midrule
        ResNet-50
            & $-0.632$ & [$-0.845$, $-0.436$] \\
        Robust RN50
            & $-0.376$ & [$-0.880$, $0.127$] \\
        ViT-B/16
            & $-0.852$ & [$-1.138$, $-0.581$] \\
        ConvNeXt-B
            & $-1.237$ & [$-1.526$, $-0.953$] \\
        SIN RN50
            & $-1.019$ & [$-1.650$, $-0.374$] \\
        DINOv2-B/14
            & $-3.559$ & [$-4.334$, $-2.771$] \\
        CLIP-B/16
            & $-2.674$ & [$-3.224$, $-2.140$] \\
        \bottomrule
    \end{tabular}
\end{table}

\begin{table*}[t]
    \centering
    \small
    \renewcommand{\arraystretch}{1.15}
    \caption{
    Paired donor shift on the existing wrong-class assignment.
    Shift is Equation~\ref{eq:donor_shift}.
    Original-class change is the source-category logit on the
    wrong image minus its logit on the correct image.
    Donor-class change is the same difference for the donor
    category. Each cell gives the class macro and its 95\%
    interval over 600 sources.
    }
    \label{tab:donor_shift}
    \begin{tabular}{@{}lccc@{}}
        \toprule
        Model & Shift & Original class & Donor class \\
        \midrule
        ResNet-50
            & \shortstack{1.236\\{}[1.008, 1.488]}
            & \shortstack{$-0.581$\\{}[$-0.759$, $-0.417$]}
            & \shortstack{0.655\\{}[0.500, 0.832]} \\
        Robust RN50
            & \shortstack{0.432\\{}[0.328, 0.549]}
            & \shortstack{$-0.233$\\{}[$-0.321$, $-0.159$]}
            & \shortstack{0.200\\{}[0.130, 0.272]} \\
        ViT-B/16
            & \shortstack{1.665\\{}[1.354, 2.007]}
            & \shortstack{$-0.848$\\{}[$-1.108$, $-0.615$]}
            & \shortstack{0.818\\{}[0.622, 1.049]} \\
        ConvNeXt-B
            & \shortstack{2.333\\{}[2.008, 2.679]}
            & \shortstack{$-1.131$\\{}[$-1.382$, $-0.897$]}
            & \shortstack{1.202\\{}[0.985, 1.425]} \\
        SIN RN50
            & \shortstack{1.686\\{}[1.337, 2.056]}
            & \shortstack{$-0.813$\\{}[$-1.075$, $-0.575$]}
            & \shortstack{0.873\\{}[0.639, 1.124]} \\
        DINOv2-B/14
            & \shortstack{6.607\\{}[5.912, 7.376]}
            & \shortstack{$-3.253$\\{}[$-3.787$, $-2.743$]}
            & \shortstack{3.354\\{}[2.917, 3.813]} \\
        CLIP-B/16
            & \shortstack{4.923\\{}[4.312, 5.587]}
            & \shortstack{$-2.353$\\{}[$-2.789$, $-1.940$]}
            & \shortstack{2.570\\{}[2.204, 2.949]} \\
        \bottomrule
    \end{tabular}
\end{table*}

\subsection{Complete Contrast Exports}
\label{app:intervention_exports}

The numerical supplement retains the carrier summaries in
\texttt{full\_N02\_carrier\_sensitivity.csv}, the original
28-test geometry/context family in
\texttt{full\_T3\_causal\_contrasts.csv}, and the 21 individual
scramble contrasts in
\texttt{scramble\_family\_effects.csv}.
Mask-matching diagnostics and within-image category margins
are retained in \texttt{scramble\_qc.csv} and
\texttt{wrong\_geometry\_margin.csv}.
Same-category, log-probability, and rank contrasts are retained
in \texttt{geometry\_contrasts.csv}.
The QC-pass sensitivity and the paired donor shift are retained
in \texttt{item9\_scramble\_sensitivity.csv} and
\texttt{item9\_donor\_shift.csv}. The fixed-plate control analysis
additionally uses \texttt{switching.csv} and
\texttt{photometry\_summary.csv}, with source-level switching
indicators in \texttt{switching\_by\_source.parquet}.

The geometry contrasts measure changes in original-target
scores under paired conditions. Carrier switching measures
prediction variation across four renders. The wrong-geometry
margin compares two categories within one altered image, and
the donor shift compares that margin with the correct image.
These estimands provide complementary evidence and are
interpreted separately.

\subsection{Same-Category Apertures, Log-Probability, and Rank}
\label{app:same_category_geometry}

The geometry contrasts in Table~\ref{tab:geometry_and_carrier}
use the native target logit. The same paired sources also carry
a target log-probability and a target rank. Log-probability is
the natural logarithm of the target-class probability. Rank 1
is the top class. For a score, the contrast remains correct
minus control, as in Equation~\ref{eq:geometry_effect}. For rank,
the reported effect is control rank minus correct rank, so a
positive value still favors the source aperture.

The same-category control keeps the reference mirror, carrier,
fog field, and placement. Its aperture is the B1 stencil of a
different photograph from the same category. Among the other
five photographs in that category, the donor minimizes the sum
of absolute differences in foreground density and in perimeter
normalized by the square root of foreground area, each divided
by its standard deviation across the 600 sources. Standard
deviations have a minimum of $10^{-6}$, and source identifiers
break ties. Model scores are not used. Every selected donor
selects the original source in return. The median matching
distance is 0.36. No donor is the source itself or the
wrong-class donor already assigned to that source.

The correct-reference image is the stored reference JPEG.
Reconstructing that reference from the same assets reproduces
the stored reconstruction error: the maximum mean absolute RGB
difference from the decoded reference is 1.45, below the
renderer gate of 2. Rescoring those stored JPEGs changes the
class-macro target logit by less than 0.0001 for every
checkpoint. The same-category image is rendered with that
locked configuration and scored in the same session.
Intervals in
Tables~\ref{tab:geometry_same_logit}--\ref{tab:geometry_rank}
are 95\% hierarchical bootstrap intervals from 10,000
replicates. They are not added to the Holm family reported
for Table~\ref{tab:geometry_and_carrier}.

\begin{table*}[t]
    \centering
    \small
    \renewcommand{\arraystretch}{1.15}
    \caption{
    Target-logit effects, including a same-category aperture.
    The first three columns are
    Table~\ref{tab:geometry_and_carrier}.
    The same-category column is the new control.
    Each cell is the class-macro paired effect over 600 sources
    and 100 categories; the second line is the 95\% interval.
    A positive value favors the source aperture.
    All seven same-category intervals include zero.
    }
    \label{tab:geometry_same_logit}
    \begin{tabular}{@{}lcccc@{}}
        \toprule
        Model & No aperture & Mean scramble & Wrong geometry & Same category \\
        \midrule
        ResNet-50
            & \shortstack{0.646\\{}[0.485, 0.828]}
            & \shortstack{0.573\\{}[0.420, 0.749]}
            & \shortstack{0.581\\{}[0.422, 0.763]}
            & \shortstack{$-0.035$\\{}[$-0.199$, $0.126$]} \\
        Robust RN50
            & \shortstack{0.293\\{}[0.142, 0.446]}
            & \shortstack{0.138\\{}[0.068, 0.221]}
            & \shortstack{0.233\\{}[0.158, 0.317]}
            & \shortstack{0.018\\{}[$-0.049$, $0.089$]} \\
        ViT-B/16
            & \shortstack{0.940\\{}[0.704, 1.218]}
            & \shortstack{0.803\\{}[0.587, 1.054]}
            & \shortstack{0.848\\{}[0.608, 1.111]}
            & \shortstack{$-0.081$\\{}[$-0.306$, $0.145$]} \\
        ConvNeXt-B
            & \shortstack{1.200\\{}[0.969, 1.458]}
            & \shortstack{1.101\\{}[0.863, 1.363]}
            & \shortstack{1.131\\{}[0.896, 1.388]}
            & \shortstack{$-0.143$\\{}[$-0.373$, $0.077$]} \\
        SIN RN50
            & \shortstack{0.950\\{}[0.679, 1.241]}
            & \shortstack{0.883\\{}[0.664, 1.115]}
            & \shortstack{0.813\\{}[0.568, 1.068]}
            & \shortstack{$-0.049$\\{}[$-0.256$, $0.169$]} \\
        DINOv2-B/14
            & \shortstack{3.364\\{}[2.832, 3.920]}
            & \shortstack{3.184\\{}[2.711, 3.692]}
            & \shortstack{3.253\\{}[2.744, 3.793]}
            & \shortstack{$-0.165$\\{}[$-0.602$, $0.258$]} \\
        CLIP-B/16
            & \shortstack{3.423\\{}[2.978, 3.864]}
            & \shortstack{3.003\\{}[2.624, 3.401]}
            & \shortstack{2.353\\{}[1.942, 2.780]}
            & \shortstack{$-0.164$\\{}[$-0.517$, $0.197$]} \\
        \bottomrule
    \end{tabular}
\end{table*}

\begin{table*}[t]
    \centering
    \small
    \renewcommand{\arraystretch}{1.15}
    \caption{
    Correct-minus-control target log-probability.
    Log-probability is the natural log of the target softmax
    probability. Aggregation matches
    Table~\ref{tab:geometry_same_logit}.
    A positive value favors the source aperture.
    }
    \label{tab:geometry_logprob}
    \begin{tabular}{@{}lcccc@{}}
        \toprule
        Model & No aperture & Mean scramble & Wrong geometry & Same category \\
        \midrule
        ResNet-50
            & \shortstack{0.637\\{}[0.465, 0.827]}
            & \shortstack{0.633\\{}[0.474, 0.816]}
            & \shortstack{0.581\\{}[0.409, 0.771]}
            & \shortstack{$-0.029$\\{}[$-0.205$, $0.136$]} \\
        Robust RN50
            & \shortstack{0.482\\{}[0.327, 0.645]}
            & \shortstack{$-0.003$\\{}[$-0.087$, $0.091$]}
            & \shortstack{0.231\\{}[0.140, 0.326]}
            & \shortstack{0.029\\{}[$-0.046$, $0.105$]} \\
        ViT-B/16
            & \shortstack{0.815\\{}[0.565, 1.095]}
            & \shortstack{0.903\\{}[0.679, 1.152]}
            & \shortstack{0.820\\{}[0.567, 1.099]}
            & \shortstack{$-0.097$\\{}[$-0.322$, $0.129$]} \\
        ConvNeXt-B
            & \shortstack{1.110\\{}[0.850, 1.390]}
            & \shortstack{1.198\\{}[0.928, 1.484]}
            & \shortstack{1.102\\{}[0.844, 1.375]}
            & \shortstack{$-0.160$\\{}[$-0.407$, $0.074$]} \\
        SIN RN50
            & \shortstack{0.119\\{}[$-0.221$, $0.475$]}
            & \shortstack{0.877\\{}[0.602, 1.160]}
            & \shortstack{0.808\\{}[0.504, 1.120]}
            & \shortstack{0.039\\{}[$-0.227$, $0.309$]} \\
        DINOv2-B/14
            & \shortstack{3.851\\{}[3.362, 4.371]}
            & \shortstack{3.508\\{}[3.071, 3.966]}
            & \shortstack{3.296\\{}[2.812, 3.799]}
            & \shortstack{$-0.167$\\{}[$-0.587$, $0.245$]} \\
        CLIP-B/16
            & \shortstack{3.185\\{}[2.764, 3.609]}
            & \shortstack{2.742\\{}[2.370, 3.137]}
            & \shortstack{2.448\\{}[2.051, 2.871]}
            & \shortstack{$-0.126$\\{}[$-0.452$, $0.199$]} \\
        \bottomrule
    \end{tabular}
\end{table*}

\begin{table*}[t]
    \centering
    \small
    \renewcommand{\arraystretch}{1.15}
    \caption{
    Target-rank advantage of the source aperture.
    The effect is the control rank minus the correct rank,
    so a positive value means the target is closer to rank 1
    under the source aperture. Aggregation matches
    Table~\ref{tab:geometry_same_logit}.
    }
    \label{tab:geometry_rank}
    \begin{tabular}{@{}lcccc@{}}
        \toprule
        Model & No aperture & Mean scramble & Wrong geometry & Same category \\
        \midrule
        ResNet-50
            & \shortstack{157.8\\{}[122.3, 195.1]}
            & \shortstack{129.4\\{}[100.6, 161.2]}
            & \shortstack{137.4\\{}[103.3, 173.1]}
            & \shortstack{$-1.1$\\{}[$-27.8$, $25.7$]} \\
        Robust RN50
            & \shortstack{28.4\\{}[9.8, 47.6]}
            & \shortstack{16.9\\{}[8.0, 26.6]}
            & \shortstack{26.4\\{}[15.9, 38.0]}
            & \shortstack{1.9\\{}[$-7.0$, $10.7$]} \\
        ViT-B/16
            & \shortstack{160.6\\{}[118.0, 203.6]}
            & \shortstack{126.2\\{}[93.8, 159.5]}
            & \shortstack{140.5\\{}[103.0, 178.8]}
            & \shortstack{$-16.0$\\{}[$-45.4$, $13.2$]} \\
        ConvNeXt-B
            & \shortstack{213.1\\{}[176.1, 251.9]}
            & \shortstack{194.3\\{}[159.6, 230.8]}
            & \shortstack{202.1\\{}[166.3, 239.8]}
            & \shortstack{$-17.7$\\{}[$-45.3$, $9.3$]} \\
        SIN RN50
            & \shortstack{90.6\\{}[63.9, 117.6]}
            & \shortstack{83.1\\{}[61.1, 106.2]}
            & \shortstack{72.5\\{}[50.2, 95.9]}
            & \shortstack{$-4.2$\\{}[$-23.4$, $14.9$]} \\
        DINOv2-B/14
            & \shortstack{239.7\\{}[200.5, 279.5]}
            & \shortstack{223.1\\{}[188.9, 258.7]}
            & \shortstack{230.4\\{}[192.9, 270.6]}
            & \shortstack{$-7.8$\\{}[$-36.1$, $20.5$]} \\
        CLIP-B/16
            & \shortstack{279.7\\{}[238.6, 321.0]}
            & \shortstack{242.3\\{}[205.5, 279.8]}
            & \shortstack{223.6\\{}[185.5, 264.6]}
            & \shortstack{$-18.6$\\{}[$-47.0$, $9.6$]} \\
        \bottomrule
    \end{tabular}
\end{table*}

All 21 same-category intervals include zero. These readouts
therefore do not separate the source aperture from a
density-and-perimeter-matched aperture taken from another
photograph of the same category. Target rank keeps a positive
interval for no aperture, the scramble mean, and wrong-class
geometry at every checkpoint. Target log-probability does the
same, except for robust RN50 against the scramble mean and
SIN RN50 against no aperture, whose intervals include zero.
For Robust RN50, the log-probability interval also includes
zero in each of the three scramble families.

\subsection{Native Top-1 Effects of Geometry}
\label{app:geometry_top1}

Using the saved native ImageNet-1000 predictions, we
measure original-target top-1 on the same 600 paired B1
references and their geometry controls, without further
inference. Each source contributes one reference per
condition; accuracies average within source and then
across 100 equally weighted categories. The scramble
mean first averages the three scramble conditions within
each source. Paired 95\% CIs use 10,000 hierarchical
class-then-source bootstrap replicates.

\begin{table*}[t]
    \centering
    \footnotesize
    \setlength{\tabcolsep}{4pt}
    \renewcommand{\arraystretch}{1.12}
    \caption{
    Native ImageNet-1000 original-target top-1 (\%) for the
    B1 geometry controls. The paired difference is correct
    geometry minus the within-source mean of three scrambles,
    in percentage points (pp), with an unadjusted 95\% CI.
    All three individual scramble families and wrong-class
    geometry have 0\% original-target top-1 for every model.
    The Shape-only--Final gap comes from the full benchmark,
    not the B1 control subset; it is shown only for scale
    comparison, not as a causal fraction explained by geometry.
    }
    \label{tab:geometry_top1_native}
    \begin{tabular}{@{}lcccccc@{}}
        \toprule
        Model & Correct & No aperture & Scramble mean
        & Wrong geometry & \shortstack{Correct$-$scramble\\(pp; 95\% CI)}
        & \shortstack{Shape$\to$Final\\gap (pp)} \\
        \midrule
        ResNet-50
        & 1.50 & 0.00 & 0.00 & 0.00
        & \shortstack{1.50\\{\scriptsize [0.17, 3.67]}}
        & 30.56 \\
        Robust RN50
        & 0.17 & 0.00 & 0.00 & 0.00
        & \shortstack{0.17\\{\scriptsize [0.00, 1.00]}}
        & 12.27 \\
        ViT-B/16
        & 2.50 & 0.00 & 0.00 & 0.00
        & \shortstack{2.50\\{\scriptsize [0.67, 5.00]}}
        & 29.77 \\
        ConvNeXt-B
        & 3.17 & 0.00 & 0.00 & 0.00
        & \shortstack{3.17\\{\scriptsize [1.17, 6.00]}}
        & 39.04 \\
        SIN RN50
        & 0.50 & 0.00 & 0.00 & 0.00
        & \shortstack{0.50\\{\scriptsize [0.00, 1.67]}}
        & 20.68 \\
        DINOv2-B/14
        & 6.67 & 0.17 & 0.00 & 0.00
        & \shortstack{6.67\\{\scriptsize [3.50, 10.67]}}
        & 49.38 \\
        CLIP-B/16
        & 5.50 & 0.00 & 0.00 & 0.00
        & \shortstack{5.50\\{\scriptsize [2.67, 9.00]}}
        & 34.89 \\
        \bottomrule
    \end{tabular}
\end{table*}

Relative to the scramble mean, correct geometry raises
native top-1 by 0.17--6.67 pp across checkpoints. The
paired 95\% CIs are above zero for five of seven models;
those of Robust RN50 and SIN RN50 include zero. These
exploratory top-1 intervals are not Holm-adjusted and
do not inherit the multiplicity-corrected significance
of the target-logit contrasts. The gains are small
relative to the full Shape-only-to-Final declines, while
showing that geometric interventions can also affect
native top-1 decisions.

\section{Probe Fitting, Transfer, and Conditional Retention}
\label{app:probe_details}
This appendix expands the frozen-feature experiments in
Section~\ref{sec:structural_stability}. We distinguish recognition
within the Shape-only view, transfer of the same fitted readout
to Final, and category decoding by a separately fitted Final-view
readout. All three evaluations use the same 100 target categories.
\subsection{Source-Disjoint Fold Construction}
\label{app:probe_fold_details}
Each category contains six source photographs. Within category,
sources are ordered by the SHA-256 hash of the probe seed and
source identifier, then assigned to six folds.
The probe seed is 20260925.
For test fold $k$, fold $(k+1)\bmod6$ is reserved for validation,
and the remaining four folds provide training sources.
Each rotation therefore contains 400 training sources,
100 validation sources, and 100 test sources.
Every aperture level and render derived from a source inherits
that source's fold.
The six rotations provide one held-out prediction for each
evaluated image. Shape-only evaluation covers 1,800 images;
Final evaluation covers 14,400 renders.
These correspond to the same 600 sources, rather than
1,800 or 14,400 independent photographs.
The splits separate source photographs while retaining the
benchmark categories and shared mirror/carrier asset banks.
They do not evaluate unseen categories or unseen rendering assets.
\subsection{Standardization and Ridge Readouts}
\label{app:ridge_probe_details}
Features are standardized using training-fold statistics only.
For feature dimension $d$, the standardized value is
\begin{equation}
\widetilde{x}_d
=
\frac{x_d-\mu_d^{\mathrm{train}}}
     {\sigma_d^{\mathrm{train}}},
\end{equation}
where training standard deviations below $10^{-6}$ are replaced
by 1. The same training statistics transform validation and
test features.
Let $Z$ contain standardized training features and $Y$ contain
one-hot labels for the 100 categories. The ridge readout~\citep{hoerl1970ridge} solves
\begin{equation}
W_C
=
\arg\min_W
\left\{
\|ZW-Y\|_F^2
+
\frac{1}{C}\|W\|_F^2
\right\}.
\label{eq:ridge_probe_objective}
\end{equation}
The implementation uses
\begin{equation}
W_C
=
\left(Z^\top Z+C^{-1}I\right)^{-1}Z^\top Y.
\end{equation}
No explicit intercept is fitted. Prediction selects the category
with the largest linear score:
\begin{equation}
\widehat{y}_{\mathrm{probe}}(x)
=
\arg\max_j
\left(\widetilde{x}^{\top}W_C\right)_j.
\end{equation}
For the Shape-trained readout, $C$ is selected from
$\{0.01,0.1,1,10\}$ using Shape-only validation accuracy.
Ties favor the smaller $C$.
All three aperture levels are included, giving 1,200 training
and 300 validation images per rotation.
Individual render features are used for fitting; features
are not averaged within source beforehand.
\subsection{Transfer and Final-View Diagnostic}
\label{app:probe_evaluation_details}
We use $S$ for Shape-only and $F$ for Final.
The three principal evaluations are:
\begin{enumerate}
    \item $S\rightarrow S$: fit and select the readout on
    Shape-only training/validation sources, then evaluate
    held-out Shape-only images.
    \item $S\rightarrow F$: apply that same readout and its
    Shape-derived standardization to Final renders from
    the held-out sources.
    \item $F\rightarrow F$: fit a new readout and new
    standardization on Final training features, then evaluate
    held-out Final renders.
\end{enumerate}
The Final-view diagnostic reuses the corresponding rotation's
Shape-selected $C$. It does not select a new $C$ using Final
validation accuracy. Its training fold contains 9,600 renders
from 400 sources; the validation-source fold remains excluded
from fitting.
Reusing $C$ fixes its numerical value, while training-set size
and feature statistics differ between views.
Consequently, this diagnostic establishes decoding under the
specified fitting procedure, rather than comparing separately
optimized heads or equal per-example regularization.
Neither the encoder nor the native ImageNet readout is retrained.
Table~\ref{tab:probe_intervals} provides the complete accuracy
intervals for these evaluations. Each estimate averages
correctness within source and then within category.
Exploratory changes to centering, scale, a training-only linear
map, penalty selection, source weighting, and the number of
Final training renders are reported in
Appendix~\ref{app:readout_sensitivity}.
\begin{table*}[t]
    \centering
    \small
    \renewcommand{\arraystretch}{1.15}
    \caption{
    Source-disjoint probe accuracy and stored 95\% confidence
    intervals, in percent.
    Each cell reports accuracy above its interval.
    All evaluations cover 600 held-out sources across the
    six rotations. Final-trained diagnostics fit fresh weights
    and standardization while reusing Shape-selected $C$.
    }
    \label{tab:probe_intervals}
    \begin{tabular}{@{}lccc@{}}
        \toprule
        Model
        & $S\rightarrow S$
        & $S\rightarrow F$
        & $F\rightarrow F$ \\
        \midrule
        ResNet-50
            & \shortstack{65.22\\{}[58.50, 71.67]}
            & \shortstack{4.12\\{}[2.60, 5.97]}
            & \shortstack{11.92\\{}[9.26, 14.82]} \\
        Robust RN50
            & \shortstack{57.22\\{}[50.11, 64.28]}
            & \shortstack{1.33\\{}[0.44, 2.56]}
            & \shortstack{1.83\\{}[1.00, 2.90]} \\
        ViT-B/16
            & \shortstack{65.72\\{}[59.00, 72.50]}
            & \shortstack{7.12\\{}[5.10, 9.53]}
            & \shortstack{16.42\\{}[12.60, 20.48]} \\
        ConvNeXt-B
            & \shortstack{78.17\\{}[72.72, 83.33]}
            & \shortstack{11.31\\{}[8.31, 14.67]}
            & \shortstack{25.01\\{}[20.28, 29.86]} \\
        SIN RN50
            & \shortstack{55.67\\{}[48.89, 62.50]}
            & \shortstack{3.58\\{}[2.15, 5.35]}
            & \shortstack{5.94\\{}[4.05, 8.13]} \\
        DINOv2-B/14
            & \shortstack{83.17\\{}[78.11, 87.89]}
            & \shortstack{12.21\\{}[8.94, 15.79]}
            & \shortstack{36.35\\{}[30.85, 41.94]} \\
        CLIP-B/16
            & \shortstack{66.83\\{}[60.11, 73.22]}
            & \shortstack{11.17\\{}[8.28, 14.33]}
            & \shortstack{25.03\\{}[20.40, 29.97]} \\
        \bottomrule
    \end{tabular}
\end{table*}

\subsection{Exploratory Readout Sensitivity}
\label{app:readout_sensitivity}
Tables~\ref{tab:readout_shift_sensitivity}
and~\ref{tab:final_head_sensitivity} use the same
source-disjoint folds as Table~\ref{tab:probe_intervals}.
The reproduced Shape$\rightarrow$Final and
Final$\rightarrow$Final point estimates match that table.
The intervals below come from 10,000 class-then-source
bootstrap draws; Table~\ref{tab:probe_intervals} retains
its stored intervals.
These ridge-based checks are separate from the
source-weighted multinomial-logistic DINOv2 head used
for aperture tracking in Table~\ref{tab:final_render_probe}.

Recentering subtracts the Final training mean and adds the
Shape training mean before the unchanged Shape readout.
Scale calibration also divides by the Final training standard
deviation and multiplies by the Shape training standard
deviation. The Shape readout then standardizes with its own
training statistics, so its weights see Final features
standardized by the Final training mean and standard deviation.
Both adjustments use training-fold statistics only.
The linear map is a ridge map with penalty $1$, fit on paired
training-source averages and applied to every held-out Final
render before the unchanged Shape readout.
Final$\rightarrow$Shape applies the reported Final head,
including its Shape-selected penalty, to held-out Shape-only
images.

Recentering accounts for most of the DINOv2 and CLIP movement,
reaching 24.62\% and 17.00\%. ResNet-50 remains at 4.60\%
after recentering and reaches 12.76\% after scale calibration.
The linear map reaches 29.17\% for DINOv2 and 18.90\% for
ConvNeXt-B. Every adjusted Shape$\rightarrow$Final estimate
remains below the corresponding Shape$\rightarrow$Shape
accuracy. Final$\rightarrow$Shape exceeds
Shape$\rightarrow$Final for every checkpoint and remains
below Shape$\rightarrow$Shape.

Selecting $C$ on Final validation accuracy leaves six
reported Final heads within 0.01 percentage points of
Table~\ref{tab:probe_intervals}. DINOv2 moves from 36.35\%
to 37.66\%, and CLIP from 25.03\% to 24.85\%.
The source-weighted ridge head gives every training source
the same total weight and selects
$\lambda\in\{10^{-4},10^{-3},10^{-2},10^{-1}\}$
by source-macro validation accuracy, retaining the smaller
$\lambda$ after a tie. Its penalty is $\lambda\|W\|_F^2$.
DINOv2 reaches 40.56\% under this ridge protocol; all six
folds select $\lambda=0.1$, the largest tested penalty.
The 23.64\% aperture-tracking result
(Table~\ref{tab:final_render_probe}) instead uses multinomial
logistic regression with a separate regularization grid.
The two results therefore differ in fitting protocols
and do not isolate the effect of probe objective.
The three-view head trains on one render from each aperture
level, chosen by sorted mirror and carrier names, selects its
own $C$ on those validation renders, and is scored on every
held-out Final render. DINOv2 falls from 36.35\% to 24.25\%.
Its source-weighted ridge counterpart remains close.
\begin{table*}[t]
    \centering
    \small
    \renewcommand{\arraystretch}{1.15}
    \caption{
    Exploratory Shape-readout sensitivity.
    Accuracy (\%) is shown above its 95\% interval.
    Recentering, scale calibration, and the linear map keep
    the Shape weights. Final$\rightarrow$Shape applies the
    reported Final head to Shape-only images.
    }
    \label{tab:readout_shift_sensitivity}
    \begin{tabular}{@{}lcccc@{}}
        \toprule
        Model
        & Recentering
        & Scale-calibrated
        & Linear map
        & Final$\rightarrow$Shape \\
        \midrule
        ResNet-50
            & \shortstack{4.60\\{}[3.08, 6.51]}
            & \shortstack{12.76\\{}[10.17, 15.60]}
            & \shortstack{11.08\\{}[8.60, 13.81]}
            & \shortstack{36.67\\{}[30.11, 43.50]} \\
        Robust RN50
            & \shortstack{1.32\\{}[0.54, 2.42]}
            & \shortstack{1.97\\{}[1.02, 3.14]}
            & \shortstack{2.00\\{}[1.16, 3.01]}
            & \shortstack{2.11\\{}[0.67, 4.17]} \\
        ViT-B/16
            & \shortstack{8.11\\{}[5.97, 10.54]}
            & \shortstack{8.17\\{}[6.02, 10.60]}
            & \shortstack{11.55\\{}[8.82, 14.56]}
            & \shortstack{41.44\\{}[34.22, 48.94]} \\
        ConvNeXt-B
            & \shortstack{13.39\\{}[10.38, 16.83]}
            & \shortstack{13.72\\{}[10.63, 17.10]}
            & \shortstack{18.90\\{}[15.17, 22.86]}
            & \shortstack{57.83\\{}[50.11, 65.28]} \\
        SIN RN50
            & \shortstack{4.71\\{}[3.17, 6.56]}
            & \shortstack{5.30\\{}[3.58, 7.40]}
            & \shortstack{5.54\\{}[3.83, 7.59]}
            & \shortstack{17.56\\{}[12.44, 23.33]} \\
        DINOv2-B/14
            & \shortstack{24.62\\{}[20.44, 29.09]}
            & \shortstack{25.49\\{}[21.15, 29.99]}
            & \shortstack{29.17\\{}[24.64, 33.96]}
            & \shortstack{37.39\\{}[30.61, 44.39]} \\
        CLIP-B/16
            & \shortstack{17.00\\{}[13.51, 20.70]}
            & \shortstack{17.31\\{}[13.91, 21.08]}
            & \shortstack{16.61\\{}[12.42, 21.21]}
            & \shortstack{44.11\\{}[36.72, 51.67]} \\
        \bottomrule
    \end{tabular}
\end{table*}
\begin{table*}[t]
    \centering
    \small
    \renewcommand{\arraystretch}{1.15}
    \caption{
    Exploratory Final-head sensitivity.
    Accuracy (\%) is shown above its 95\% interval.
    Own-penalty selection uses Final validation accuracy.
    Source-weighted rows give each training source equal total
    weight. Three-view training uses one render from each
    aperture level and is scored on every held-out Final render.
    }
    \label{tab:final_head_sensitivity}
    \begin{tabular}{@{}lcccc@{}}
        \toprule
        Model
        & Own penalty
        & Source-weighted
        & Three-view
        & Three-view, weighted \\
        \midrule
        ResNet-50
            & \shortstack{11.92\\{}[9.33, 14.90]}
            & \shortstack{12.68\\{}[9.88, 15.74]}
            & \shortstack{5.77\\{}[4.15, 7.71]}
            & \shortstack{5.92\\{}[4.26, 7.93]} \\
        Robust RN50
            & \shortstack{1.84\\{}[1.02, 2.87]}
            & \shortstack{1.83\\{}[1.02, 2.88]}
            & \shortstack{1.12\\{}[0.49, 2.03]}
            & \shortstack{0.72\\{}[0.28, 1.38]} \\
        ViT-B/16
            & \shortstack{16.42\\{}[12.58, 20.60]}
            & \shortstack{16.63\\{}[12.78, 20.94]}
            & \shortstack{7.63\\{}[5.51, 9.96]}
            & \shortstack{7.84\\{}[5.69, 10.25]} \\
        ConvNeXt-B
            & \shortstack{25.01\\{}[20.29, 29.94]}
            & \shortstack{25.25\\{}[20.40, 30.39]}
            & \shortstack{12.92\\{}[9.78, 16.48]}
            & \shortstack{13.17\\{}[9.92, 16.69]} \\
        SIN RN50
            & \shortstack{5.94\\{}[4.03, 8.13]}
            & \shortstack{6.94\\{}[4.65, 9.59]}
            & \shortstack{3.14\\{}[2.05, 4.44]}
            & \shortstack{2.82\\{}[1.86, 3.98]} \\
        DINOv2-B/14
            & \shortstack{37.66\\{}[32.28, 43.31]}
            & \shortstack{40.56\\{}[34.69, 46.32]}
            & \shortstack{24.25\\{}[19.34, 29.47]}
            & \shortstack{24.46\\{}[19.45, 29.75]} \\
        CLIP-B/16
            & \shortstack{24.85\\{}[20.14, 29.79]}
            & \shortstack{25.02\\{}[20.24, 30.01]}
            & \shortstack{16.37\\{}[12.63, 20.41]}
            & \shortstack{16.60\\{}[12.76, 20.69]} \\
        \bottomrule
    \end{tabular}
\end{table*}
\begin{figure}[H]
    \centering
    \includegraphics[width=\columnwidth]
    {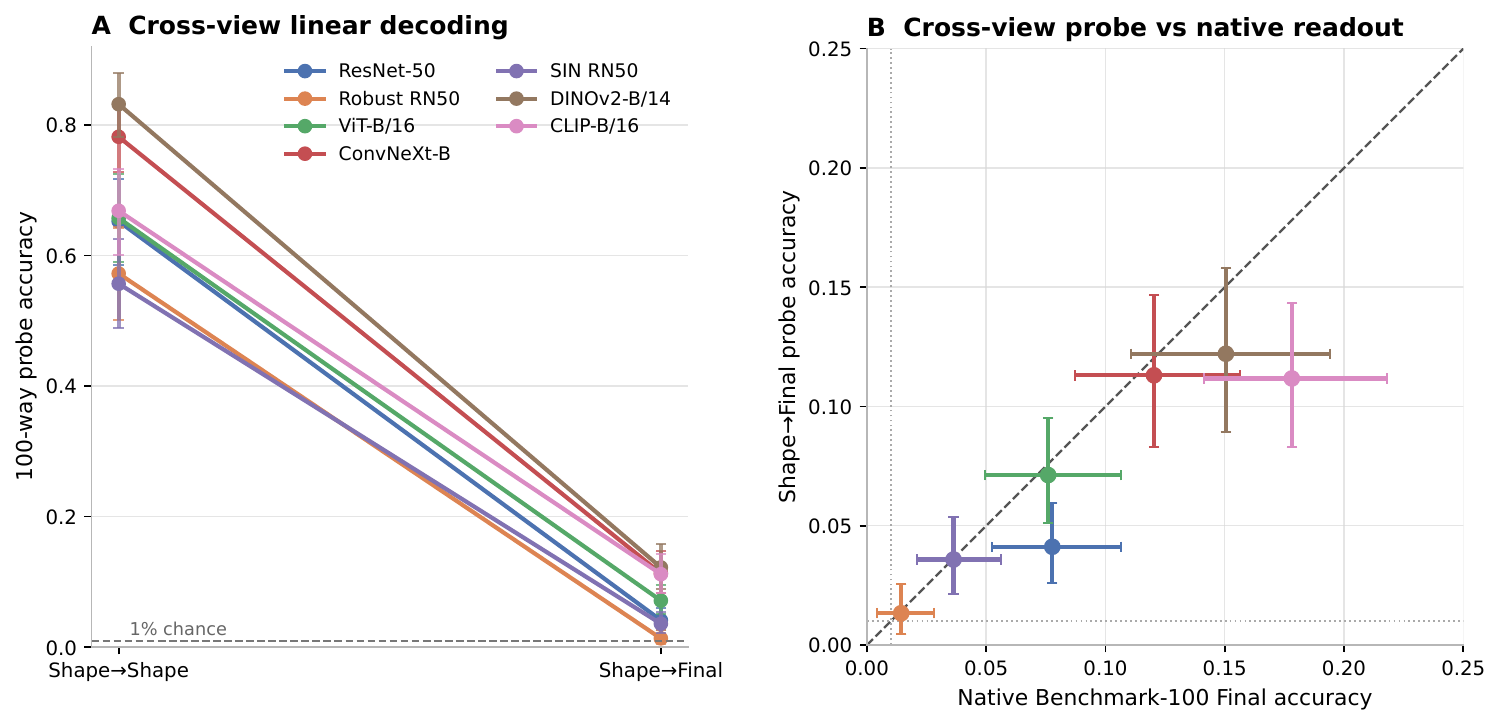}
    \caption{
    \textbf{Cross-view linear readout transfer.}
    Left: Shape-trained probe accuracy on held-out
    Shape-only and Final images.
    Right: transferred-probe versus native Benchmark-100
    accuracy on identical Final renders; the diagonal
    marks equality. Error bars show 95\% CIs;
    the 1\% line marks 100-way chance.
    Paired gaps and the render-level Final$\to$Final
    head appear in Table~\ref{tab:structural_transfer}.
    }
    \label{fig:structural_transfer}
\end{figure}

\begin{table*}[t]
    \centering
    \small
    \caption{
    \textbf{Exploratory readout sensitivity.}
    Source-disjoint 100-class accuracy (\%).
    Shape$\to$Final and Final$\to$Final reproduce
    Table~\ref{tab:structural_transfer}: the latter is
    the render-level head that reuses the Shape-selected
    penalty.
    Scale-calibrated Shape$\to$Final matches each Final
    coordinate to the training-source Shape mean and
    standard deviation, then applies the unchanged Shape
    readout.
    Final$\to$Shape applies that reported Final head to
    held-out Shape-only images.
    Three-view trains on one render from each aperture
    level, selects its own penalty, and is scored on
    every held-out Final render.
    Bold indicates column maxima.
    }
    \label{tab:readout_sensitivity}
    \setlength{\tabcolsep}{3pt}
    \renewcommand{\arraystretch}{1.10}
    \begin{tabular*}{0.92\textwidth}{
        @{\extracolsep{\fill}}
        lccccc
        @{}
    }
        \toprule
        \textbf{Model}
        & \textbf{Shape$\to$Final}
        & \textbf{Scale-calibrated}
        & \textbf{Final$\to$Shape}
        & \textbf{Final$\to$Final}
        & \textbf{Three-view} \\
        \midrule
        ResNet-50
        & 4.12 & 12.76 & 36.67 & 11.92 & 5.77 \\
        Robust RN50
        & 1.33 & 1.97 & 2.11 & 1.83 & 1.12 \\
        ViT-B/16
        & 7.12 & 8.17 & 41.44 & 16.42 & 7.63 \\
        ConvNeXt-B
        & 11.31 & 13.72 & \textbf{57.83} & 25.01 & 12.92 \\
        SIN RN50
        & 3.58 & 5.30 & 17.56 & 5.94 & 3.14 \\
        DINOv2-B/14
        & \textbf{12.21} & \textbf{25.49} & 37.39
        & \textbf{36.35} & \textbf{24.25} \\
        CLIP-B/16
        & 11.17 & 17.31 & 44.11 & 25.03 & 16.37 \\
        \bottomrule
    \end{tabular*}
\end{table*}

\paragraph{Sensitivity of the ridge readout.}
Table~\ref{tab:readout_sensitivity} reproduces the
reported Shape$\to$Final and Final$\to$Final estimates
and adds three exploratory checks.
Matching each Final coordinate to the training-source
Shape mean and standard deviation raises
Shape$\to$Final, including ResNet-50 from 4.12\% to
12.76\%, DINOv2 from 12.21\% to 25.49\%, and CLIP from
11.17\% to 17.31\%. The calibrated readout remains below
Shape-only accuracy.
A Final-trained readout evaluated on Shape-only exceeds
Shape$\to$Final for every checkpoint and remains below
Shape$\to$Shape, from 2.11\% for Robust RN50 to 57.83\%
for ConvNeXt-B.
A Final head trained on one render from each aperture
level and scored on every held-out Final render lowers
Final$\to$Final, including DINOv2 from 36.35\% to
24.25\%.

\subsection{Paired Comparison with the Native Readout}
\label{app:probe_native_comparison}
The native comparison restricts saved ImageNet scores to the
same 100 benchmark labels
(Appendix~\ref{app:native_readout_details}).
For each Final render, the comparison subtracts native
correctness from Shape-trained probe correctness.
These paired differences are aggregated by source and category
before uncertainty estimation.
Table~\ref{tab:structural_transfer} reports all seven
probe-minus-native gaps and their intervals.
Every point estimate is negative. The ResNet-50 and CLIP-B/16
intervals lie below zero, while the other five include zero.
Intervals containing zero do not establish equivalence,
and these comparisons do not support a consistent advantage
for the transferred Shape-trained readout.

\subsection{Conditional Recognition Retention}
\label{app:conditional_retention_details}
Aggregate accuracies do not identify which initially recognized
Shape-only images remain recognized under composition.
We therefore pair held-out predictions by checkpoint,
source identifier, and aperture level.
Each Shape-only prediction is paired with eight Final renders:
two mirrors crossed with four carriers at that same level.
Both predictions use the Shape-trained readout associated
with the source's held-out fold.
Let $h_{i,b}$ indicate correct Shape-only recognition for
source $i$ and aperture level $b$, and let $f_{i,b,r}$ indicate
correct recognition of its Final render $r$.
Conditional retention is
\begin{equation}
\widehat{\rho}
=
\frac{
\sum_{i,b}\sum_{r=1}^{8}h_{i,b}f_{i,b,r}
}{
\sum_{i,b}\sum_{r=1}^{8}h_{i,b}
}.
\label{eq:paired_conditional_retention}
\end{equation}
The denominator counts paired Final instances whose associated
Shape-only image was correctly recognized.
Each recognized Shape-only image contributes eight eligible
instances; these are repeated observations of the same source.
Final successes following an incorrect Shape-only prediction
are counted separately:
\begin{equation}
N_{\mathrm{new}}
=
\sum_{i,b}\sum_{r=1}^{8}
(1-h_{i,b})f_{i,b,r}.
\end{equation}
Because such successes can occur, dividing marginal
$S\rightarrow F$ accuracy by marginal $S\rightarrow S$
accuracy does not yield conditional retention.
Table~\ref{tab:conditional_retention} reports the paired
numerators, denominators, retention estimates, and newly
correct instances. The conditional estimate weights eligible
instances; it is not an average of category-specific
conditional ratios.
\begin{table*}[t]
    \centering
    \small
    \caption{
    Recognition on Final conditional on correct Shape-only
    recognition by the held-out Shape-trained readout.
    Counts refer to paired render instances among 14,400 Final
    renders per checkpoint, not independent photographs.
    Joint correct is the numerator and eligible is the
    denominator in Equation~\ref{eq:paired_conditional_retention}.
    Newly correct counts Final successes following Shape-only
    failures. Retention and interval endpoints are percentages.
    }
    \label{tab:conditional_retention}
    \begin{tabular}{@{}lrrrcr@{}}
        \toprule
        Model
        & Joint correct
        & Eligible
        & Retention
        & 95\% CI
        & Newly correct \\
        \midrule
        ResNet-50
            & 507 & 9,392 & 5.40 & [3.53, 7.48] & 86 \\
        Robust RN50
            & 116 & 8,240 & 1.41 & [0.31, 2.84] & 76 \\
        ViT-B/16
            & 938 & 9,464 & 9.91 & [7.32, 12.69] & 88 \\
        ConvNeXt-B
            & 1,556 & 11,256 & 13.82 & [10.53, 17.46] & 72 \\
        SIN RN50
            & 426 & 8,016 & 5.31 & [3.31, 7.55] & 90 \\
        DINOv2-B/14
            & 1,657 & 11,976 & 13.84 & [10.64, 17.31] & 101 \\
        CLIP-B/16
            & 1,399 & 9,624 & 14.54 & [11.13, 18.10] & 210 \\
        \bottomrule
    \end{tabular}
\end{table*}
This supplementary ratio analysis uses 10,000 class-then-source
bootstrap draws with seed 20261001. Each draw samples 100
categories with replacement and six sources within each sampled
category occurrence. Repeated occurrences receive fresh source
draws. Source-level numerator and denominator contributions
remain paired, and their ratio is recomputed for each draw.
The 2.5th and 97.5th percentiles provide the interval.
These estimates show limited persistence of correct recognition
by the Shape-trained readout among initially recognized
instances. They do not establish that category information
has disappeared from the frozen representation.

\subsection{Scope of the Label-Remapping Control}
\label{app:probe_label_remapping}
The local analysis implementation also contains a class-level
training-label permutation. Within a rotation, every training
example of a category receives the same permuted category
identity, while validation and test labels retain their original
identities.
This transformation preserves category grouping in the training
data and changes its alignment with the evaluation labels.
It is therefore distinct from independently shuffling labels
across sources. We exclude it as evidence for an independent
source-label-shuffle null, and no main claim depends on its
exported accuracy.

\subsection{Numerical Exports and Interpretation}
\label{app:probe_exports}
The selected probe export,
\texttt{probe\_full\_selected.csv}, retains evaluation identities,
accuracy intervals, native comparisons, paired gaps, test-source
and render counts, and fold-specific selected $C$ values.
The sensitivity estimates are retained in
\texttt{item11\_probe\_sensitivity.csv}.
Auxiliary geometry-control evaluations retain their metric
identities; replacement-category prediction rates are not
relabeled as original-target accuracy.
The paired retention counts and uncertainty are provided in
\texttt{conditional\_retention.csv}.
They are computed from existing held-out predictions and do
not require new encoder inference or probe fitting.
Together, the experiments establish poor transfer of the
Shape-trained decision rule and category information recoverable
by a separately fitted Final-view rule within the benchmark.
They do not establish that the recovered category information
is specifically geometric or that the results generalize
beyond the shared categories and rendering assets.

\subsection{Aperture-Tracking Probe and Held-Out Assets}
\label{app:aperture_tracking_details}

The DINOv2 head in Section~\ref{sec:structural_stability}
is a separate fit on every Final render of the training
sources.
Each source receives the same total weight, with weights
rescaled to mean one. Standardization is the
source-weighted mean and variance of that fold's training
rows. $C$ is chosen by source-macro validation accuracy,
ties keep the smaller value, and the scaler and weights
are frozen before testing. The frozen head is then applied
to held-out Final renders, correct geometry, a same-class
aperture, a different-class aperture, the three scrambles,
no aperture, and a mirror--carrier identifier. A
source-and-asset-disjoint split removes held-out mirror
and carrier assets from training; its test set contains
the 32 sources whose reference uses one of those assets.

\paragraph{Aperture-category tracking.}
The 36.35\% DINOv2 entry in
Table~\ref{tab:structural_transfer} is the render-level
fit that reuses the Shape-selected penalty.
The separate equal-source-weight DINOv2 head
reaches 23.64\% on held-out Final images
(95\% CI: $[18.93,28.58]$) and 25.00\% on correct
geometry (95\% CI: $[18.83,31.33]$).
A same-class aperture scores 42.67\%
(95\% CI: $[35.50,49.83]$).
A different-class aperture scores 0\% on the target
(95\% CI: $[0.00,0.00]$) and 53.17\% on the donor class
(95\% CI: $[46.83,59.50]$).
Correct geometry exceeds the mean of the three scrambles
by 24.11 percentage points
(95\% CI: $[18.17,30.45]$) and no aperture by 24.33
(95\% CI: $[18.17,31.17]$).
Tile, row-band, and column-band scrambles score
0.50\%, 0.83\%, and 1.33\%; no aperture scores 0.67\%;
and a mirror--carrier identifier scores 0\%.
On the scramble subsets that pass the low-level geometry
check, top-1 is 0.00\%, 1.11\%, and 1.73\%, with pass
rates 38.17\%, 67.33\%, and 71.67\%.
In the source-and-asset-disjoint split, 32 sources from
32 classes score 15.76\% on Final
(95\% CI: $[8.33,24.22]$) and 12.50\% on correct geometry
(95\% CI: $[3.13,25.00]$).
Correct geometry exceeds the scramble mean by 10.42
percentage points (95\% CI: $[1.04,21.88]$) and no
aperture by 12.50 (95\% CI: $[3.13,25.00]$).
Donor-class accuracy is 43.75\%
(95\% CI: $[28.13,62.50]$), and target accuracy on
wrong-class geometry is 0\%.
This limited evaluation includes one test source per
category. Its wide intervals and lack of within-category
replication restrict conclusions about generalization
to unseen rendering assets.

\paragraph{Replacement-aperture provenance.}
Evaluation targets are held out from probe training,
but replacement apertures are selected independently
of the probe's source folds. In the six-fold evaluation,
468 of 600 same-class donors originate from training
sources and 132 from validation sources. For wrong-class
replacements, 397 originate from training, 101 from
validation, and 102 from test sources.
In the 32-source asset-disjoint evaluation, 22 same-class
and 25 wrong-class donors originate from training sources
of the corresponding head. These replacement comparisons
therefore characterize category sensitivity under the
evaluated donor assignments, rather than generalization
to entirely unseen replacement apertures. The correct-
geometry, scramble, and no-aperture comparisons do not
use these replacement donors.


\section{Human Validation: Protocol and Results}
\label{app:human_validation}

This appendix presents the human-recognition evaluation
summarized in Section~\ref{sec:human_recognition},
including the survey design, response coverage, scoring,
statistical analysis, and variation in stimulus difficulty.
The study examines whether respondents can identify
source-object categories from Semantic Apertures
composites under a multiple-choice recognition task.
The 500 Final images and ten options are shared with
the paired VLM evaluation in
Appendix~\ref{app:extended_vlm}.
Prompts, the number of respondents, and aggregation
still differ. Native ImageNet-1000 and Benchmark-100
recognition use different candidate sets.

\subsection{Survey Design and Response Coverage}
\label{app:human_coverage}

\paragraph{Evaluation design.}
The human study evaluated 500 Final-view stimuli
representing 100 target categories, with five stimuli
per category. The evaluation was organized into ten
survey forms containing 50 questions each.
Respondents were asked to identify the object
represented in each image by selecting a category
from the displayed multiple-choice alternatives.

\paragraph{Stimulus selection.}
The 500 stimuli were drawn from the Final benchmark,
with five images per target category. The selected set
contains 167 B0, 167 B1, and 166 B2 images; mirror
and carrier assignments also vary across images.
Selection preceded construction of the model-informed
answer options, and model predictions were not used
to select the stimuli. Class and aperture-level balance
do not imply random sampling from all 14,400 composites.

\paragraph{Ethics and consent.}
The human evaluation involved a low-risk visual recognition
survey using synthetic images. Participants were informed
of the research purpose and consented to the use of their
responses. No formal institutional ethics review was
conducted. Responses were analyzed using anonymized
records without linking identities across survey forms.

\paragraph{Model-informed answer options.}
Three ImageNet-1K-pretrained classifiers---ConvNeXt-Tiny,
Swin-T, and MaxViT-T---were evaluated on clean
source-object crops with the background replaced
by neutral gray. Crops were centered on square
canvases and processed using each model's
pretrained image transforms. Neither the Final
composites nor their fogged carrier appearance
were used for candidate selection.

Each classifier ranked the 100 benchmark categories.
The union of their top-ten predictions was combined
using the mean reciprocal rank across the three
complete category rankings. Predefined ambiguous
label pairs were excluded when selecting alternatives
to the ground-truth category. The ten highest-ranked
eligible categories were retained, with the correct
source-object category explicitly included if absent.
Thus, each question provided the correct category
and nine model-informed alternatives.

The collected data comprise 51 complete survey
submissions and 2,550 individual judgments,
of which 1,774 were correct.
Each form received between three and eleven
complete submissions, corresponding to 3--11
judgments per stimulus. Every analyzed submission
contained responses to all 50 questions in its form.

Submissions were represented by anonymized
record identifiers. Because participant identities
were not linked across forms, the 51 submissions
are treated as response records rather than
verified distinct individuals.

\begin{figure*}[p]
    \centering
    \includegraphics[
        width=\textwidth,
        height=0.80\textheight,
        keepaspectratio
    ]{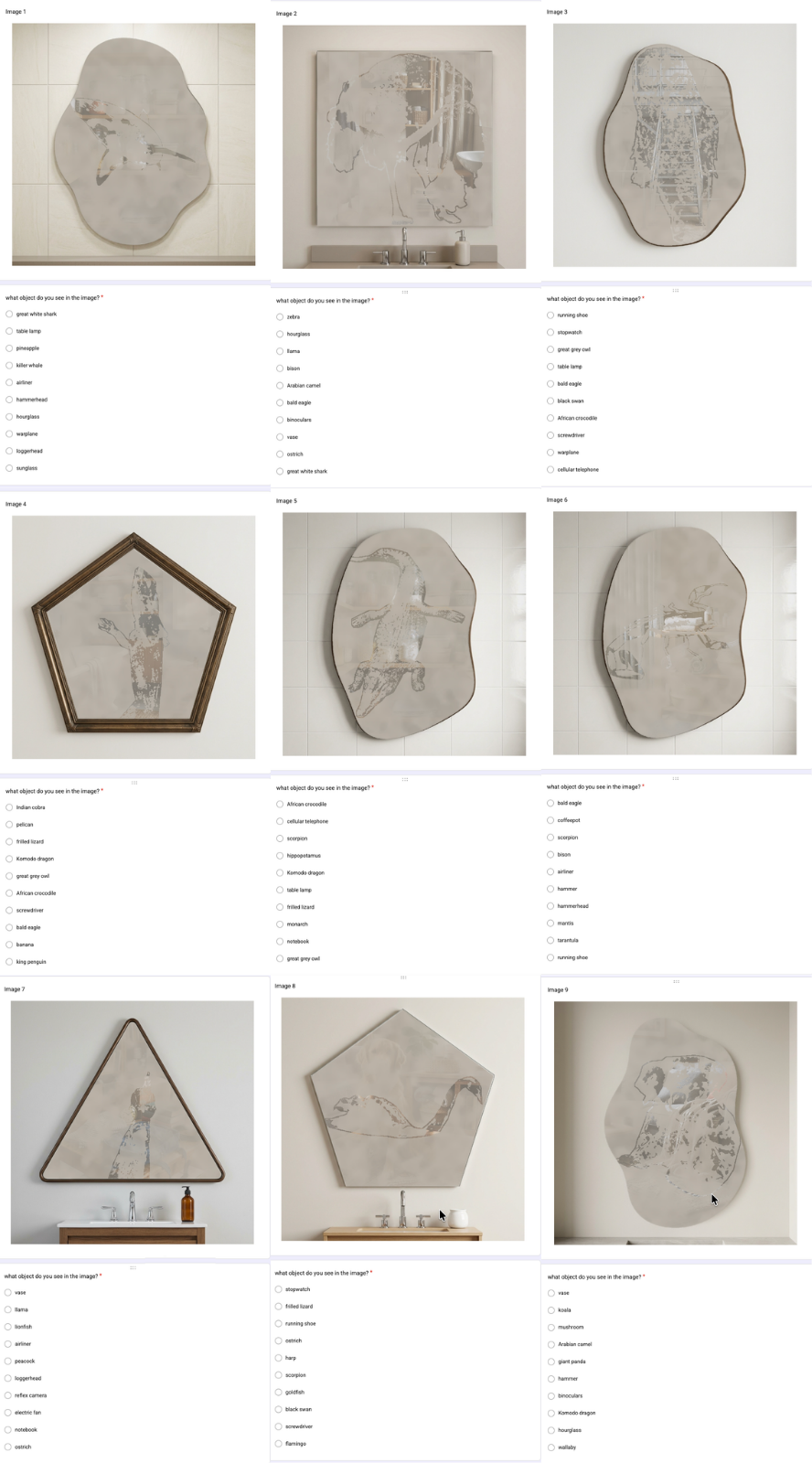}
    \caption{
    \textbf{Example items from the human-recognition survey.}
    Nine representative questions illustrate the image
    presentation, question wording, and multiple-choice
    response format. Each displayed example provides
    ten candidate labels and permits one selection.
    }
    \label{fig:human_survey_examples}
\end{figure*}

\paragraph{Survey presentation.}
Figure~\ref{fig:human_survey_examples} illustrates
nine representative questions from the administered
survey interface. The displayed examples include
a Semantic Apertures image, the question
``What object do you see in the image?'',
and ten candidate category labels presented
as single-choice options.

\paragraph{Response coverage.}
Table~\ref{tab:human_per_form} summarizes the response
coverage and accuracy of each form. Since every
submission contains 50 responses, all questions
within a form have equal response coverage.
However, the number of submissions differs across
forms. We therefore use stimulus-level aggregation
as the primary accuracy measure, giving equal
weight to each evaluated image.

\begin{table}[t]
    \centering
    \footnotesize
    \setlength{\tabcolsep}{3pt}
    \renewcommand{\arraystretch}{1.08}
    \caption{
    \textbf{Human survey coverage and recognition.}
    Each form contains 50 stimuli. Records are complete
    submissions, not verified unique participants.
    Accuracies and 95\% CIs are percentages.
    The total reports image-macro accuracy over 500 images.
    The matched VLM comparison uses 498 images
    (Figure~\ref{fig:human_vlm_matched}).
    Survey examples and recognition distributions appear
    in Figures~\ref{fig:human_survey_examples}
    and~\ref{fig:human_recognition}.
    }
    \label{tab:human_per_form}

    \resizebox{\columnwidth}{!}{%
    \begin{tabular}{@{}rrrrrc@{}}
        \toprule
        Form & Images & Records & Judgments
        & Accuracy & 95\% CI \\
        \midrule
        1  & 50 & 3  & 150 & 66.00 & [52.00, 78.67] \\
        2  & 50 & 4  & 200 & 66.50 & [49.00, 82.00] \\
        3  & 50 & 3  & 150 & 60.00 & [44.00, 76.00] \\
        4  & 50 & 8  & 400 & 63.75 & [50.25, 76.50] \\
        5  & 50 & 7  & 350 & 77.71 & [66.29, 88.00] \\
        6  & 50 & 3  & 150 & 73.33 & [59.33, 86.00] \\
        7  & 50 & 4  & 200 & 77.00 & [66.00, 86.50] \\
        8  & 50 & 4  & 200 & 81.00 & [71.00, 90.00] \\
        9  & 50 & 4  & 200 & 66.50 & [45.50, 85.50] \\
        10 & 50 & 11 & 550 & 66.55 & [54.91, 77.64] \\
        \midrule
        Total & 500 & 51 & 2,550
              & 69.83 & [65.41, 74.10] \\
        \bottomrule
    \end{tabular}%
    }
\end{table}

\subsection{Answer Mapping and Recognition Scoring}
\label{app:human_scoring}

\paragraph{Answer-key alignment.}
Ground-truth labels were associated with survey
questions using their form identifiers and positions.
The answer-key workbook contains ten columns,
\texttt{table\_0} through \texttt{table\_9},
corresponding to Forms 1--10. Within each column,
the 50 entries follow the administered question order.

Responses were aligned with these entries by form
and question position. Their original column order
was preserved to distinguish questions sharing
identical headings. The response file named
\texttt{Mirror Images 55.csv} corresponds to Form 5.
All analyzed forms contained 50 nonblank responses
per submission and complete corresponding
ground-truth labels.

\paragraph{Response normalization.}
Responses and reference category labels were
normalized using Unicode NFC, surrounding-whitespace
removal, and case folding. A prediction was scored
as correct only when its normalized text exactly
matched the corresponding target label.
No synonym-based matching, fuzzy matching,
or partial credit was applied.

\paragraph{Primary recognition measure.}
Let $k_i$ denote the number of correct judgments
for stimulus $i$, and let $n_i$ denote its total
number of judgments. Image-macro accuracy is

\begin{equation}
    \widehat{a}_{\mathrm{image}}
    =
    \frac{1}{500}
    \sum_{i=1}^{500}
    \frac{k_i}{n_i}.
    \label{eq:human_appendix_macro}
\end{equation}

This measure assigns equal weight to each
evaluated stimulus, independently of its
response count. The resulting accuracy is
69.83\%. Since each of the 100 target
categories contributes five stimuli,
image-macro and category-macro accuracy
coincide for this evaluation.

For completeness, judgment-pooled accuracy is

\begin{equation}
    \widehat{a}_{\mathrm{pooled}}
    =
    \frac{\sum_i k_i}{\sum_i n_i}
    =
    \frac{1{,}774}{2{,}550}
    =
    69.57\%.
    \label{eq:human_pooled_accuracy}
\end{equation}

The difference reflects unequal response
coverage across survey forms. We report
image-macro accuracy as the primary result
throughout the paper.

\subsection{Uncertainty and Recognition Distributions}
\label{app:human_uncertainty}

\paragraph{Confidence intervals.}
We estimate uncertainty using 10,000 crossed
response-record and stimulus bootstrap replicates,
with random seed 20261001.

The ten forms remain fixed. Within each form,
complete response records and the 50 question
positions are independently resampled with
replacement. A sampled record contributes its
responses across the selected questions, preserving
the dependence among judgments from the same
submission.

For each replicate, recognition accuracy is
calculated separately for each form and then
averaged equally across the ten forms.
This reproduces the weighting of the primary
image-macro estimator.

The 2.5th and 97.5th percentiles of the
bootstrap distribution provide the reported
95\% confidence interval of
$[65.41,74.10]\%$.
Form-specific intervals appear in
Table~\ref{tab:human_per_form}.

These intervals describe uncertainty under
the observed survey design and its resampling
assumptions. They do not account for possible
dependence between submissions made by the
same individual across different forms.
The resampling unit for survey items is the
administered question, rather than the
underlying source photograph.

\begin{figure}[H]
    \centering
    \includegraphics[width=\columnwidth]
    {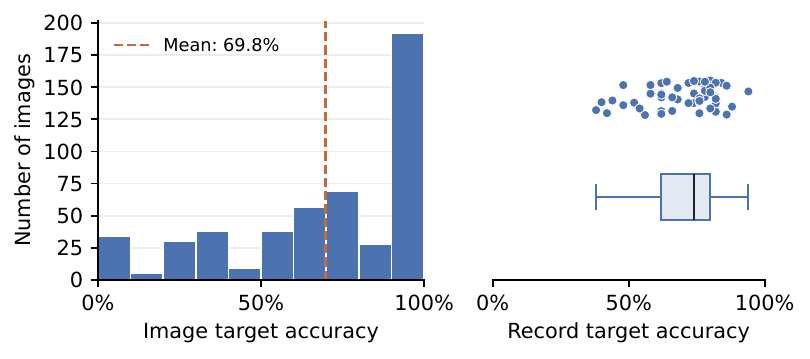}
    \caption{
    \textbf{Human target recognition and stimulus difficulty.}
    Left: per-image recognition accuracy across 500
    Final stimuli; the dashed line indicates the
    69.83\% image-macro mean (95\% CI:
    65.41--74.10\%). Each stimulus received
    3--11 judgments. Right: accuracy across
    51 complete survey submissions, with
    50 judgments per record. Submission records
    do not necessarily represent unique individuals.
    }
    \label{fig:human_recognition}
\end{figure}

\paragraph{Distribution of recognition outcomes.}
Figure~\ref{fig:human_recognition} summarizes
recognition across individual stimuli and complete
survey submissions.

Recognition varies across the 500 stimuli:
188 (37.6\%) received only correct judgments,
31 (6.2\%) received none, and the remaining
281 (56.2\%) had intermediate accuracy.

These outcomes indicate considerable variation
in stimulus difficulty under the administered
task. Because each image received only 3--11
judgments, unanimous outcomes do not establish
that an image is universally easy or difficult.

\paragraph{Zero-correct stimuli.}
All 31 zero-correct images represent failures under
the administered ten-choice task, not proof of
unrecognizability. Each had between 3 and 11 judgments,
but response coverage varies by survey form; zero
correct responses from three judgments carry less
evidence of difficulty than zero from eleven. The
aggregate count alone cannot reveal whether errors
cluster by target category, aperture level, carrier,
or answer-option confusions, nor can it isolate effects
of fog, contrast, or aperture visibility.

Accuracy across complete response records
ranges from 38\% to 94\%, with a median
of 74\%. This variation may reflect differences
in the evaluated questions, respondent
performance, or survey-form composition.
The available design does not separately
identify these contributions.

The image-level and record-level distributions
characterize complementary aspects of the
observed responses. Neither distribution is
an inter-rater agreement statistic or a
substitute for uncertainty estimation of
the overall recognition rate.

\subsection{Human Recognizability and Interpretation}
\label{app:human_validation_interpretation}
\label{app:human_vlm_protocols}

\paragraph{Recognition of target categories.}
The human study provides evidence that a
substantial proportion of the evaluated
Semantic Apertures composites permit
recognition of their source-object categories.

The 69.83\% image-macro accuracy demonstrates
task-specific recognizability under the ten-choice
setting with model-informed distractors.
At the same time, the per-image distribution
indicates that recognizability is not uniform
across the evaluated stimuli.

The result concerns the sampled 500 images
and does not establish human recognition
accuracy across all 14,400 benchmark
composites or every rendering configuration.
No matched human Shape-only baseline was collected,
so the study does not estimate a human Shape-to-Final
decline or attribute it to composition.

\paragraph{Interpretation of visual evidence.}
A correct category response does not reveal
which visual features were used to make
the decision. Respondents may rely on
aperture boundaries, internal structural
features, contextual cues, residual
appearance, or combinations of these signals.

Accordingly, the human results support
the presence of recognizable target-related
information without establishing that
recognition depends exclusively on aperture
geometry. They also do not isolate the
contribution of transmitted carrier appearance.

The structural properties of the generated
apertures and the effects of controlled
geometry interventions are examined
separately in the benchmark construction
and model analyses.

\paragraph{Relationship to model evaluation.}
Native ImageNet-1000 and Benchmark-100
recognition use different candidate sets,
so those accuracies are not estimates on
the human task.

\begin{figure}[H]
    \centering
    \includegraphics[width=\columnwidth]
    {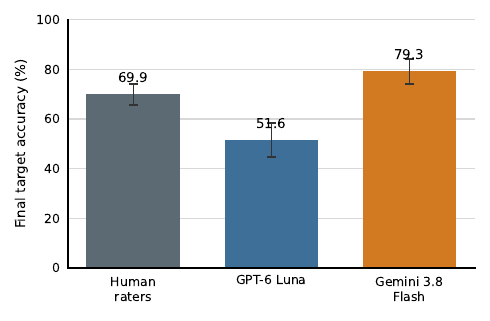}
    \caption{%
    \textbf{Final-target recognition on the same images and the same ten options.}
    The human bar is the category macro of per-image rater accuracy.
    Each VLM bar is one scored request per image.
    All three estimates use the same 498 images.
    Error bars are 95\% hierarchical bootstrap intervals.
    The human interval resamples images within categories;
    the published human interval also resamples respondents and uses all 500 images.}
    \label{fig:human_vlm_matched}
\end{figure}

The paired VLM evaluation uses the same
500 Final images and the same ten options.
On the 498 images scored by both endpoints,
category-macro Final-target accuracy is
51.6\% for \texttt{gpt-6-luna} and 79.3\%
for \texttt{gemini-3.8-flash}, against a
human rate of 69.9\% on those images
(Figure~\ref{fig:human_vlm_matched}).
Gemini lies above the raters and GPT lies
below them. Prompts, the number of
respondents, and aggregation still differ:
human accuracy averages the raters on each
image, and each endpoint contributes one
request.

The same ten options, applied to the seven
encoders' stored ImageNet-1000 logits on
all 500 Final images, give 16.6\%
(ResNet-50), 3.8\% (Robust RN50), 14.8\%
(ViT-B/16), 19.6\% (ConvNeXt-B), 7.2\%
(SIN RN50), 26.8\% (DINOv2-B/14), and
30.2\% (CLIP-B/16).

\subsection{Provenance and Evaluation Scope}
\label{app:human_protocol_scope}

\paragraph{Response and stimulus documentation.}
Survey responses were aligned deterministically
with ground-truth category labels using form
identifiers and question positions.
This mapping supports reproducible scoring
of the recorded judgments.

\paragraph{Sampling and statistical scope.}
The 51 complete submissions cannot be verified
as originating from 51 distinct respondents.
The confidence intervals assume independence
across recorded submissions and may not capture
dependence from repeated participation.

The study covers a category-balanced subset of
500 composites, with 167/167/166 images at B0/B1/B2,
not a verified random sample supporting population-level
estimates over all 14,400 composites. Each image has
only 3--11 judgments, and no matched human Shape-only
baseline is available. Results therefore concern
recognition under these stimuli and answer options,
not human cross-view robustness or geometry-specific
cue use.

These considerations limit population-level
generalization. They do not change the reported
scoring of the collected survey responses.
The VLM comparison uses the shared images and
ten options, while prompts, respondent count,
and aggregation still differ.

\paragraph{Numerical supplement.}
The deidentified numerical supplement includes
\texttt{human\_per\_image.csv},
\texttt{human\_per\_record.csv},
\texttt{human\_per\_form.csv}, and
\texttt{table\_human\_per\_class.csv}.
These tables retain the derived recognition
results, response coverage, and associated
statistical summaries without participant
names or response timestamps.
Overall, the human evaluation provides evidence
of target-category recognizability in Semantic
Apertures. It does not establish that respondents
relied specifically on aperture geometry.
On the shared images and ten options, Gemini
lies above the human rate and GPT lies below it
(Figure~\ref{fig:human_vlm_matched}).

\section{Extended VLM Results and Sensitivity Analyses}
\label{app:extended_vlm}
\label{app:vlm_protocols}

This appendix provides the evaluation protocol, complete
supporting results, and sensitivity analyses for the paired
VLM study reported in Section~\ref{sec:vlm_recognition}.
We analyze two recorded endpoints, GPT-6 Luna and
Gemini 3.8 Flash, covering stimulus preparation,
candidate construction, independent queries, response
parsing, coverage, statistical estimation, and recognition
under competing visual content.
Human evaluation is documented separately in
Appendix~\ref{app:human_validation}.

\subsection{Evaluation Design and Stimulus Preparation}
\label{app:api_coverage}
\label{app:vlm_paired_core}

\paragraph{Image selection and scheduled evaluation.}
The evaluation uses the 500 Final images from the human
study, five images in each of 100 target categories.
The set contains 167 images at $B0$, 167 at $B1$,
and 166 at $B2$. Mirror and carrier attributes vary
across images, but the five queries for one image
share that image's aperture, mirror, and carrier.

For image $i$, the three target-recognition views are
\begin{equation}
    \left(
        X_i^{\mathrm{Original}},
        X_i^{\mathrm{Placed\ Shape},b_i},
        X_i^{\mathrm{Final},b_i}
    \right).
    \label{eq:vlm_paired_views}
\end{equation}

Two additional queries assess carrier identity using
the clean-carrier baseline and the selected Final image.
Each image therefore contributes five queries per endpoint:
three target-recognition and two carrier-recognition tasks.

Each analyzed endpoint has 2,500 scheduled queries:
1,500 target queries and 1,000 carrier queries.
The two-endpoint analysis consequently covers
5,000 scheduled queries. These queries form the basis
of the reported VLM evaluation and comparisons.

Because each image contributes only one aperture level
and one mirror--carrier assignment, comparisons across
$B0$, $B1$, and $B2$ involve different images and assets.
They describe between-image variation rather than
within-image aperture interventions.

\paragraph{Placed Shape-only stimulus.}
Placed Shape-only uses the selected source aperture
and the selected mirror's placement procedure.
The aperture support is rendered in black on an
RGB-127 canvas matching the dimensions of the
corresponding Final image.

Unlike the centered Shape-only diagnostic used
in the encoder and probe experiments, this version
preserves the selected aperture's placement within
the composition canvas.

\paragraph{Clean-carrier stimulus.}
The clean-carrier baseline applies the glass-related
appearance preprocessing to the selected carrier
and cover-crops it to the glass bounding box.
It excludes the aperture, fog, and mirror frame.

The carrier query on Final uses the same composite
image as the Final target-recognition query,
but with a different task instruction.

\paragraph{Stimulus and execution provenance.}
Target queries for both endpoints use one shared
option file and one shared letter order.
That order is also shared by Original, Placed
Shape-only, and Final. Carrier queries use a
separate letter order, shared by the clean-carrier
and Final-carrier queries and by both endpoints.

These checks establish agreement of the recorded
candidate lists and letter orders. They do not
independently revalidate every stimulus pixel.

\subsection{Model Endpoints and Query Coverage}
\label{app:vlm_endpoints}

\paragraph{Recorded endpoints.}
The two evaluated model identifiers are
\texttt{gpt-6-luna} and
\texttt{gemini-3.8-flash}.

Both are provider-hosted aliases with unpinned
revisions, rather than independently verified
checkpoint identities. Their recorded names
identify the configurations used during evaluation,
not immutable underlying model weights.

\paragraph{Query coverage.}
Table~\ref{tab:vlm_coverage} reports the complete
query-status counts and paired availability
for the two analyzed endpoints.

Valid and invalid observed outputs are scored;
invalid outputs count as incorrect. Technical
failures without scored responses remain missing.

All-five coverage counts images with scored
responses to all five tasks for an endpoint.
S/F counts images with paired Placed-Shape and
Final-target responses, while T/C counts images
with paired Final-target and Final-carrier responses.

\begin{table*}[t]
    \centering
    \footnotesize
    \setlength{\tabcolsep}{4pt}
    \renewcommand{\arraystretch}{1.15}
    \caption{
    \textbf{VLM query coverage and paired availability.}
    Valid and invalid outputs are both scored;
    technical failures are missing.
    All five denotes images with scored responses
    to all five queries. S/F denotes paired
    Placed-Shape/Final-target responses, and
    T/C denotes paired Final-target/Final-carrier
    responses. Invalid responses remain incorrect
    within the paired samples.
    }
    \label{tab:vlm_coverage}
    \begin{tabular}{@{}lrrrrrrr@{}}
        \toprule
        Recorded model
        & Scheduled & Valid & Invalid & Missing
        & All five & S/F & T/C \\
        \midrule
        GPT-6 Luna
        & 2,500 & 2,498 & 1 & 1
        & 499 & 499 & 499 \\
        Gemini 3.8 Flash
        & 2,500 & 2,497 & 2 & 1
        & 499 & 500 & 499 \\
        \bottomrule
    \end{tabular}
\end{table*}

Across the two endpoints, the recorded responses
contain 4,995 valid outputs, three invalid outputs,
and two technical failures among 5,000 scheduled
queries.

The two technical failures are length terminations
before a valid candidate identifier: one GPT
Final-target query and one Gemini carrier-on-Final
query. These failures reflect execution or
response-format availability, not observed incorrect
image recognition.

\paragraph{Common evaluation cohort.}
The primary paired analysis retains 498 images
with scored responses to all five tasks from
both endpoints. Inclusion depends on response
availability, not correctness.

All 100 target categories remain represented:
98 categories retain five images and two retain four.

The common cohort permits image-paired comparison
of the two endpoints and their five task conditions
without changing the evaluated image set between
tasks.

\subsection{Candidates, Prompts, and Response Scoring}
\label{app:api_scoring}
\label{app:vlm_candidates}

\paragraph{Target candidate construction.}
Each image has the ten target options used in the
human study: its ground-truth category and nine
alternatives
(Appendix~\ref{app:human_validation}).
Those alternatives were chosen from clean
source-object crops, not from Final-image predictions.

Candidate identities and ordering remain identical
across the three target-recognition views and both
evaluated endpoints. In the 500-image set, the
correct target answer occupies each position A--J
exactly 50 times.

Carrier-recognition queries use the ten carrier
categories with a separately constructed order.
In the 500-image set, the correct carrier likewise
occupies each position A--J exactly 50 times.

\paragraph{Task-specific prompts.}
The target-recognition prompt asks for the category
represented by the target structure and explicitly
directs attention away from unrelated content
visible through the aperture.

The carrier-recognition prompt instead asks for
the identity of the transmitted content rather
than the aperture's source-object category.

Both prompts request exactly one answer identifier
from A--J.

These are task-directed zero-shot evaluations.
In particular, the structure-focused target prompt
does not measure spontaneous cue preference under
an unconstrained instruction.

\paragraph{Independent model requests.}
Each query is submitted as a fresh single-user-message
request containing one image and its task-specific
prompt. Earlier responses and other construction
views are not provided in the same conversation.

For GPT, the runner requests low reasoning effort.
For Gemini, it requests low thinking and
temperature zero.

The supplied runner requests high image detail
and a maximum output length of 1,024 tokens.
It supports fallback behavior if a provider
rejects generation settings, but no exported
query is marked as having used that fallback.

Requested settings do not independently verify
the provider's actual internal processing,
numerical precision, or served model revision.

\paragraph{Strict response parsing.}
The primary parser applies Unicode NFKC
normalization, removes byte-order-mark characters,
trims surrounding whitespace, and converts
the response to uppercase.

It accepts exactly one A--J identifier, optionally
enclosed in a single backtick wrapper.
Explanations, class names, multiple identifiers,
or other output formats are invalid under
the primary scoring rule.

A parsed response is correct only when its
identifier matches the stored correct option.
Invalid observed responses count as incorrect.
Request errors and length-terminated outputs
without a scored candidate remain missing.
The common-cohort analysis requires scored
responses to all five tasks from both endpoints.

\subsection{Statistical Analysis and Reproducibility}
\label{app:api_uncertainty}

\paragraph{Aggregation.}
The primary paired analysis uses the same
498 common images for both endpoints.

Accuracy and paired accuracy changes are
category-macro estimates, giving equal weight
to the 100 target categories.
Conditional recognition retention and joint
target--carrier outcomes instead use pooled
image-level ratios.

The precise estimands are defined in
Appendix~\ref{app:vlm_transition_metrics}.

\paragraph{Bootstrap uncertainty.}
Primary 95\% confidence intervals use 10,000
hierarchical bootstrap replicates with seed
20261010.

Each replicate samples 100 target categories
with replacement and then samples the retained
images within each selected category occurrence.
Repeated category occurrences receive independent
image resamples.

The same resampled images are used across
views and endpoints, preserving pairing.
Conditional metrics are recomputed from their
sampled numerators and denominators rather
than by averaging image-level ratios.

The 2.5th and 97.5th bootstrap percentiles
define the reported 95\% intervals.

\paragraph{Available-case analyses.}
Available-case estimates are computed separately
using the scored image set of each task.
Their intervals use the same hierarchical
category/image bootstrap.

These analyses do not necessarily use identical
image sets across tasks. They are reported
as descriptive complements to the common-cohort
paired analysis.

Neither this resampling procedure nor the
missing-response sensitivity bounds eliminate
selection bias from unobserved technical failures.

\paragraph{Reproduction scope.}
The common-cohort analysis is a post-run
recomputation using the stored query outputs;
it does not involve additional model inference.

The numerical supplement retains the merged
query records, image-level results, common-cohort
manifest, candidate assignments, exact prompt
templates, run metadata, and scoring checks.

The intervals are conditional on the evaluated
stimuli, asset banks, prompts, response records,
and provider-hosted endpoint configurations.

They do not measure repeated-query variability,
model revision drift, unseen-asset generalization,
or robustness to alternative prompting.

\subsection{Transition Accuracy and Conditional Retention}
\label{app:vlm_transition_metrics}

Let $\mathcal{I}_c$ denote the retained images
in target category $c$, with
$n_c=|\mathcal{I}_c|$ and $C=100$.

For model $m$, let $h_{mi}^{v}$ indicate
correct target recognition for image $i$
in view $v$.
Invalid observed outputs have
$h_{mi}^{v}=0$; technical failures are
excluded from the complete common cohort.

Category-macro accuracy, expressed as a
percentage, is
\begin{equation}
    A_m^v
    =
    \frac{100}{C}
    \sum_{c=1}^{C}
    \frac{1}{n_c}
    \sum_{i\in\mathcal{I}_c}
    h_{mi}^{v}.
    \label{eq:vlm_category_macro_accuracy}
\end{equation}

The primary paired accuracy change, in
percentage points, is
\begin{equation}
\begin{aligned}
    \Delta_{\mathrm{Shape}\rightarrow\mathrm{Final},m}
    &=
    \frac{100}{C}
    \sum_{c=1}^{C}
    \frac{1}{n_c}
    \sum_{i\in\mathcal{I}_c}
    \bigl(
        h_{mi}^{\mathrm{Final}}
\\[-2pt]
    &\qquad\qquad
        -
        h_{mi}^{\mathrm{Placed\ Shape}}
    \bigr).
\end{aligned}
    \label{eq:vlm_transition_change}
\end{equation}

This category-macro estimand differs from
a pooled image-average difference when
retained category sizes are unequal.

Conditional retention instead measures the
fraction of initially correct Placed-Shape
cases that remain correct on Final:
\begin{equation}
    R_{\mathrm{Shape}\rightarrow\mathrm{Final},m}
    =
    \frac{
        \sum_i
        h_{mi}^{\mathrm{Placed\ Shape}}
        h_{mi}^{\mathrm{Final}}
    }{
        \sum_i
        h_{mi}^{\mathrm{Placed\ Shape}}
    }.
    \label{eq:vlm_conditional_retention}
\end{equation}

Reported retention percentages equal $100R$.
Analogous Original-to-Final and
clean-carrier-to-Final retention quantities
use the corresponding correctness indicators.
Conditional retention must not be calculated
as a ratio of marginal accuracies.

\paragraph{Paired recognition outcomes.}
On the common 498-image cohort, GPT retains
236 of 397 initially Shape-correct cases,
loses 161, and gains 21 Final successes
following Shape-only failures.

Gemini retains 388 of 460 initially
Shape-correct cases, loses 72,
and gains 8 previously incorrect cases.

The corresponding conditional
Shape-to-Final retention estimates are
59.4\% (95\% CI: [52.3, 66.5])
for GPT and 84.3\% ([79.5, 88.7])
for Gemini.

Category-macro Placed-Shape-to-Final
target accuracy decreases by
28.0 percentage points for GPT
(95\% CI: [$-34.8$, $-21.5$])
and 13.0 points for Gemini
([$-17.8$, $-8.4$]).

Clean-carrier to Final-carrier accuracy
decreases by 38.2 percentage points for GPT
([$-44.3$, $-32.2$]) and 24.1 points
for Gemini ([$-29.3$, $-19.1$]).

The full target and carrier accuracy trajectories
are reported in Section~\ref{sec:vlm_recognition}.

\subsection{Available-Case Recognition Results}
\label{app:vlm_available_results}

Table~\ref{tab:vlm_available_cases} reports all
scored responses from the two evaluated endpoints
under each task condition.

These estimates use pooled scored-image denominators.
Where a task is complete, the pooled percentage
agrees with the category macro at one decimal.
Task columns can contain different image sets;
their marginal accuracies therefore do not constitute
a paired recognition transition.

Coverage and paired availability are summarized
in Table~\ref{tab:vlm_coverage}.

\begin{table*}[t]
    \centering
    \footnotesize
    \setlength{\tabcolsep}{4pt}
    \renewcommand{\arraystretch}{1.2}
    \caption{
    \textbf{Available-case VLM recognition.}
    Each cell reports pooled accuracy (\%),
    its 95\% category/image bootstrap interval,
    and the scored denominator.
    Invalid outputs count as incorrect;
    technical failures are excluded.
    Task columns may contain different images,
    so marginal results do not form a matched transition.
    }
    \label{tab:vlm_available_cases}

    \begin{tabular}{@{}lccccc@{}}
        \toprule
        Recorded model
        & Original
        & \shortstack{Placed\\Shape-only}
        & \shortstack{Final\\target}
        & \shortstack{Clean\\carrier}
        & \shortstack{Final\\carrier} \\
        \midrule

        GPT-6 Luna
        & \shortstack{97.8\\{}[95.6, 99.4]\\$n=500$}
        & \shortstack{79.6\\{}[74.4, 84.6]\\$n=500$}
        & \shortstack{51.5\\{}[44.8, 58.3]\\$n=499$}
        & \shortstack{92.6\\{}[89.6, 95.4]\\$n=500$}
        & \shortstack{54.4\\{}[48.2, 60.6]\\$n=500$} \\
        \addlinespace

        Gemini 3.8 Flash
        & \shortstack{99.4\\{}[98.4, 100.0]\\$n=500$}
        & \shortstack{92.2\\{}[89.0, 95.0]\\$n=500$}
        & \shortstack{79.4\\{}[74.2, 84.4]\\$n=500$}
        & \shortstack{90.2\\{}[86.6, 93.6]\\$n=500$}
        & \shortstack{66.1\\{}[60.6, 71.6]\\$n=499$} \\

        \bottomrule
    \end{tabular}
\end{table*}

The available-case estimates are consistent
with substantial differences in target recognition
between the two endpoints, but direct paired
inference relies on the common 498-image cohort.

The available-case table also shows reduced
carrier recognition on Final relative to the
clean-carrier baseline for both endpoints.
These conditions differ in rendering and task
context, so the observed decline cannot be
assigned to a single compositional factor.

\subsection{Joint Target and Carrier Recognition}
\label{app:vlm_content_task}

Target and carrier correctness on each selected
Final image are obtained from separate
task-specific requests.

Let $t_{mi}$ and $c_{mi}$ denote target and
carrier correctness indicators for model $m$
on image $i$ in the common cohort.

The four mutually exclusive outcomes are
\begin{align}
    \mathrm{Both}_{mi}
        &= t_{mi}c_{mi}, \\
    \mathrm{TargetOnly}_{mi}
        &= t_{mi}(1-c_{mi}), \\
    \mathrm{CarrierOnly}_{mi}
        &= (1-t_{mi})c_{mi}, \\
    \mathrm{Neither}_{mi}
        &= (1-t_{mi})(1-c_{mi}).
\end{align}

Each outcome share is averaged over the same
$N=498$ images. These pooled image shares
are distinct from category-macro accuracies.

\begin{figure}[H]
    \centering
    \includegraphics[width=\columnwidth]
    {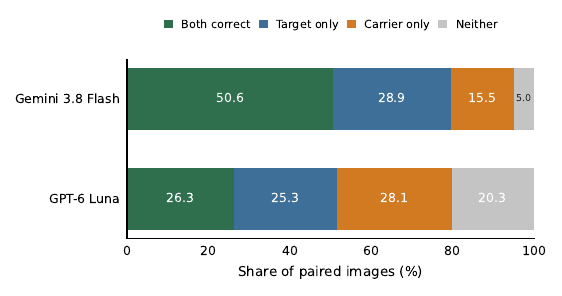}
    \caption{
    \textbf{Joint target and carrier recognition.}
    Stacked segments show pooled image shares
    on the common 498-image cohort.
    Target and carrier answers are obtained
    from separate task-specific requests
    on the same Final image.
    The four outcomes are mutually exclusive
    and exhaustive. Displayed values are rounded
    percentages. These results do not establish
    a causal trade-off between target and carrier
    recognition.
    }
    \label{fig:vlm_joint_outcomes}
\end{figure}

Figure~\ref{fig:vlm_joint_outcomes}
summarizes the four outcomes.

For GPT, the corresponding counts are
131 both correct, 126 target only,
140 carrier only, and 101 neither.

For Gemini, the counts are
252 both correct, 144 target only,
77 carrier only, and 25 neither.

Carrier-only correctness shows that the
carrier category can sometimes be recovered
when target recognition fails.

It does not establish that carrier recognition
caused the target error or that target-related
information is absent from the representation.

Furthermore, carrier recognition also declines
relative to the clean-carrier condition for
both endpoints, as shown in
Section~\ref{sec:vlm_recognition}.

The numerical supplement contains exact joint
counts and hierarchical confidence intervals.
The stacked figure displays point estimates only.

\subsection{Missing Responses and Sensitivity Analyses}
\label{app:vlm_execution_controls}

\paragraph{Missing-response sensitivity.}
We compute additional bounds over all
500 scheduled images using a fixed denominator.

For a given task, let $k$ be the number of
correct scored responses and $u$ the number
of technical failures without scored responses.
Then the possible accuracy range, expressed
as a fraction, is
\begin{equation}
    \left[
        \frac{k}{500},
        \frac{k+u}{500}
    \right].
    \label{eq:vlm_missing_response_bounds}
\end{equation}

The lower bound treats every missing response
as incorrect, whereas the upper bound treats
every missing response as correct.
Invalid observed outputs remain incorrect
under both bounds.

GPT's missing response is a Final-target query,
so its Final-target bounds are 51.4--51.6\%.
Gemini has no missing Final-target response,
so both bounds equal 79.4\%.
Gemini's missing response is a carrier-on-Final
query, with bounds 66.0--66.2\%.
GPT has no missing carrier response.

These are pooled fixed-denominator sensitivity
bounds, not confidence intervals or the
common-cohort category-macro estimates.
The bounds in
Equation~\ref{eq:vlm_missing_response_bounds}
are multiplied by 100 when presented as percentages.

The full task-specific bounds are retained
in the numerical supplement.

\paragraph{Post-hoc formatting sensitivity.}
A secondary parsing check also accepts
an isolated candidate identifier followed
by a period or enclosed in bold markers.

This alternative rule changes no
correct-response totals for GPT or Gemini.
It is excluded from the primary evaluation
and serves only as a formatting-sensitivity check.

Free-form explanations with a trailing
candidate identifier remain invalid under
this limited alternative rule.

\paragraph{Interpretation and unperformed controls.}
The findings concern the two recorded
provider-hosted endpoints, evaluated with
fixed task-specific prompts.

They do not establish a controlled comparison
of independently verified local checkpoints,
different quantization levels, or numerical
precision configurations.

Provider-default precision fields do not
independently establish the inference precision
used internally.

The target prompt explicitly directs attention
toward target-derived structure.
A neutral-versus-structure-focused prompt
comparison was not performed.

Likewise, the submitted results do not include
higher-precision-versus-quantized evaluations,
repeated-query variability tests, or controlled
provider-revision comparisons.

The Placed-Shape-to-Final transition introduces
carrier appearance, fog, and mirror/context
simultaneously. Its recognition changes therefore
cannot be attributed solely to carrier semantics.

Finally, the joint target--carrier outcomes arise
from separate task-specific queries.
They do not establish a causal competition between
target and carrier recognition.

The images and ten target options match the human
study. Prompts, the number of respondents, and
aggregation still differ
(Appendix~\ref{app:human_vlm_protocols}).

Generalization to different prompts, unseen
rendering assets, or future endpoint revisions
remains untested.

\subsection{Summary of VLM Evidence}
\label{app:vlm_summary}

The paired study evaluates target and carrier
recognition on the human-study Final images.

Across 498 common images, both evaluated endpoints
lose target accuracy when moving from Placed
Shape-only to Final, although the magnitude
of the decline and conditional retention differ.
On this cohort, Final-target accuracy is 51.6\%
for GPT and 79.3\% for Gemini.

Both endpoints also exhibit lower carrier
recognition on Final than on the clean-carrier
baseline. Joint-outcome analysis shows that
successful carrier recognition can coexist
with unsuccessful target recognition.

These findings characterize recognition
under the evaluated prompts, answer spaces,
and rendering conditions. They do not identify
a unique visual mechanism or establish
a general causal preference for carrier
content over target-derived geometry.

\section{Aperture-Level and Context Checks}
\label{app:critical_ablations}

This appendix provides the complete aperture-level estimates
and reference-paired context-removal results supporting
Section~\ref{sec:ablations_robustness}.
The analyses retain the frozen benchmark configuration:
one heavy-fog condition without droplets.

\subsection{Recognition Across Aperture Levels}
\label{app:aperture_level_results}

For each source, B0, B1, and B2 use the same assigned mirrors
and carriers. Each level contributes eight Final renders per
source, giving 4,800 renders per level.
Correctness and original-target scores are averaged over
these eight renders before source/category aggregation.

\begin{figure}[H]
    \centering
    \includegraphics[width=\columnwidth]
    {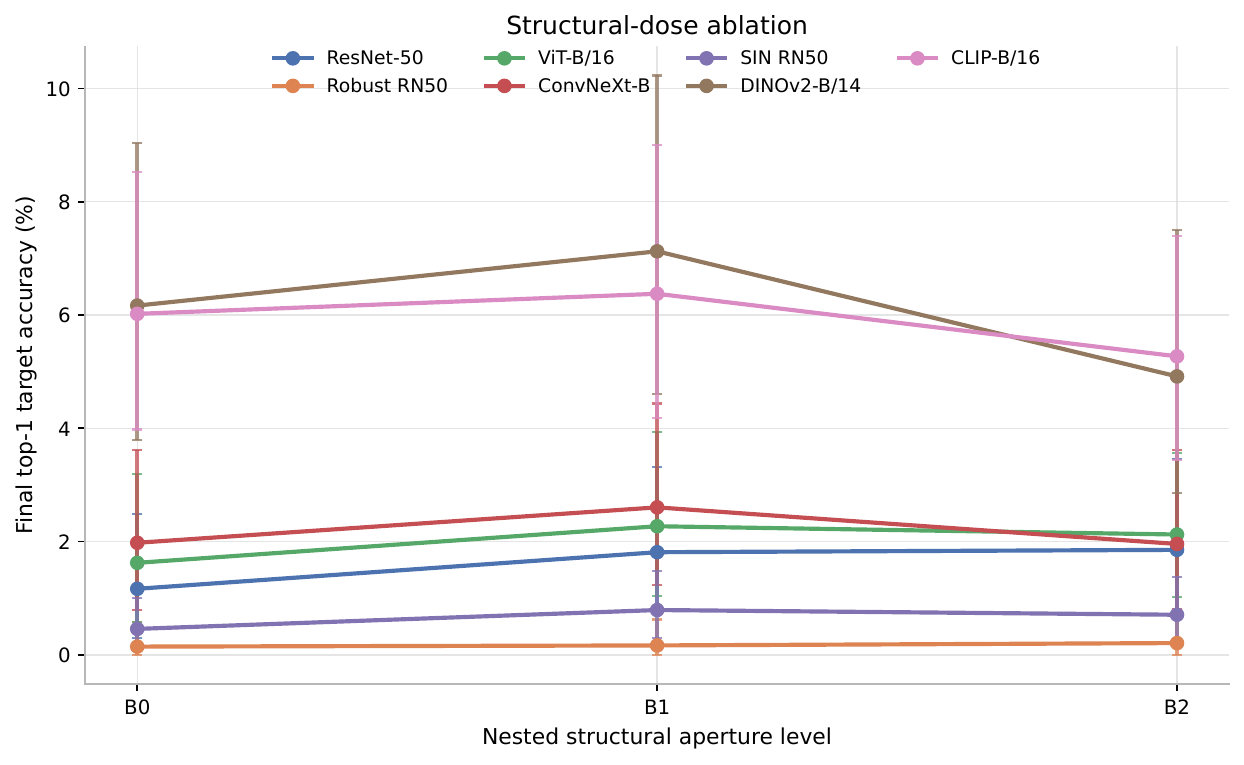}
    \caption{
    Final native ImageNet-1000 top-1 accuracy across the three
    nested source-aperture levels.
    Points show source/category-aggregated estimates and bars
    show stored marginal 95\% confidence intervals.
    The assigned mirrors and carriers are identical across levels.
    Connecting lines organize the levels.
    The paired B1-versus-B2 comparison is reported in
    Table~\ref{tab:b1_b2_paired}.
    }
    \label{fig:aperture_levels_supp}
\end{figure}

Figure~\ref{fig:aperture_levels_supp} shows native
ImageNet-1000 top-1 accuracy.
Table~\ref{tab:aperture_level_results} additionally reports
original-target scores and the stored confidence intervals.
The aperture-level export contains these two metrics;
it does not contain level-specific top-5 estimates.

\begin{table*}[t]
    \centering
    \small
    \renewcommand{\arraystretch}{1.12}
    \caption{
    Complete aperture-level estimates and stored 95\% confidence
    intervals. Each cell reports the estimate above its interval.
    Top-1 values are percentages; target scores retain their
    checkpoint-specific units.
    Every estimate aggregates 600 sources and 100 categories.
    }
    \label{tab:aperture_level_results}
    \begin{tabular}{@{}llccc@{}}
        \toprule
        Model & Metric & B0 & B1 & B2 \\
        \midrule
        ResNet-50
        & Top-1 (\%)
            & \shortstack{1.17\\{}[0.29, 2.48]}
            & \shortstack{1.81\\{}[0.75, 3.31]}
            & \shortstack{1.85\\{}[0.75, 3.46]} \\
        & Target score
            & \shortstack{0.456\\{}[0.306, 0.626]}
            & \shortstack{0.600\\{}[0.429, 0.786]}
            & \shortstack{0.625\\{}[0.454, 0.811]} \\
        \midrule
        Robust RN50
        & Top-1 (\%)
            & \shortstack{0.15\\{}[0.00, 0.58]}
            & \shortstack{0.17\\{}[0.00, 0.62]}
            & \shortstack{0.21\\{}[0.00, 0.75]} \\
        & Target score
            & \shortstack{$-0.315$\\{}[$-0.718$, 0.109]}
            & \shortstack{$-0.288$\\{}[$-0.684$, 0.123]}
            & \shortstack{$-0.252$\\{}[$-0.653$, 0.162]} \\
        \midrule
        ViT-B/16
        & Top-1 (\%)
            & \shortstack{1.62\\{}[0.58, 3.19]}
            & \shortstack{2.27\\{}[1.04, 3.94]}
            & \shortstack{2.12\\{}[1.02, 3.56]} \\
        & Target score
            & \shortstack{0.528\\{}[0.312, 0.780]}
            & \shortstack{0.691\\{}[0.441, 0.960]}
            & \shortstack{0.727\\{}[0.485, 0.999]} \\
        \midrule
        ConvNeXt-B
        & Top-1 (\%)
            & \shortstack{1.98\\{}[0.79, 3.63]}
            & \shortstack{2.60\\{}[1.23, 4.44]}
            & \shortstack{1.96\\{}[0.81, 3.63]} \\
        & Target score
            & \shortstack{0.668\\{}[0.445, 0.901]}
            & \shortstack{0.839\\{}[0.599, 1.088]}
            & \shortstack{0.765\\{}[0.536, 1.017]} \\
        \midrule
        SIN RN50
        & Top-1 (\%)
            & \shortstack{0.46\\{}[0.10, 1.00]}
            & \shortstack{0.79\\{}[0.29, 1.48]}
            & \shortstack{0.71\\{}[0.23, 1.38]} \\
        & Target score
            & \shortstack{0.170\\{}[$-0.330$, 0.680]}
            & \shortstack{0.224\\{}[$-0.280$, 0.734]}
            & \shortstack{0.230\\{}[$-0.263$, 0.737]} \\
        \midrule
        DINOv2-B/14
        & Top-1 (\%)
            & \shortstack{6.17\\{}[3.79, 9.04]}
            & \shortstack{7.13\\{}[4.60, 10.23]}
            & \shortstack{4.92\\{}[2.85, 7.50]} \\
        & Target score
            & \shortstack{3.159\\{}[2.497, 3.844]}
            & \shortstack{3.283\\{}[2.601, 3.986]}
            & \shortstack{2.713\\{}[2.072, 3.388]} \\
        \midrule
        CLIP-B/16
        & Top-1 (\%)
            & \shortstack{6.02\\{}[3.98, 8.52]}
            & \shortstack{6.38\\{}[4.19, 9.00]}
            & \shortstack{5.27\\{}[3.44, 7.40]} \\
        & Target score
            & \shortstack{24.429\\{}[24.044, 24.817]}
            & \shortstack{24.402\\{}[24.020, 24.786]}
            & \shortstack{24.076\\{}[23.699, 24.452]} \\
        \bottomrule
    \end{tabular}
\end{table*}

B1 has higher top-1 point estimates than B2 for ViT-B/16,
ConvNeXt-B, SIN RN50, DINOv2-B/14, and CLIP-B/16.
Standard and robust ResNet-50 instead have their highest
point estimates at B2.

Target scores and accuracy do not always follow the same
ordering. For example, ViT-B/16 and SIN RN50 have higher
mean target scores at B2 than B1 despite lower top-1 accuracy.
This illustrates why target-score changes and correct
classification are reported separately.

Table~\ref{tab:b1_b2_paired} reports an exploratory paired
comparison. Each source has the same eight mirror--carrier
assignments at B1 and B2. Top-1 and target logit are averaged
over those eight renders, and the reported value is the class
macro of the within-source B1-minus-B2 difference.
The top-1 interval excludes zero only for DINOv2-B/14, at
$+2.21$ percentage points (95\% CI: $[1.19,3.42]$).
The target-logit interval excludes zero for ConvNeXt-B,
DINOv2-B/14, and CLIP-B/16 in favor of B1, and for Robust
RN50 in favor of B2.
Holm adjustment is applied within the seven checkpoints of
one metric and is separate from the published 28-contrast
geometry family.

\begin{table*}[t]
    \centering
    \small
    \caption{
    Exploratory paired B1-minus-B2 differences.
    Top-1 entries are percentage points. Target-logit entries
    retain checkpoint-specific units.
    Each source contributes the mean of eight matched
    mirror--carrier assignments, and the estimate is the class
    macro of that within-source difference.
    Intervals are 95\% hierarchical bootstrap intervals.
    Holm adjustment is within the seven checkpoints of the
    same metric and is separate from the published geometry
    family. Positive values favor B1.
    }
    \label{tab:b1_b2_paired}
    \begin{tabular}{@{}lcccccc@{}}
        \toprule
        Model
        & Top-1 (pp) & 95\% CI & $p_{\mathrm{Holm}}$
        & Logit & 95\% CI & $p_{\mathrm{Holm}}$ \\
        \midrule
        ResNet-50
            & $-0.04$ & [$-0.71$, 0.69] & 1.00000
            & $-0.025$ & [$-0.075$, 0.026] & 0.18149 \\
        Robust RN50
            & $-0.04$ & [$-0.25$, 0.00] & 1.00000
            & $-0.036$ & [$-0.054$, $-0.017$] & 0.00035 \\
        ViT-B/16
            & 0.15 & [$-0.63$, 1.15] & 1.00000
            & $-0.036$ & [$-0.101$, 0.035] & 0.18149 \\
        ConvNeXt-B
            & 0.65 & [$-0.02$, 1.50] & 0.02910
            & 0.075 & [0.009, 0.146] & 0.00040 \\
        SIN RN50
            & 0.08 & [$-0.19$, 0.42] & 1.00000
            & $-0.007$ & [$-0.054$, 0.041] & 0.64682 \\
        DINOv2-B/14
            & 2.21 & [1.19, 3.42] & 0.00035
            & 0.570 & [0.444, 0.709] & 0.00035 \\
        CLIP-B/16
            & 1.10 & [$-0.17$, 2.56] & 0.03775
            & 0.326 & [0.218, 0.445] & 0.00035 \\
        \bottomrule
    \end{tabular}
\end{table*}

The levels retain progressively more source-coordinate
support, but their comparison jointly changes transmitting
area, spatial organization, and potentially placement scale
(Appendix~\ref{app:aperture_details}).

\subsection{Reference-Paired Context Removal}
\label{app:context_removal_results}

The context analysis compares the correct B1 reference with
its matched context-removed condition.
Before JPEG encoding, pixels outside the glass mask are
replaced by RGB value 127; the reference content inside
the glass is retained.

For source $i$, define
\begin{equation}
d_i^{\mathrm{context}}
=
z_{y_i}(x_i^{\mathrm{context\ removed}})
-
z_{y_i}(x_i^{\mathrm{reference}}).
\label{eq:context_removal_effect}
\end{equation}
Positive effects indicate higher original-target scores
after surrounding-context removal.

Table~\ref{tab:context_reference_paired} reports the
source-paired estimates. Six confidence intervals lie
above zero and have Holm-adjusted $p<0.01$.
Robust RN50 has an interval including zero.
These results concern original-target scores; they do
not establish recovery of correct classification.

\begin{table*}[t]
    \centering
    \small
    \caption{
    Context-removed minus correct-reference original-target
    scores. Each checkpoint uses 600 paired B1 references
    from 100 categories.
    Intervals are stored 95\% source/category bootstrap
    intervals after sign reversal.
    Holm-adjusted $p$-values retain the original 28-contrast
    geometry/context family.
    Score units are checkpoint-specific.
    }
    \label{tab:context_reference_paired}
    \begin{tabular}{@{}lrcr@{}}
        \toprule
        Model & Effect & 95\% CI & $p_{\mathrm{Holm}}$ \\
        \midrule
        ResNet-50
            & 0.226 & [0.124, 0.336] & 0.00140 \\
        Robust RN50
            & 0.029 & [$-0.179$, 0.228] & 0.58007 \\
        ViT-B/16
            & 0.483 & [0.301, 0.675] & 0.00140 \\
        ConvNeXt-B
            & 0.557 & [0.399, 0.725] & 0.00140 \\
        SIN RN50
            & 0.836 & [0.520, 1.167] & 0.00140 \\
        DINOv2-B/14
            & 1.053 & [0.733, 1.387] & 0.00140 \\
        CLIP-B/16
            & 0.850 & [0.612, 1.092] & 0.00140 \\
        \bottomrule
    \end{tabular}
\end{table*}

The original contrast export uses reference minus
context-removed scores. For presentation here, its
estimate $\delta$ and interval $[L,U]$ become
$-\delta$ and $[-U,-L]$.
This sign conversion requires no new bootstrap and
does not change the two-sided test or correction family.

\subsection{Baseline Alignment and Crop Interpretation}
\label{app:context_baseline_scope}

The context intervention uses one selected B1 render
per source. Its baseline is that same reference image,
rather than the source's average over all 24 Final renders.
Comparing the latter average with context-removed images
would also change the distribution of mirrors, carriers,
and aperture levels.

The mirror-region condition crops the reference to the
glass-mask bounding box. Subsequent model preprocessing
changes the effective scale of the aperture and visible
carrier content. Cropping therefore combines context
removal with a change in image framing and scale.
No localization gain is inferred from an unmatched
full-Final baseline.

The aperture and paired-context exports are retained in
\texttt{stencil\_levels.csv} and
\texttt{context\_reference\_paired.csv}.
The paired B1--B2 estimates are retained in
\texttt{item11\_b1\_b2.csv}.
Variation across fixed rendering assets is reported
separately in Appendix~\ref{app:factor_heterogeneity};
historical fog and droplet calibration is documented in
Appendix~\ref{app:pilot_calibration}.

\section*{Appendix Part II: Additional Analyses and Extended Material}

\section{Rendering-Factor Heterogeneity}
\label{app:factor_heterogeneity}

This appendix characterizes variation within the fixed rendering
asset bank. These exploratory summaries complement the paired
interventions, but are not used to establish causal effects of
individual rendering factors or generalization to unseen assets.

\subsection{Factor-Level Aggregation}
\label{app:factor_aggregation}

We group Final images by mirror shape, frame style, surrounding
type, carrier category, and aperture level.
These factors contain five, three, three, ten, and three levels,
respectively.

Within each checkpoint and factor level, original-target scores
are first averaged over the relevant renders of each source.
The estimate then averages source means within category and
gives equal weight to the categories represented in that stratum.
Stored 95\% confidence intervals use 2,000 source/category
bootstrap replicates
(Appendix~\ref{app:bootstrap_details}).

The groups overlap because a source contributes multiple
rendering assignments. Each mirror-shape and carrier-category
stratum contains 240 sources, and each aperture-level stratum
contains all 600 sources.
Frame-style strata contain 399 sources for frameless mirrors,
386 for regular frames, and 380 for ornate frames.
Surrounding strata contain 393 sources for the stored
\texttt{none} level, 394 for texture, and 393 for object surrounds.

These unique-source counts differ from the assignment counts
in Table~\ref{tab:benchmark_balance}: a source can have two
assigned mirrors sharing a frame style or surrounding type.
Factor strata therefore do not represent disjoint samples.

\subsection{Display Centering}
\label{app:factor_display_centering}

\begin{figure*}[p]
    \centering
    \includegraphics[width=\textwidth]
    {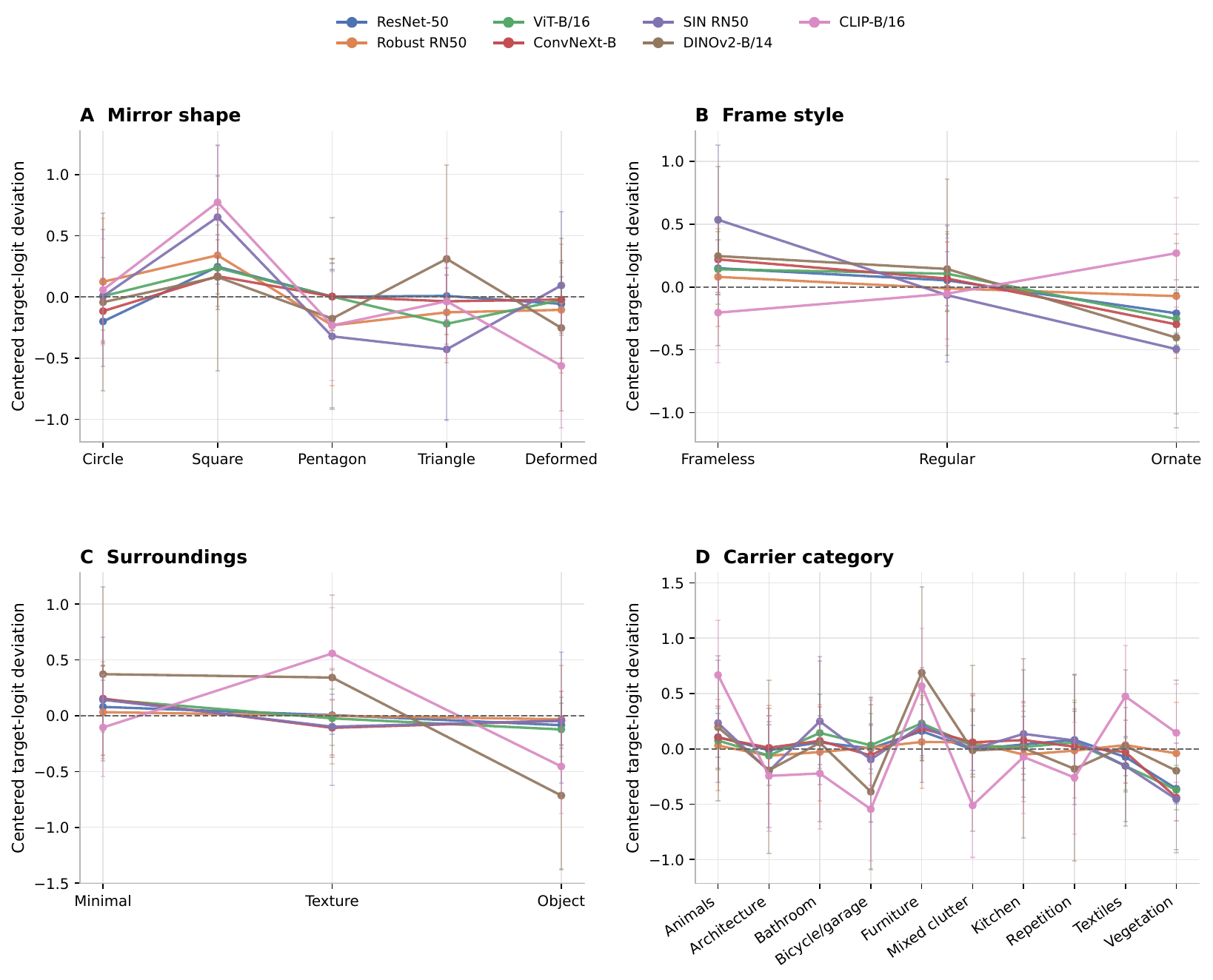}
    \caption{
    Descriptive target-score variation within the fixed asset bank.
    Top left: mirror shape; top right: frame style;
    bottom left: surrounding type; bottom right: carrier category.
    Each checkpoint/factor profile is centered using
    Equation~\ref{eq:factor_display_center}.
    Error bars are raw 95\% intervals shifted by that same center;
    they are not confidence intervals for paired deviations
    from zero. Lines connect categorical levels for readability.
    The label Minimal corresponds to the stored surrounding
    level \texttt{none}; carrier labels are shortened for display.
    }
    \label{fig:factor_heterogeneity_supp}
\end{figure*}

Figure~\ref{fig:factor_heterogeneity_supp} displays the four
asset-factor profiles. For checkpoint $m$, factor $f$, and
level $\ell$, let $\overline{s}_{m,f,\ell}$ be the exported
target-score estimate and $n_{f,\ell}$ its source count.
The display center is
\begin{equation}
c_{m,f}
=
\frac{
\sum_{\ell}n_{f,\ell}\overline{s}_{m,f,\ell}
}{
\sum_{\ell}n_{f,\ell}
},
\label{eq:factor_display_center}
\end{equation}
and the plotted value is
\begin{equation}
d_{m,f,\ell}
=
\overline{s}_{m,f,\ell}-c_{m,f}.
\label{eq:factor_display_deviation}
\end{equation}

The center is computed separately for every checkpoint/factor
profile. It is a source-count-weighted summary of the exported
stratum means, rather than necessarily the checkpoint's overall
Final mean.

If the raw interval is $[L_{m,f,\ell},U_{m,f,\ell}]$,
its displayed endpoints are
\begin{equation}
\left[
L_{m,f,\ell}-c_{m,f},
U_{m,f,\ell}-c_{m,f}
\right].
\label{eq:factor_display_interval}
\end{equation}
This operation shifts the stored interval by a fixed descriptive
center. It does not propagate uncertainty in the center or
estimate a paired level-versus-center difference.
Consequently, the dashed zero line is a display reference,
not a significance threshold.

Centering removes an additive offset but does not standardize
score scales across checkpoints. Differences in profile amplitude
must therefore be interpreted within checkpoint.

\subsection{Ranges of Uncentered Estimates}
\label{app:factor_score_ranges}

Table~\ref{tab:factor_ranges} summarizes the minimum and maximum
exported mean score across levels of each factor.
For example, DINOv2-B/14 surrounding-type means range from
2.337 to 3.425, while its carrier-category means range from
2.657 to 3.732. These values describe variation among the
observed strata.

The extrema are selected descriptively. Their difference is
not a source-paired intervention estimate, and no significance
test of the minimum versus maximum is reported.
Aperture-level ranges are included for completeness;
their full estimates and intervals appear in
Appendix~\ref{app:critical_ablations}.

\begin{table*}[t]
    \centering
    \footnotesize
    \setlength{\tabcolsep}{4pt}
    \caption{
    Minimum and maximum uncentered mean original-target scores
    across factor levels.
    Each pair is $(\mathrm{minimum},\mathrm{maximum})$;
    these pairs are not confidence intervals.
    Columns summarize five mirror shapes, three frame styles,
    three surrounding types, ten carrier categories, and three
    aperture levels. Score units are checkpoint-specific.
    }
    \label{tab:factor_ranges}
    \begin{tabular}{@{}lccccc@{}}
        \toprule
        Model
        & Mirror shape
        & Frame
        & Surround
        & Carrier category
        & Aperture level \\
        \midrule
        ResNet-50
            & $(0.343,0.790)$
            & $(0.360,0.719)$
            & $(0.479,0.642)$
            & $(0.197,0.717)$
            & $(0.456,0.625)$ \\
        Robust RN50
            & $(-0.500,0.072)$
            & $(-0.351,-0.197)$
            & $(-0.314,-0.251)$
            & $(-0.362,-0.237)$
            & $(-0.315,-0.252)$ \\
        ViT-B/16
            & $(0.414,0.870)$
            & $(0.408,0.803)$
            & $(0.525,0.795)$
            & $(0.265,0.866)$
            & $(0.528,0.727)$ \\
        ConvNeXt-B
            & $(0.623,0.908)$
            & $(0.470,0.987)$
            & $(0.648,0.910)$
            & $(0.314,0.938)$
            & $(0.668,0.839)$ \\
        SIN RN50
            & $(-0.225,0.855)$
            & $(-0.270,0.760)$
            & $(0.114,0.356)$
            & $(-0.262,0.442)$
            & $(0.170,0.230)$ \\
        DINOv2-B/14
            & $(2.755,3.319)$
            & $(2.674,3.324)$
            & $(2.337,3.425)$
            & $(2.657,3.732)$
            & $(2.713,3.283)$ \\
        CLIP-B/16
            & $(23.719,25.054)$
            & $(24.107,24.581)$
            & $(23.835,24.847)$
            & $(23.778,24.990)$
            & $(24.076,24.429)$ \\
        \bottomrule
    \end{tabular}
\end{table*}

\subsection{Interpretation and Numerical Supplement}
\label{app:factor_heterogeneity_scope}

The profiles describe target-score heterogeneity within the
observed bank. Factor levels correspond to different fixed
assets and source subsets; they can also interact with aperture
placement, carrier appearance, and other rendering properties.
Global assignment balance does not make these strata a fully
crossed, within-source intervention design.

Accordingly, the profiles do not isolate the effect of changing
only mirror shape, frame, or surrounding type.
They also do not measure carrier-category recognition,
establish accuracy differences from score differences,
or evaluate new asset banks.

The raw export,
\texttt{factor\_heterogeneity.csv}, contains 168 estimates:
seven checkpoints crossed with the 24 levels across all five
factors. It retains means, interval endpoints, and source counts.
The four-panel figure displays the 147 estimates for the
21 asset-factor levels. Their centered values and shifted
intervals are retained in
\texttt{factor\_centered\_for\_display.csv}.

These descriptive analyses remain separate from the explicitly
paired geometry and context effects in
Appendices~\ref{app:geometry_carrier_results}
and~\ref{app:critical_ablations}.

\section{Historical Pilot and Calibration}
\label{app:pilot_calibration}

\subsection{Purpose and Scope}
\label{app:pilot_scope}

We conducted a preliminary calibration study to examine
structural coverage, carrier visibility, and rendering
behavior before constructing the final benchmark.
The pilot used 30 ImageNet-S source photographs from
30 categories: ten animals and five categories each
of devices, vehicles, household objects, and food.
These categories belong to the benchmark's 100-category
vocabulary and span varied structural properties,
including thin parts, openings, wings, and wheels.

The calibration examined aperture construction and
synthetic rendering independently of the recognition
experiments. It assessed structural preservation and
rendering properties rather than classification accuracy.
The pilot and final benchmark therefore represent
distinct experimental stages, and calibration measurements
are not treated as recognition or robustness results.

\subsection{Structural and Rendering Calibration}
\label{app:pilot_sweeps}

\paragraph{Structural aperture sweep.}
We evaluated seven nested aperture-coverage targets:
\[
    0.20,\;0.27,\;0.34,\;0.41,\;0.48,\;0.55,\;0.62.
\]
For each source, candidate masks were constructed
by retaining the highest-ranked pixels from a common
structural-priority map within the object segmentation.
This procedure produced 210 candidate masks across
30 source photographs.

We measured realized coverage, nesting, support
confinement, connected components, and retention
of object boundaries, edges, local detail, and
thin structures.

Following this analysis, coverage targets of
34\%, 48\%, and 62\% were selected as the low,
medium, and high pilot levels, respectively.
Their mean realized coverage closely matched
the requested fractions.

Mean boundary recall increased from 82.73\%
at 34\% coverage to 87.84\% at 48\% and
91.74\% at 62\%. Mean edge retention increased
from 66.64\% to 77.73\% and 86.71\%,
respectively. Thin-structure retention similarly
increased from 50.25\% to 63.22\% and 74.76\%.

These measurements characterize the trade-off
between aperture coverage and structural-feature
retention in the pilot construction. They do not
establish improved recognition or isolate the
effect of geometry from the amount of transmitted
carrier appearance.

\paragraph{Fog calibration.}
We evaluated seven fog-rendering recipes, including
light, medium, and heavy configurations and four
intermediate or bracketing variants.

The recipes jointly varied opacity, clipping limits,
background and aperture transmission, blur, bloom,
and color mixing. Thus, their effects cannot be
attributed to opacity alone.

The initial fog sweep used a fixed mirror--carrier
configuration, a common underlying fog-field
realization, and a 41\% pilot aperture target.
It produced 210 composites across the 30 sources
without a droplet overlay.

Following calibration, the light, medium, and heavy
recipes were retained as the three pilot fog levels.
The selected heavy configuration uses opacity
0.97, clipping limits $[0.93,0.995]$,
background transmission 0.006, and aperture
transmission 0.52.

\paragraph{Joint calibration and context variation.}
We evaluated the interaction between aperture coverage,
fog rendering, and condensation treatment using
12 source photographs.

The complete factorial grid crossed seven aperture
targets, seven fog recipes, and two condensation
conditions, yielding
\[
    12\times7\times7\times2=1{,}176
\]
composite observations.

The selected three aperture levels and three fog
recipes define a further subset of
\[
    12\times3\times3\times2=216
\]
calibration observations.

A separate context-variation experiment used
ten sources, ten mirror--carrier assignments per
source, and the selected aperture, fog, and
condensation conditions. This produced
\[
    10\times10\times3\times3\times2=1{,}800
\]
additional composite observations.

These experiments extended calibration beyond
the initial fixed mirror--carrier configuration,
allowing rendering diagnostics to be examined
under different spatial structures and visual
contexts. They were not used to estimate
recognition performance or generalization
to unseen rendering assets.

\subsection{Calibration Diagnostics and Interpretation}
\label{app:pilot_diagnostics}

Structural diagnostics measured aperture coverage,
nesting, off-support leakage, connected components,
and retention of boundaries, edges, local detail,
and thin structures.

Rendering diagnostics compared each composite
with its corresponding clean carrier inside
the mirror glass region. They included luminance
contrast retention, gradient retention, and separate
contrast measurements inside and outside the
aperture.

Placement checks measured aperture clipping,
while compositing checks examined preservation
of the mirror surroundings outside the glass
region before image encoding.

The calibration evaluated both a fog-only veil
and a synthetic condensation-relief condition.
These conditions modify rendering appearance
rather than providing a physically measured
simulation of condensation.

The measurements supported parameter selection
and examination of rendering behavior.
They do not establish human recognizability,
perceptually uniform differences between settings,
or robustness of the evaluated recognition models.

The independent 500-image human-recognition study
is reported in Section~\ref{sec:human_recognition}
and Appendix~\ref{app:human_validation}.

\subsection{Relationship to the Frozen Benchmark}
\label{app:pilot_final_relationship}

The historical pilot and final benchmark use
different structural-aperture construction rules.

The pilot apertures were generated by selecting
fixed fractions of pixels from a structural-priority
ranking. In contrast, the final benchmark constructs
$B0$ and $B1$ using structural extraction thresholds,
with $B2$ extending the retained support toward a
55\% coverage target when necessary
(Appendix~\ref{app:aperture_details}).

Consequently, the selected pilot fractions of
34\%, 48\%, and 62\% are not fixed coverage
definitions for the final $B0$, $B1$, and $B2$
apertures. The final levels are nested in source
coordinates, but their realized coverage varies
across source objects.

The frozen benchmark also uses a single
heavy-fog configuration without droplets,
as specified in Appendix~\ref{app:rendering_details}.
Its background and aperture transmissions are
$t_{\mathrm{bg}}=0.006$ and
$t_{\mathrm{clear}}=0.52$, respectively.

The additional fog recipes and condensation
conditions investigated during calibration are
not factors in the final recognition evaluation.
Accordingly, the reported recognition results
do not establish robustness across fog strengths,
droplet conditions, or physical condensation.

The aperture-level and context interventions
performed on the frozen benchmark are reported
separately in
Appendix~\ref{app:critical_ablations}.


\section{Semantic alignment of Final top-1 predictions}
\label{app:prediction_alignment}

Each Final top-1 prediction is labeled from the photo it was rendered with, not from the carrier category. A class is eligible for a photo only when it occurs as a stored Final top-1 index, on any of the seven models, for a render that uses that photo. Eligible classes were marked before the shares below were computed. A class is marked on a carrier photo when that object is visible in the photo. It is marked on a mirror photo when that object is visible outside the glass. The glass region is the mirror mask at luminance of at least 128, and it is not part of the surround.

If the winning class is marked on that image's carrier photo, the prediction is carrier. If it is marked on that image's mirror surround and not on the carrier photo, the prediction is context. Otherwise it is other. When the same object is visible in both photos, carrier takes priority. If a visible object could be either of two ImageNet classes and the photo does not decide, neither class is marked. A related or typical class is not marked unless that object is present. A correct target class follows the same rule, so a correct prediction is other unless the target object is visible in the carrier or the surround.

The neither rule covers the ambiguous cases in these photographs. Unlabeled toiletry bottles are not marked as soap dispensers or lotion; a pump bottle alone beside a bathroom faucet is marked as a soap dispenser. A sofa is not a studio couch, a living-room rug is not a prayer rug or a doormat, and cloth towels are not paper towels. A road or hybrid bicycle is not a mountain bike. A cropped vessel that could be a cup, a vase, or a pitcher is none of those. The brown striped cat is marked tabby and not tiger cat. A wall-mounted bathroom cabinet is a medicine chest; a wooden vanity is not. A flowerpot is pot. A built-in oven or cooktop is stove. A freestanding wooden dining table is a dining table; a counter, desk, or workbench is not.

A review of all 75 photographs produced the maps below. Washbasin is unmarked on the frameless circle surround because the basin bowl is not visible there. The bath mats in the second and third bathrooms are marked doormat. The jacketed books in the third clutter photograph are marked book jacket, and the closed laptop in the office photograph is marked laptop. The second garage is marked carton, coil for the coiled hose, and mountain bike for the green bicycle. A living-room rug, a shop rug, a window with no mesh screen, a portable radio, pantry shelves, and the drop-bar bicycle in the third garage are unmarked.

The shares are class-macro means. Each source contributes the fraction of its 24 Final renders in a segment, those source fractions are averaged within the ImageNet class, and the 100 classes are averaged without weights. Intervals are 95\% hierarchical bootstrap intervals with 10{,}000 replicates, resampling sources within a class and then resampling classes. This is semantic agreement between the winning class name and an object visible in the named photo. It is not a measurement of where the model looked.

\begin{figure}[H]
    \centering
    \includegraphics[width=\columnwidth]
    {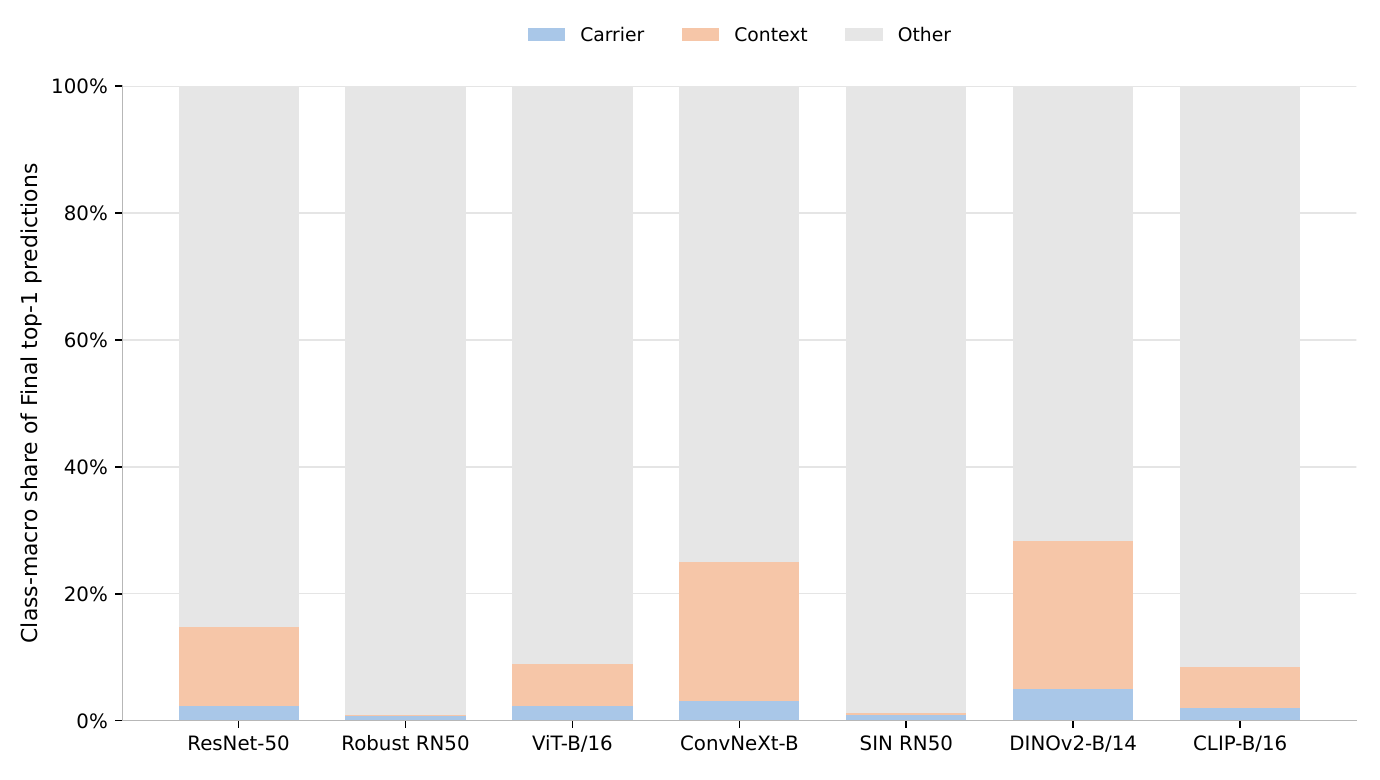}

\caption{
\textbf{What Final top-1 predictions name.}
Class-macro share of stored Final top-1 predictions.
Carrier matches an object in that render's carrier
photograph, context matches the mirror surround but
not the carrier, and Other includes all unmatched
predictions, including correct target labels not marked
as visible in either photograph.
Context is washbasin, except for SIN RN50, whose
context predictions are soap dispenser.
}

    \label{fig:prediction_alignment}
\end{figure}


\begin{table*}[t]
\centering
\small
\setlength{\tabcolsep}{8pt}
\caption{Class-macro share of Final top-1 predictions (percent). Intervals are 95\% hierarchical bootstrap intervals.}
\label{tab:prediction_alignment}
\begin{tabular}{lccc}
\toprule
Model & Carrier & Context & Other \\
\midrule
ResNet-50
& 2.3 (1.7--3.0)
& 12.4 (10.3--14.8)
& 85.3 (82.9--87.6) \\

Robust RN50
& 0.7 (0.4--1.1)
& 0.1 (0.0--0.2)
& 99.2 (98.8--99.5) \\

ViT-B/16
& 2.2 (1.6--2.9)
& 6.6 (5.0--8.4)
& 91.1 (89.2--93.0) \\

ConvNeXt-B
& 3.1 (2.4--4.0)
& 21.9 (19.3--24.6)
& 75.0 (72.1--77.7) \\

SIN RN50
& 0.9 (0.5--1.3)
& 0.4 (0.1--0.7)
& 98.8 (98.2--99.3) \\

DINOv2-B/14
& 4.9 (4.0--6.0)
& 23.3 (20.6--26.1)
& 71.7 (68.8--74.7) \\

CLIP-B/16
& 2.1 (1.5--2.7)
& 6.4 (4.8--8.1)
& 91.6 (89.7--93.3) \\
\bottomrule
\end{tabular}
\end{table*}


\begin{table*}[t]
\centering
\footnotesize
\setlength{\tabcolsep}{4pt}
\renewcommand{\arraystretch}{1.12}
\caption{Carrier-photo map after human review. Only classes that are visible in that photo and that occur in the stored Final top-1 predictions are listed. Photos with no such class are omitted.}
\label{tab:carrier_map}

\begin{tabular}{@{}p{0.42\textwidth}p{0.54\textwidth}@{}}
\toprule
\textbf{Carrier photo} & \textbf{Marked classes} \\
\midrule

\texttt{\detokenize{aspect_animal_objects_01.png}}
& tabby, lampshade, pillow \\

\texttt{\detokenize{aspect_animal_objects_02.png}}
& golden retriever \\

\texttt{\detokenize{aspect_animal_objects_03.png}}
& pot \\

\texttt{\detokenize{aspect_architectural_geometry_01.png}}
& bannister \\

\texttt{\detokenize{aspect_architectural_geometry_03.png}}
& bannister \\

\texttt{\detokenize{aspect_chairs_furniture_01.png}}
& coffee mug \\

\texttt{\detokenize{aspect_chairs_furniture_02.png}}
& coffee mug \\

\texttt{\detokenize{aspect_chairs_furniture_03.png}}
& desk, laptop \\

\texttt{\detokenize{aspect_mixed_clutter_01.png}}
& analog clock, backpack, carton, dining table, pitcher \\

\texttt{\detokenize{aspect_mixed_clutter_02.png}}
& carton, coil \\

\texttt{\detokenize{aspect_mixed_clutter_03.png}}
& carton, dining table, lampshade, table lamp, book jacket, plate \\

\texttt{\detokenize{aspect_organized_repetition_02.png}}
& bookcase, lampshade, library, table lamp \\

\texttt{\detokenize{aspect_organized_repetition_03.png}}
& carton \\

\texttt{\detokenize{aspect_textile_materials_02.png}}
& quilt \\

\texttt{\detokenize{aspect_textile_materials_03.png}}
& pillow \\

\texttt{\detokenize{aspect_vegetation_01.png}}
& pot \\

\texttt{\detokenize{aspect_vegetation_02.png}}
& greenhouse, pot \\

\texttt{\detokenize{aspect_vegetation_03.png}}
& pot \\

\texttt{\detokenize{scene_bathroom_01.png}}
& shower curtain, toilet seat, washbasin \\

\texttt{\detokenize{scene_bathroom_02.png}}
& bathtub, doormat, medicine chest, soap dispenser \\

\texttt{\detokenize{scene_bathroom_03.png}}
& doormat, shower curtain, soap dispenser, washbasin, toilet tissue \\

\texttt{\detokenize{scene_garage_01.png}}
& carton, disk brake, mountain bike \\

\texttt{\detokenize{scene_garage_02.png}}
& carton, coil, mountain bike \\

\texttt{\detokenize{scene_garage_03.png}}
& screwdriver \\

\texttt{\detokenize{scene_kitchen_01.png}}
& spatula, teapot \\

\texttt{\detokenize{scene_kitchen_02.png}}
& mixing bowl, stove, toaster, plate \\

\texttt{\detokenize{scene_kitchen_03.png}}
& pitcher, stove, strainer, plate \\

\bottomrule
\end{tabular}
\end{table*}


\begin{table*}[t]
\centering
\footnotesize
\setlength{\tabcolsep}{4pt}
\renewcommand{\arraystretch}{1.12}
\caption{Mirror-surround map after human review. Objects inside the glass are not marked. Mirror photos with no marked class are omitted.}
\label{tab:mirror_map}

\begin{tabular}{@{}p{0.42\textwidth}p{0.54\textwidth}@{}}
\toprule
\textbf{Mirror photo} & \textbf{Marked classes} \\
\midrule

\texttt{\detokenize{mirror_circle_object_frameless.png}}
& soap dispenser \\

\texttt{\detokenize{mirror_circle_object_ornate.png}}
& washbasin \\

\texttt{\detokenize{mirror_circle_object_regular.png}}
& soap dispenser, washbasin \\

\texttt{\detokenize{mirror_deformed_object_frameless.png}}
& washbasin \\

\texttt{\detokenize{mirror_deformed_object_ornate.png}}
& soap dispenser, washbasin \\

\texttt{\detokenize{mirror_deformed_object_regular.png}}
& soap dispenser, washbasin \\

\texttt{\detokenize{mirror_pentagon_object_frameless.png}}
& washbasin \\

\texttt{\detokenize{mirror_pentagon_object_ornate.png}}
& washbasin \\

\texttt{\detokenize{mirror_pentagon_object_regular.png}}
& washbasin \\

\texttt{\detokenize{mirror_square_none_ornate.png}}
& washbasin \\

\texttt{\detokenize{mirror_square_object_frameless.png}}
& soap dispenser, washbasin \\

\texttt{\detokenize{mirror_square_object_ornate.png}}
& soap dispenser, washbasin \\

\texttt{\detokenize{mirror_square_object_regular.png}}
& washbasin \\

\texttt{\detokenize{mirror_triangle_object_frameless.png}}
& soap dispenser \\

\texttt{\detokenize{mirror_triangle_object_ornate.png}}
& washbasin \\

\texttt{\detokenize{mirror_triangle_object_regular.png}}
& washbasin \\

\bottomrule
\end{tabular}
\end{table*}



\end{document}